\documentclass[10pt,journal,compsoc]{IEEEtran}

\ifCLASSOPTIONcompsoc
  \usepackage[nocompress]{cite}
\else
  \usepackage{cite}
\fi
\ifCLASSINFOpdf
\else
\fi
\usepackage{amsmath}
\usepackage{amssymb}
\usepackage{amsthm}
\usepackage{hyperref}
\usepackage{multirow}
\usepackage{graphicx}
\usepackage{adjustbox}
\usepackage[section]{placeins}
\usepackage[table]{xcolor}

\DeclareMathOperator*{\argmin}{arg\,min}

\begin{document}
%
\title{Hitchhiker's Guide to Super-Resolution: Introduction and Recent Advances}
%
%
%
%

\author{Brian B. Moser$^{1,2}$, Federico Raue$^{1}$, Stanislav Frolov$^{1}$, Sebastian Palacio$^{1}$, Jörn Hees$^{1}$, Andreas Dengel$^{1,2}$\\
$^1$ German Research Center for Artificial Intelligence (DFKI), Germany\\
$^2$ TU Kaiserslautern, Germany\\
{\tt\small first.second@dfki.de}}

%
%

\markboth{IEEE TRANSACTIONS ON PATTERN ANALYSIS AND MACHINE INTELLIGENCE,~Vol.~0, No.~0, 2023}%
{Moser \MakeLowercase{\textit{et al.}}: Hitchhiker's Guide to Super-Resolution: Introduction and Recent Advances}
%



\IEEEtitleabstractindextext{%
\begin{abstract}
With the advent of Deep Learning (DL), Super-Resolution (SR) has also become a thriving research area. However, despite promising results, the field still faces challenges that require further research, e.g., allowing flexible upsampling, more effective loss functions, and better evaluation metrics. We review the domain of SR in light of recent advances and examine state-of-the-art models such as diffusion (DDPM) and transformer-based SR models. We critically discuss contemporary strategies used in SR and identify promising yet unexplored research directions. We complement previous surveys by incorporating the latest developments in the field, such as uncertainty-driven losses, wavelet networks, neural architecture search, novel normalization methods, and the latest evaluation techniques. We also include several visualizations for the models and methods throughout each chapter to facilitate a global understanding of the trends in the field. This review ultimately aims at helping researchers to push the boundaries of DL applied to SR.
\end{abstract}

}

\maketitle

\IEEEdisplaynontitleabstractindextext

%
\IEEEpeerreviewmaketitle

\ifCLASSOPTIONcompsoc
\IEEEraisesectionheading{\section{Introduction}\label{sec:introduction}}
\else
\section{Introduction}
\label{sec:introduction}
\fi

%
%
%
%
\IEEEPARstart{S}{uper-Resolution} (SR) is the process of enhancing Low-Resolution (LR) images to High-Resolution (HR). 
The applications range from natural images \cite{zeyde2010single, martin2001database} to highly advanced satellite \cite{valsesia2021permutation}, and medical imaging \cite{bashir2021comprehensive}.
Despite its long history \cite{9044873}, SR remains a challenging task in computer vision because it is notoriously ill-posed: several HR images can be valid for any given LR image due to many aspects like brightness and coloring \cite{anwar2020densely, sun2020learned}.
The fundamental uncertainties in the relation between LR and HR images pose a complex research task.
Thanks to rapid advances in Deep Learning (DL), SR has made significant progress in recent years.
Unfortunately, entry into this field is overwhelming because of the abundance of publications.
It is knotty work to get an overview of the advantages and disadvantages between publications.
This work unwinds some of the most representative advances in the crowded field of SR.

Existing surveys have primarily focused on classical methods \cite{Yang2014SingleImageSA, Thapa2016APC}, whereas we, in contrast, concentrate exclusively on DL methods given their proven superiority. 
The most similar survey to this work is the one by Bashir et al. \cite{bashir2021comprehensive}, which itself is a logical continuation of Wang et al. \cite{9044873}.
However, our work focuses on methods that have recently gained popularity: uncertainty-driven loss, advances in image quality assessment methods, new datasets, denoising diffusion probabilistic models, advances in normalization techniques, and new architecture approaches.
This work also reviews pioneering work that combines neural architecture search with SR, which automatically derives designs instead of relying solely on human dexterity \cite{chu2020multi}.
Finally, this work aims to give a broader overview of the field and highlight challenges as well as future trends.

\autoref{sec:definitions} lays out definitions and introduces known metrics used by most SR publications.
This section also introduces datasets and data types found in SR research.
\autoref{sec:learning_objectives} presents commonly used regression-based learning objectives (pixel, uncertainty-driven, and content loss) as well as more sophisticated objectives like adversarial loss and denoising diffusion probabilistic models. 
\autoref{sec:upsampling_methods} covers interpolation- and learning-based upsampling.
\autoref{sec:attentionMechanisms} goes over the basic mechanisms of attention used in SR.
\autoref{sec:additionalStrategies} explains additional learning strategies, covering a variety of techniques that can be applied for performance improvement.
It discusses curriculum learning, enhanced predictions, network fusion, multi-task learning, and normalization.
\autoref{sec:srmodels} introduces SR models. It goes through different architecture types, such as simple, residual, recurrent-based, lightweight, and wavelet transform-based networks.
Several graphical visualizations (also in supplementary material) accompany the chapter, highlighting the difference between the proposed architectures.
\autoref{sec:unsupervised} and \autoref{sec:nas} introduce unsupervised SR and neural architecture search combined with SR.
Finally, this work summarizes and points to future directions in \autoref{sec:discussion} and \autoref{sec:conclusion}.

\section{Setting and Terminology}
\label{sec:definitions}
This section focuses on four questions:
(1) What is SR? (2) How good is a stated SR solution? (3) What SR datasets are available to test the solution? (4) How are images represented in SR?

The first question introduces the fundamental definitions as well as the terminology that is specific to SR.
The second relates to evaluation metrics that assess any proposed SR solution, such as Peak Signal-to-Noise Ratio (PSNR) and Structural Similarity Index (SSIM). 
The third question is linked to numerous datasets that provide various data types, such as 8K resolution images or video sequences.
Last but not least, the fourth question addresses different image representations (i.e., color spaces) that SR models can use.

\subsection{Problem Definition: Super-Resolution}
\label{sec:probdef}
Super-Resolution (SR) refers to methods that can develop High-Resolution (HR) images from at least one Low-Resolution (LR) image \cite{9044873}.
The SR field divides into Single Image Super-Resolution (SISR) and Multi-Image Super-Resolution (MISR).
In SISR, one LR image leads to one HR image, whereas MISR generates many HR images from many LR images.
Regarding popularity, most researchers focus on SISR because its techniques are extendable to MISR. 
\subsubsection{Single Image Super-Resolution (SISR)}
\label{sec:sisr}
The goal of Single Image Super-Resolution (SISR) is to scale up a given Low-Resolution (LR) image $\mathbf{x}  \in \mathbb{R}^{\bar{w} \times \bar{h} \times c}$ to a High-Resolution (HR) image $\mathbf{y} \in \mathbb{R}^{w \times h \times c}$, with $\bar{w} \leq w$ and $\bar{h} \leq h$. 
Throughout this work, $N_\mathbf{x} = w \cdot h \cdot c$ defines the amount of pixels of an image $\mathbf{x} \in \mathbb{R}^{w \times h \times c}$ and $\Omega_\mathbf{x}$ the set of all valid positions in $\mathbf{x}$:
\begin{equation}
    \Omega_\mathbf{x} = \{ \left( i,j,k \right) \in \mathbb{N}_1^3 | i \leq h, j \leq w, k \leq c \}
\end{equation}
Let $s \in \mathbb{N}_1$ be a scaling factor, it holds that $h = s \cdot \bar{h}$ and $w = s \cdot \bar{w}$.
Furthermore, let $\mathcal{D}: \mathbb{R}^{w \times h \times c} \rightarrow \mathbb{R}^{\bar{w} \times \bar{h} \times c}$ be a degradation mapping that describes the inherent relationship between the two entities LR ($\mathbf{x}$) and HR ($\mathbf{y}$):
\begin{equation}
    \mathbf{x} = \mathcal{D} \left( \mathbf{y}; \delta \right),
\end{equation}
in which $\delta$ are parameters of $\mathcal{D}$ that contain, for example, the scaling factor $s$ and other elements like blur type.

In practice, the degradation mapping is often unknown and therefore modeled, e.g., with bicubic downsampling.
The underlying challenge of SISR is to perform the inverse mapping of $\mathcal{D}$. Unfortunately, this problem is ill-posed because one LR image can lead to multiple nonidentical HR images. 
The goal is to find a SR model $\mathcal{M}: \mathbb{R}^{\bar{w} \times \bar{h} \times c} \rightarrow \mathbb{R}^{w \times h \times c}$, s.t.: 
\begin{equation}
    \mathbf{\hat{y}} = \mathcal{M} \left( \mathbf{x}; \theta \right),
\end{equation}
where $\mathbf{\hat{y}}$ is the predicted HR approximation of the LR image $\mathbf{x}$ and $\theta$ the parameters of $\mathcal{M}$. 

For Deep Learning (DL), this translates into an optimization objective that minimizes the difference between the estimation $\mathbf{\hat{y}}$ and the ground-truth HR image $\mathbf{y}$ under a given loss function $\mathcal{L}$: 
\begin{equation}
    \hat{\theta} = \argmin_{\theta} \mathcal{L} \left( \mathbf{\hat{y}}, \mathbf{y}\right)
\end{equation}
The setting is illustrated in the supplementary material.
\subsubsection{Multi-Image Super-Resolution (MISR)}
Multi-Image Super-Resolution (MISR) is the task of yielding one or more HR images from many LR images \cite{kappeler2016video}. 
An example is satellite imagery \cite{valsesia2021permutation}, where many LR examples direct to a single HR prediction, a so-called many-to-one approach. 
An alternative is the many-to-many approach, where many LR images lead to more than one HR image. It is usually employed for video sequence enhancing \cite{kappeler2016video}. 
The LR images are generally of the same scene, e.g., multiple satellite images of the same geographical location. 
Given a sequence $\mathbf{x} = \left( \mathbf{x}_1, ..., \mathbf{x}_T\right)$ with $T \in \mathbb{N}_1$ and $\mathbf{x}_t \in \mathbb{R}^{\bar{w} \times \bar{h} \times c}$, $0 < t \leq T$, the task is to predict $\mathbf{y} = \left( \mathbf{y}_1, ..., \mathbf{y}_{T'}\right)$ with $T' \in \mathbb{N}_1$ and $\mathbf{y}_{t'} \in \mathbb{R}^{w \times h \times c}$, $0 < t' \leq T'$. 
The most frequent case is $T = T'$, where $\mathbf{y}_{t}$ is supposed to be the HR image of $\mathbf{x}_t$. 
Generally, MISR is an extension of the SISR setting. 

\subsection{Evaluation: Image Quality Assessment (IQA)}
Many properties are associated with excellent image quality, such as sharpness, contrast, or the absence of noise. Thus, fair evaluation of SR models is challenging. 
This section shows different evaluation methods that fall under the umbrella term Image Quality Assessment (IQA).
Broadly speaking, IQA refers to any metric based on perceptual assessments of human viewers, i.e., how realistic the image appears after applying SR methods.
IQA can be subjective (e.g., human raters) or objective (e.g., formal metrics). 

\subsubsection{Mean Opinion Score (MOS)}

Digital images are ultimately meant to be viewed by human beings. 
Thus, the most appropriate way of assessing images is a subjective evaluation \cite{sajjadi2017enhancenet, ledig2017photo}. 
One commonly used subjective IQA method is the Mean Opinion Score (MOS). 
Human viewers rate images with quality scores, typically 1 (bad) to 5 (good). 
MOS is the arithmetic mean of all ratings. 
Despite reliability, mobilizing human resources is time-consuming and cumbersome, especially for large datasets.

\subsubsection{Peak Signal-to-Noise Ratio (PSNR)}
\label{sec:psnr}
Objectively assessing quality is of indisputable importance due to the massive amount of images produced in recent years and the weaknesses of subjective measurements. 
One popular objective quality measurement is Peak Signal-to-Noise Ratio (PSNR). 
It is the ratio between the maximum possible pixel-value $L$ (255 for 8-bit representations) and the Mean Squared Error (MSE) of reference images. 
Given the approximation $\mathbf{\hat{y}}$ and the ground-truth $\mathbf{y}$, PSNR is a logarithmic quantity using the decibel scale [dB]:
\begin{equation}
    \text{PSNR}  \left( \mathbf{y}, \mathbf{\hat{y}} \right) = 10 \cdot \log_{10} \frac{L^2}{\frac{1}{N_\mathbf{y}} \sum_{p \in \Omega_\mathbf{y}} \left[ \mathbf{y}_p - \mathbf{\hat{y}}_p \right]^2}
\end{equation}

Although it is widely used as an evaluation criterion for SR models, it often leads to mediocre results in real scenarios. 
It focuses on pixel-level differences instead of mammalian visual perception, which is more attracted to structures \cite{wang2004image}. 
Subsequently, it correlates poorly with subjectively perceived quality. 
Slight changes in pixels (e.g., shifting) can lead to a significantly decreased PSNR, while humans barely recognize the difference. 
Consequently, new metrics focus on more structural features within an image. 

\subsubsection{Structural Similarity Index (SSIM)}

The Structural Similarity Index (SSIM) depends on three relatively independent entities: luminance, contrast, and structures \cite{wang2004image}. 
It is widely known and better meets the requirements of perceptual assessment \cite{9044873}.
SSIM estimates for an image $\mathbf{y}$ the luminance $\mu_\mathbf{y}$ as the mean of the intensity, while it is estimating contrast $\sigma_\mathbf{y}$ as its standard deviation:
\begin{equation}
    \mu_\mathbf{y} = \frac{1}{N_\mathbf{y}} \sum_{p \in \Omega_\mathbf{y}} \mathbf{y}_p,
\end{equation}
\begin{equation}
    \sigma_\mathbf{y} = \frac{1}{N_\mathbf{y} - 1} \sum_{p \in \Omega_\mathbf{y}} \left[ \mathbf{y}_p - \mu_\mathbf{y}\right]^2
\end{equation}
In order to compare the entities, the authors of SSIM introduced a similarity comparison function $S$:
\begin{equation}
    \label{eq:simfun}
    S \left( x, y, c\right) = \frac{2 \cdot x \cdot y + c}{x^2 + y^2 + c}\,,
\end{equation}
where $x$ and $y$ are the compared scalar variables, and $c = \left(k \cdot L \right)^2$, $0 < k \ll 1$ is a constant to avoid instability.
Given a ground-truth image $\mathbf{y}$ and its approximation $\mathbf{\hat{y}}$, the comparisons on luminance ($\mathcal{C}_l$) and contrast ($\mathcal{C}_c$) are
\begin{equation}
    \label{eq:ssim_luminance}
    \mathcal{C}_l \left( \mathbf{y}, \mathbf{\hat{y}} \right) = 
    S \left( \mu_\mathbf{y}, \mu_\mathbf{\hat{y}}, c_1\right) \text{ and } \mathcal{C}_c \left( \mathbf{y}, \mathbf{\hat{y}} \right) = 
    S \left( \sigma_\mathbf{y}, \sigma_\mathbf{\hat{y}}, c_2 \right)
\end{equation}
where $c_1, c_2 > 0$.
The empirical co-variance
\begin{equation}
    \sigma_{\mathbf{y}, \mathbf{\hat{y}}} = \frac{1}{ N_\mathbf{y} - 1} \sum_{p \in \Omega_\mathbf{y}} \left( \mathbf{y}_p- \mu_\mathbf{y} \right) \cdot \left( \mathbf{\hat{y}}_p - \mu_\mathbf{\hat{y}} \right),
\end{equation}
determines the structure comparison ($\mathcal{C}_s$), expressed as the correlation coefficient between $\mathbf{y}$ and $\mathbf{\hat{y}}$:
\begin{equation}
    \mathcal{C}_s \left( \mathbf{y}, \mathbf{\hat{y}} \right) = \frac{\sigma_{\mathbf{y}, \mathbf{\hat{y}}} + c_3}{\sigma_\mathbf{y} \cdot \sigma_\mathbf{\hat{y}} + c_3},
\end{equation}
where $c_3 > 0$. Finally, the SSIM is defined as:
\begin{equation}
    \label{eq:ssim}
    \text{SSIM} \left( \mathbf{y}, \mathbf{\hat{y}}\right) = \left[ \mathcal{C}_l \left( \mathbf{y}, \mathbf{\hat{y}} \right)\right]^\alpha \cdot \left[ \mathcal{C}_c \left( \mathbf{y}, \mathbf{\hat{y}} \right)\right]^\beta \cdot\left[ \mathcal{C}_s \left( \mathbf{y}, \mathbf{\hat{y}} \right)\right]^\gamma
\end{equation}
where $\alpha > 0, \beta > 0$ and $\gamma > 0$ are adjustable control parameters for weighting relative importance of all components.

\subsubsection{Learning-based Perceptual Quality (LPQ)}
Lately, researchers have tried to mitigate some weak points of MOS by using DL, the so-called Learning-based Perceptual Quality (LPQ).
In essence, LPQ tries to approximate a variety of subjective ratings by applying DL methods.
One way is to use datasets that contain subjective scores, such as TID2013 \cite{ponomarenko2015image}, and a neural network to predict human rating scores, e.g., DeepQA \cite{kim2017deep} or NIMA \cite{talebi2018nima}. 

A significant drawback of LPQ is the limited availability of annotated samples.
One can augment a small-sized dataset by applying noise and ranking. 
Adding minimal noise to an image should lead to poorer quality, making the noisy image and the original counterpart pairwise discriminable.
These pairs are called Quality-Discriminable Image Pairs (DIP) \cite{ma2017dipiq}. 
The DIP Inferred Quality (dipIQ) index uses RankNet \cite{burges2005learning}, which is based on a pairwise learning-to-rank algorithm \cite{liu2011learning}.
The authors of dipIQ show higher accuracy and robustness in variation-rich content than by training directly on IQA databases like TID2013 \cite{ponomarenko2015image}.
Another example is RankIQA \cite{liu2017rankiqa}, in which a Siamese Network \cite{koch2015siamese} is used to rank image quality by using artificial distortions.

Another inventive way to calculate the similarity between two images is to use DL to extract and compare features. One well-known representative is Learned Perceptual Image Patch Similarity (LPIPS) \cite{zhang2018unreasonable}, which uses $L$ feature maps generated by an extractor $\varphi$, e.g., VGG \cite{simonyan2014very}. Let $H_l$, $W_l$ be the height and width of the $l$-th feature map, respectively, and $\alpha_l \in \mathbb{R}^{C_l}$ a scaling vector, then LPIPS is formulated as
\begin{equation}
\text{LPIPS} \left( \mathbf{y}, \mathbf{\hat{y}}\right) = \sum^L_{l=1} \frac{\sum_{h,w} \left\lVert \alpha_l \odot \left( \varphi^l \left( \mathbf{\hat{y}} \right)_{h,w} - \varphi^l \left(  \mathbf{y} \right)_{h,w}\right) \right\rVert^2_2}{H_lW_l}
\end{equation}
The authors showed that LPIPS aligns better with human judgments than PSNR or SSIM. However, the quality depends on the feature extractor underneath.

Another example is Deep Image Structure and Texture Similarity (DISTS) \cite{ding2020image}, which combines spatial averages (texture) with the correlations of feature maps (structure).

\subsubsection{Task-based Evaluation (TBE)}
\label{sec:tbe}
Alternatively, one can focus on task-oriented features. For instance, one can measure quality differences from DL models that solve other Computer Vision (CV) tasks like image classification.
Other CV tasks can also benefit from including SR as a pre-processing step. Measuring the performance on tasks, with and without SR, is yet another ingenious way to measure the quality of an SR method.
HR images provide more details, which are highly desirable for CV tasks like object recognition \cite{ valsesia2021permutation}.
Nevertheless, it requires extra training steps, which can be avoided by using predefined features like those presented in the following section.

\subsubsection{Evaluation with defined Features}
\label{sec:definedFeatures}
One example is the Gradient Magnitude Similarity Deviation (GMSD) \cite{xue2013gradient}, which uses the pixel-wise Gradient Magnitude Similarity (GMS).
Based on the variation of local quality, which arises from the diversity of local image structures,
GMS calculates the gradient magnitudes with:
\begin{equation}
    \mathbf{m}_\mathbf{x} \left( p \right) = \sqrt{ \left( \nabla_h \mathbf{x} \right)^2_p + \left( \nabla_v \mathbf{x} \right)^2_p}, p \in \Omega_\mathbf{x}
\end{equation}
where $\nabla_h \mathbf{x}$ and $\nabla_v \mathbf{x}$ are the horizontal and vertical gradient images of $\mathbf{x}$, respectively.
The GMS map, similar to \autoref{eq:ssim_luminance}, is given by
\begin{equation}
    \label{eq:gms}
    \text{GMS} \left( \mathbf{y}, \mathbf{\hat{y}} \right)_p = 
    S \left(  \mathbf{m}_\mathbf{y} \left( p \right), \mathbf{m}_\mathbf{\hat{y}} \left( p \right), c\right)
\end{equation}
where $c$ is a positive constant.
The final IQA score is given by the average of the GMS map:
\begin{equation}
    \label{eq:gmsd}
    \text{GMSD} \left( \mathbf{y}, \mathbf{\hat{y}} \right) = \frac{1}{N_\mathbf{y}} \sum_{p \in \Omega_\mathbf{y}} \text{GMS} \left( \mathbf{y}, \mathbf{\hat{y}} \right)_p
\end{equation}
    
An alternative is the Feature Similarity (FSIM) Index. It also uses gradient magnitudes, but combines them with Phase Congruency (PC), a local structure measurement, as feature points \cite{zhang2011fsim}. 
The phase congruency model postulates a biologically plausible model of how human visual systems detect and identify image features. 
These features are perceived where the Fourier components are maximal in phase.
An extension to FSIM is the Haar wavelet-based Perceptual Similarity Index (HaarPSI), which uses a Haar wavelet decomposition to assess local similarities \cite{reisenhofer2018haar}. 
Its authors claim that it outperforms state-of-the-art IQA methods like SSIM and FSIM in terms of execution time and higher correlations with human opinion scores.
They suggest that a multi-scale complex-valued wavelet filterbank directly influences the computation of FSIM in the computation of the Fourier components.
The approach is similar to FSIM but relies on 2D discrete Haar wavelet transform.
    
\subsubsection{Multi-Scale Evaluation}
In practice, SR models usually super-resolve to different scaling factors, known as Multi-Scaling (MS). Thus, evaluating metrics should address this scenario.
The MS-Structural Similarity (MS-SSIM) index \cite{wang2003multiscale} is a direct extension of SSIM that incorporates MS. 
Compared to SSIM, it can be more robust to variations in viewing conditions.
It adds parameters that weight the relative importance of different scales.
MS-SSIM applies a low-pass filter and downsamples the filtered image by a factor of 2 for every scale level $1 \leq i \leq s_{max}$, where $s_{max}$ is the largest scale. 
By doing so, it calculates the contrast and structure comparisons at each scale and the luminance comparison for the largest scale $s_{max}$.
MS-SSIM is formulated as 
    \begin{equation}
        \text{MS-SSIM} \left( \mathbf{y}, \mathbf{\hat{y}}\right) \\ = \mathcal{C}_{l}  \left( \mathbf{y}, \mathbf{\hat{y}} \right)^{\alpha} \cdot \prod_{i=1}^{s_{max}}  \mathcal{C}_{c, i} \left( \mathbf{y}, \mathbf{\hat{y}} \right)^{\beta_i} \cdot \mathcal{C}_{s, i} \left( \mathbf{y}, \mathbf{\hat{y}} \right)^{\gamma_i},
    \end{equation}
    where $\alpha\negmedspace >\negmedspace 0$, $\beta_i\negmedspace >\negmedspace 0$ and $\gamma_i\negmedspace >\negmedspace 0$ are adjustable control parameters. 
    Similarly, the GMSD can be extended with 
    \begin{equation}
        \text{MS-GMSD} \left( \mathbf{y}, \mathbf{\hat{y}}\right) = \sqrt{\sum^{s_{max}}_{i=0} \alpha_i \cdot \left(\text{GMSD}_i \left( \mathbf{y}, \mathbf{\hat{y}}\right) \right)^2},
    \end{equation}
    where $\alpha_i$ are adjustable control parameters and $\text{GMSD}_i$ is the GMSD score at $i_{th}$ scale \cite{zhang2017gradient}.

\subsection{Datasets and Challenges}
\begin{figure}
    \begin{center}
        \includegraphics[width=.45\textwidth]{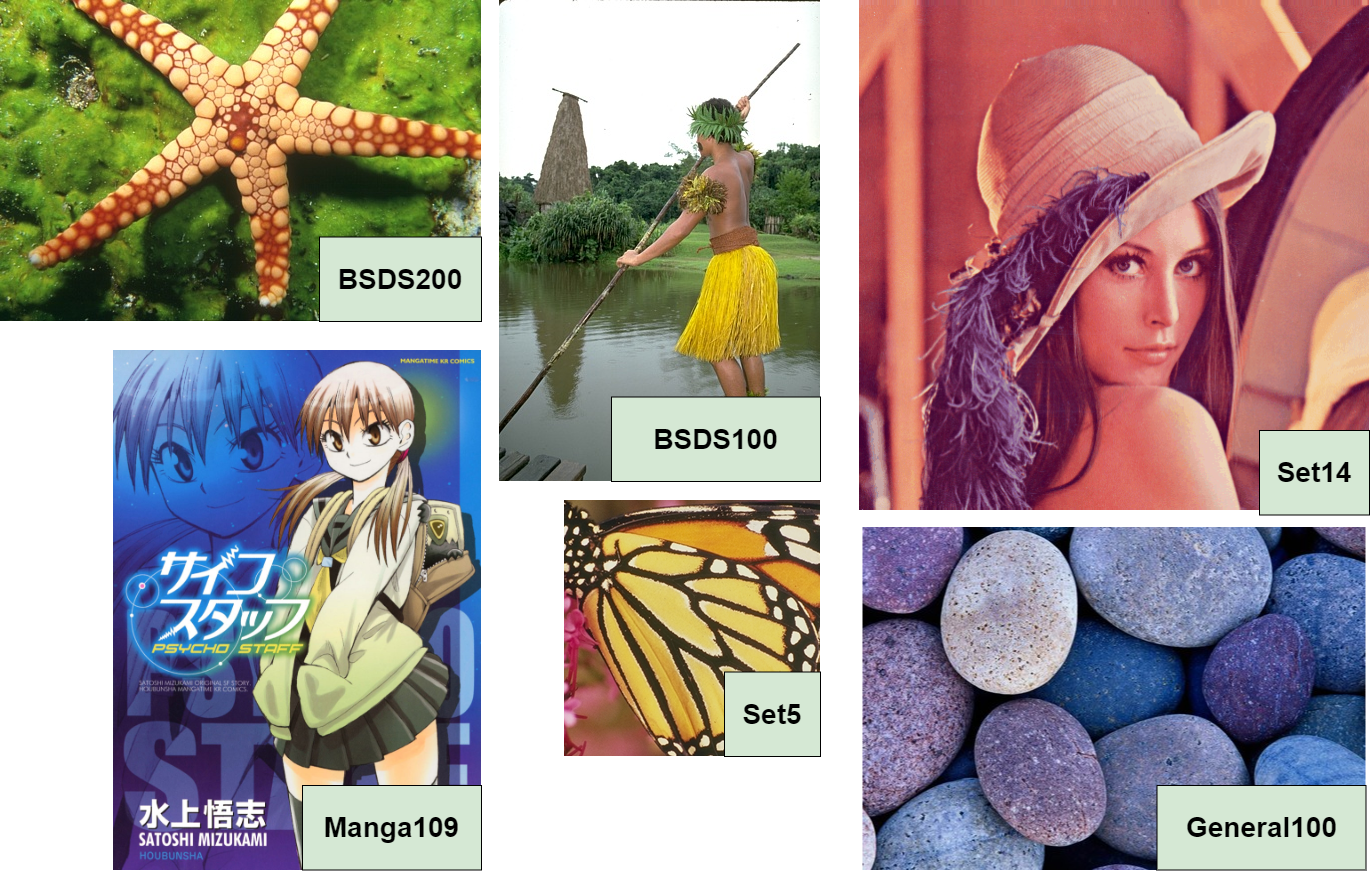}
        \caption{
        Example images from different SR datasets: 
        Set5 \cite{bevilacqua2012low},
        Set14 \cite{zeyde2010single},
        Manga109 \cite{matsui2017sketch},
        General100 \cite{dong2016accelerating},
        BSDS100 \cite{martin2001database},
        and BSDS200 \cite{martin2001database}.
        The ratio of the size differences is preserved.
        }
        \label{fig:examples_datasets}
    \end{center}
\end{figure}
For SR, more extensive datasets have been made available over the past years.
Various accessible datasets contain vast amounts of images of different qualities and content, as shown in \autoref{fig:examples_datasets}. 
Some are created explicitly for SR in a supervised manner with LR-HR pairs. 
Moreover, one can exploit datasets that do not come with SR annotations by generating LR pairs and treating original samples as the HR version. LR samples are typically computed using bicubic interpolation, and anti-aliasing \cite{MATLAB:2017b}.
The supplementary material list commonly used datasets. 
Some of them were published as part of a SR challenge. 
Two of the most famous challenges are the New Trends in Image Restoration and Enhancement (NTIRE) challenge \cite{lugmayr2020ntire}, and the Perceptual Image Restoration and Manipulation (PIRM) challenge \cite{blau20182018}.

\subsection{Color Spaces}
Applications of SR can expand into other domains with additional modalities such as depth- or hyper-spectral SR. However relevant, we restrict the scope of this review to the more prevalent use case of color images (i.e., c=3).
Data representation (i.e., color space) has played a crucial role in SR methods before DL.
Nowadays, most researchers use the RGB color space \cite{zhang2018image, lim2017enhanced}.
Nonetheless, researchers have tried to combine the advantages of other color spaces. 
The first DL-based SR model \cite{dong2015image} used the first channel of the YCbCr space.
It is standard to capture only the Y channel if YCbCr is employed \cite{tong2017image} since focusing on structure instead of color is preferred.
Recent SR models use the RGB color space, also because using something other than RGB color spaces is ill-defined when comparing state-of-the-art results if PSNR is used as evaluation metric \cite{hu2019meta, zhang2018residual}.
Exploring other color spaces for DL-based SR methods is nearly nonexistent, which presents an exciting research gap. 

\section{Learning Objectives}
\label{sec:learning_objectives}
The ultimate goal of SR is to provide a model that maps an LR image to an HR image.
Therefore, the question naturally arises: how to train a given SR model?
The answer to this question is given in the following sections.

\subsection{Regression-based Objectives}
Regression-based objectives attempt to model the relation between input and the desired output explicitly. 
The parameters of the SR model are estimated directly from the data, often by minimizing the $L1$ and $L2$ losses.
While both loss functions are easily applicable, the results tend to be blurry.
We also discuss one way to mitigate this shortcoming, namely by modeling uncertainty into the loss function itself.

\subsubsection{Pixel Loss}
The pixel loss measures the pixel-wise difference, as illustrated in the supplementary material.
There are two well-known pixel loss functions in literature. 
The first one is the Mean Absolute Error (MAE), or $L1$-loss:
\begin{equation}
    \label{eq:l1}
    \mathcal{L}_{\text{L1}} \left( \mathbf{y}, \mathbf{\hat{y}}\right) =  \frac{1}{N_\mathbf{y}} \sum_{p \in \Omega_\mathbf{y}} \left| \mathbf{y}_p - \mathbf{\hat{y}}_p \right|
\end{equation}
It takes the absolute differences between every pixel of both images and returns the mean value.
The second well-known pixel loss function is the Mean Squared Error (MSE), or $L2$-loss. 
It weights high-value differences higher than low-value differences due to an additional square operation:
\begin{equation}
    \label{eq:l2}
    \mathcal{L}_{\text{L2}} \left( \mathbf{y}, \mathbf{\hat{y}}\right) =  \frac{1}{N_\mathbf{y}} \sum_{p \in \Omega_\mathbf{y}} \left| \mathbf{y}_p - \mathbf{\hat{y}}_p \right| ^2
\end{equation}
However, it is more common to see the $L1$-loss in literature because $L2$ is very reactive to extraordinary values: too smooth for low values and too variable for high values \cite{hui2018fast, lim2017enhanced}.
There are variants in the literature, depending on the task and other specifications. 
A popular one is the Charbonnier-loss \cite{lai2017deep}, which is defined by
\begin{equation}
    \label{eq:charbonnier}
    \mathcal{L}_\text{Charbonnier} \left( \mathbf{y}, \mathbf{\hat{y}}\right) =  \frac{1}{N_\mathbf{y}} \sum_{p \in \Omega_\mathbf{y}} \sqrt{\left| \mathbf{y}_p - \mathbf{\hat{y}}_p \right| ^2 + \epsilon^2},
\end{equation}
where $0 < \epsilon \ll 1$ is a small constant such that the inner term is non-zero. 
\autoref{eq:l1} can be seen as a special case of Charbonnier with $\epsilon = 0$. 

Pixel loss functions favor a high PSNR because both formulations use pixel differences. 
Also, PSNR does not correlate well with subjectively perceived quality (see \autoref{sec:psnr}), which makes pixel loss functions sub-optimal. 
Another observation is that the resultant images tend to be blurry: sharp edges need to be modeled better.
One way to combat this is by adding uncertainty.

\subsubsection{Uncertainty-Driven Loss}
Modeling uncertainty in DL improves the performance and robustness of deep networks \cite{kendall2017uncertainties}. 
Therefore, Ning et al. proposed an adaptive weighted loss for SISR \cite{ning2021uncertainty}, which aims at prioritizing texture and edge pixels that are visually more significant than pixels in smooth regions. Thus, the adaptive weighted loss treats every pixel unequally. 

Inspired by insights from Variational Autoencoders (VAE) \cite{kingma2013auto}, they first model an estimated uncertainty.
Let $\mathcal{M}$ be the SR model with parameters $\theta$, which learns two intermediate results: $\mu_\theta \left( \mathbf{x} \right)$, the mean image, and $\sigma_\theta \left( \mathbf{x} \right)$, the variance image (uncertainty).
Consequently, the approximated image $\mathbf{\hat{y}}$ is given by 
\begin{equation}
    \mathbf{\hat{y}} = \mathcal{M} \left( \mathbf{x} ; \theta \right) = \underbrace{\mu_\theta \left( \mathbf{x} \right)}_{= \mathbf{\hat{y}}_\mu} + \epsilon \cdot \underbrace{\sigma_\theta \left( \mathbf{x} \right)}_{= \mathbf{\hat{y}}_\sigma},
\end{equation}
where $\epsilon \sim \mathcal{N} \left( \mathbf{0}, \mathbf{I} \right)$.
Most DL-based SISR methods estimate only the mean $\mathbf{\hat{y}}_\mu$. 
In contrast, Ning et al. proposed estimating the uncertainty $\mathbf{\hat{y}}_\sigma$ simultaneously \cite{ning2021uncertainty}.

Based on the observation that the uncertainty is sparse for SISR due to many smooth areas, Ning et al. present an Estimating Sparse Uncertainty (ESU) loss:
\begin{equation}
    \label{eq:lesu}
    \mathcal{L}_\text{ESU} \left( \mathbf{y}, \mathbf{\hat{y}}\right) = \exp \left( - \ln \mathbf{\hat{y}}_\sigma\right) \cdot \left\lVert \mathbf{y} - \mathbf{\hat{y}}\right\rVert_1 + 2 \cdot \ln \mathbf{\hat{y}}_\sigma
\end{equation}

However, they observed that $\mathcal{L}_{ESU}$ lowers the performance and assumed that $\mathcal{L}_{ESU}$ is unsuitable for SISR. 
They also concluded that prioritizing pixels with high uncertainty is necessary to benefit from uncertainty estimation.

As a result, they proposed an adaptive weighted loss named Uncertainty-Driven Loss (UDL) and used a monotonically increasing function instead of $\exp \left( - \ln \mathbf{\hat{y}}_\sigma\right)$ in \autoref{eq:lesu}:
\begin{equation}
\label{eq:ludl}
    \mathcal{L}_\text{UDL} \left( \mathbf{y}, \mathbf{\hat{y}}\right) =  \left[ \ln \mathbf{\hat{y}}_\sigma - \min \left( \ln \mathbf{\hat{y}}_\sigma\right)\right] \cdot \left\lVert \mathbf{y} - \mathbf{\hat{y}}_\mu\right\rVert_1,
\end{equation}
where $\left[ \ln \mathbf{\hat{y}}_\sigma - \min \left( \ln \mathbf{\hat{y}}_\sigma\right)\right]$ is a non-negative linear scaling function.
There are two variables that need to be learned: $\mathbf{\hat{y}}_\mu$ and $\mathbf{\hat{y}}_\sigma$. 
Ning et al. propose to use $\mathcal{L}_{ESU}$ to learn $\mathbf{\hat{y}}_\sigma$ and then to train a new network with $\mathcal{L}_{UDL}$ to learn $\mathbf{\hat{y}}_\mu$, but with the pre-trained $\mathbf{\hat{y}}_\sigma$ fixed \cite{ning2021uncertainty}.
The idea of using two train processes is to prevent the variance image from degenerating into zeros.
With this setup, they achieve better results than $L1$ and $L2$ for a wide range of network architectures \cite{Lim_2017_CVPR_Workshops, dong2018denoising}. 

However, it requires a two-step training procedure, which adds extra training time.
To this day, we have yet to find any techniques that circumvent such a two-step training process.
The supplementary material contains additional visualizations.

\subsubsection{Content Loss}
\label{sec:contentLoss}
Instead of using the difference between the approximated and the ground-truth image, one can transform both entities further into a more discriminant domain.
Thus, the resultant loss function utilizes feature maps from an external feature extractor.
Such an approach is similar to TBE (see \autoref{sec:tbe}).
In more detail, the feature extractor is pre-trained on another task, i.e., image classification or segmentation.
During the training of the actual SR model on the difference of feature maps, the parameters of the feature extractor remain fixed.
Thus, the goal of the SR model is not to generate pixel-perfect estimations. Instead, it produces images whose features are close to the features of the target. 

More formally, let $\varphi$ be a Convolutional Neural Network (CNN) that extracts features like VGG \cite{simonyan2014very} and let $\varphi^l$ be the $l$-th feature map.
The content loss describes the difference of feature maps generated from the HR image $\varphi^l \left( \mathbf{y} \right)$ and the approximation of it, $\varphi^l \left( \mathbf{\hat{y}} \right)$, and is defined as
\begin{equation}
    \mathcal{L}_\text{Content} \left( \mathbf{y}, \mathbf{\hat{y}}, l\right) = \left\lVert \varphi^l \left( \mathbf{\hat{y}} \right) - \varphi^l \left(  \mathbf{y} \right) \right\rVert_2
\end{equation}

The supplementary material provides a visualization of this approach. 
The motivation is to incorporate image features (content) instead of pixel-level details; a strategy that has been frequently used for Generative Adversarial Networks (GANs), e.g., SRGAN \cite{ledig2017photo}.

\subsection{Generative Adversarial Networks}
Since the early days of GANs \cite{goodfellow2014generative}, they have had a variety of applications in CV tasks e.g., in text-to-image synthesis \cite{frolov2021adversarial}. 
The core idea is to use two distinct networks: a generator $G$ and a discriminator $D$. 
The generator network learns to produce samples close to a given dataset and to fool the discriminator.

In the case of SR, the generator is the SR model ($G = \mathcal{M}$). 
The discriminator tries to distinguish between samples coming from the generator and samples from the actual dataset.
The interaction between the generator and discriminator corresponds to a minimax two-player game and is optimized using the adversarial loss \cite{ledig2017photo, sajjadi2017enhancenet}. 
Given the ground-truth HR image $\mathbf{y}$ of the LR image $\mathbf{x}$ and the approximated SR image $\mathbf{\hat{y}} = G_{\theta_G} \left( \mathbf{x} \right)$, the loss functions based on Cross-Entropy (CE) are:
\begin{equation}
    \label{eq:gan_g_ce}
    \mathcal{L}^{\text{CE}}_{G} =  - \mathbb{E}_{\mathbf{x} \sim \mathbb{P}_\text{LR} \left( \mathbf{x} \right)} \log D_{\theta_D} \left[ G_{\theta_G} \left( \mathbf{x} \right) \right]
\end{equation}
\begin{equation}
\begin{aligned}
    \label{eq:gan_d_ce}
    \mathcal{L}^{\text{CE}}_{D} =  & - \mathbb{E}_{\mathbf{y} \sim \mathbb{P}_\text{HR}\left( \mathbf{y} \right)} \left[ \log D_{\theta_D} \left( \mathbf{y} \right)  \right] \\& - \mathbb{E}_{\mathbf{x} \sim \mathbb{P}_\text{LR} \left( \mathbf{x} \right)} \left[\log \left( 1 - D_{\theta_D} \left[ G_{\theta_G} \left( \mathbf{x} \right) \right] \right) \right]
\end{aligned}
\end{equation}

An alternative is to use the Least Square (LS) \cite{yuan2018unsupervised, wang2018fully}, which yields better quality and training stability:
\begin{equation}
    \label{eq:gan_g_ls}
    \mathcal{L}^{\text{LS}}_{G} = \mathbb{E}_{\mathbf{x} \sim \mathbb{P}_\text{LR}\left( \mathbf{x} \right)} \left( D_{\theta_D} \left[ G_{\theta_G} \left( \mathbf{x} \right) \right] - 1 \right)^2
\end{equation}
\begin{equation}
\begin{aligned}
    \label{eq:gan_d_ls}
    \mathcal{L}^{\text{LS}}_{D} = & \mathbb{E}_{\mathbf{x} \sim \mathbb{P}_\text{LR}\left( \mathbf{x} \right)}\left( D_{\theta_D} \left[ G_{\theta_G} \left( \mathbf{x} \right) \right] \right)^2 \\ & + \mathbb{E}_{\mathbf{y} \sim \mathbb{P}_\text{HR}\left( \mathbf{y} \right)}\left( D_{\theta_D} \left( \mathbf{y} \right) - 1 \right)^2
\end{aligned}
\end{equation}

Famously known for the GAN research field, but less explored for SR \cite{yu2018single}, is the Wasserstein loss with Gradient Penalty (WGAN-GP) \cite{gulrajani2017improved}, which is the extended version of the Wasserstein loss (WGAN). 
Despite variations, Lucic et al. have shown that most adversarial losses can reach comparable scores with enough hyper-parameter tuning and random restarts \cite{lucic2017gans}. 
However, this statement remains unproven for SR and requires more rigorous confirmation in the future. 

A recent review by Singla et al. \cite{singla2022review} examines GANs for SR in more detail.
Generally, SR models perform better if an adversarial loss is incorporated.
The primary disadvantage is that sufficient training stability is hard to reach due to GAN-specific issues like mode collapse \cite{ledig2017photo}. 
One way to improve the training stability is to introduce regularization terms.

\subsubsection{Total Variation Loss}
\label{sec:tvLoss}
One way to regularize GANs is to use a Total Variation (TV) denoising technique known from image processing. First introduced by Rudin et al. \cite{rudin1992nonlinear}, it filters noise by reducing the TV of a given signal.
The TV loss \cite{vella2019single} measures the difference of neighboring pixels in the vertical and horizontal direction in one image. 
It is defined as
\begin{equation}
    \text{TV} \left( \mathbf{y} \right) = \frac{1}{N_\mathbf{y}} \sum_{i,j,k} \sqrt{
    \underbrace{\left( \mathbf{y}_{i+1,j,k} - \mathbf{y}_{i,j,k} \right) ^2}_\text{diff. first axis} + \underbrace{\left( \mathbf{y}_{i,j+1,k} - \mathbf{y}_{i,j,k} \right)^2}_\text{diff. second axis}}
\end{equation}
and minimizing it results in images with smooth instead of sharp edges.
This term helps to stabilize the training of GANs \cite{ledig2017photo, yuan2018unsupervised}.
However, it can hurt the overall image quality since sharp edges are essential. 

\subsubsection{Texture Loss}
\label{sec:textureLoss}
Texture synthesis with parametric texture models has a long history with the goal of transferring global texture onto other images \cite{portilla2000parametric}.
Due to the advent of DL, Gatys et al. \cite{gatys2015texture} proposed a style transfer method (e.g., painting style) that utilized a pre-trained neural network $\varphi$.
Based on that, EnhanceNet \cite{sajjadi2017enhancenet} used the texture loss to enforce textural similarity.
It uses the Gram Matrix to capture correlations of different channels between feature maps:
\begin{equation}
    \mathcal{G}^l_{\left(c_1,c_2\right)} \left( \mathbf{y} \right)= 
    \sum_{\left(i,j,. \right) \in \Omega_\mathbf{y}} \varphi^l \left( \mathbf{y} \right)_{\left(i,j, c_1\right)} \cdot \varphi^l \left( \mathbf{y} \right)_{\left(i,j, c_2\right)}
\end{equation}
where $\varphi^l$ is the $l$-th feature map. The texture loss is 
\begin{equation}
    \mathcal{L}_\text{TEX} \left( \mathbf{y}, \mathbf{\hat{y}}, l\right) =
    \left\lVert \mathcal{G}^l \left( \mathbf{\hat{y}} \right) - \mathcal{G}^l \left( \mathbf{y} \right) \right\rVert_2^2
\end{equation}

\subsection{Denoising Diffusion Probabilistic Models}
Interestingly, DL methods are well suited for denoising Gaussian noise.
Denoising Diffusion Probabilistic Models (DDPMs) \cite{ho2020denoising} exploit this insight by formulating a Markov chain to alter one image into a noise distribution gradually, and the other way around.
The idea of this approach is that estimating small perturbations is more tractable for neural networks than explicitly describing the whole distribution with a single, non-analytically-normalizable function.

\begin{figure}
    \begin{center}
        \includegraphics[width=.49\textwidth]{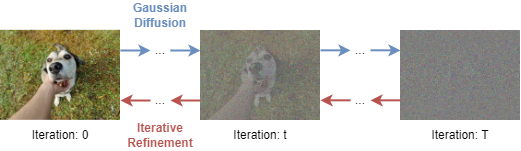}
        
        \caption{\label{fig:ddpms}
        Principle of DDPMs. The Gaussian diffusion process adds noise iteratively. The iterative refinement process reverts the process. The task of the SR model is to predict the noise added between two iterations. The predicted noise is then used to revert one iteration. 
        }
    \end{center}
\end{figure}

First in SR was ``Super-Resolution via Repeated Refinement'' (SR3) \cite{saharia2021image}. It adds noise to the LR image until $\mathbf{y}_T \sim \mathcal{N} \left( \mathbf{0}, \mathbf{I} \right)$ and generates a target HR image $\mathbf{y}_0$ iteratively in $T$ refinement steps. 
While adding noise is straightforward, it uses a DL-based model that transforms a standard normal distribution into an empirical data distribution by reverting the noise-adding process as shown in \autoref{fig:ddpms}.

The diffusion process $q$ adds Gaussian noise to a LR image $\mathbf{y} = \mathbf{y}_0$ over $T$ iterations with
\begin{equation}
    \label{eq_ddpmyty0}
    q \left( \mathbf{y}_t | \mathbf{y}_0 \right) = \mathcal{N} \left( \mathbf{y}_t | \sqrt{\gamma_t} \cdot \mathbf{y}_0, \left(1- \gamma_t \right) \cdot \mathbf{I}\right),
\end{equation}
where $\gamma_t = \prod^T_{i=1}\alpha_i$ and $0 < \alpha_t < 1$ are hyper-parameters to determine the variance of added noise per iteration. 

Consequently, the noisy image $\widetilde{\mathbf{y}}$ can be expressed as 
\begin{equation}
    \label{eq_sr3yt}
    \widetilde{\mathbf{y}} = \underbrace{\sqrt{\gamma} \cdot \mathbf{y}_0}_\text{mean}  + \underbrace{\sqrt{1-\gamma} \cdot \epsilon}_\text{variance}, \quad \epsilon \sim \mathcal{N} \left( \mathbf{0}, \mathbf{I} \right) 
\end{equation}
Given $\gamma$ and $\widetilde{\mathbf{y}}$, one can derive $\mathbf{y}_0$ from $\epsilon$ and vice versa by rearranging \autoref{eq_sr3yt}. 
Thus, a denoising model $\varphi_\theta$ with parameters $\theta$ has to either predict $\mathbf{y}_0$ or $\epsilon$. 

SR3 applies the denoising model to predict the noise $\epsilon$.
The model $\varphi_\theta \left( \mathbf{x}, \widetilde{\mathbf{y}}, \gamma \right)$ takes as input the LR image $\mathbf{x}$, the variance of the noise $\gamma$, and the noisy target image $\widetilde{\mathbf{y}}$ and is optimized by the following loss function
\begin{equation}
    \mathcal{L}_{\text{SR3}} \left( \mathbf{x}, \mathbf{y}_0 \right) = \mathbb{E}_{\epsilon, \gamma}
    \left\lVert \varphi_\theta \left( \mathbf{x}, \underbrace{\sqrt{\gamma} \cdot \mathbf{y}_0  + \sqrt{1-\gamma} \cdot \epsilon}_{= \widetilde{\mathbf{y}} \text{ by \autoref{eq_sr3yt}}}, \gamma \right) - \epsilon \right\rVert_d^d,
\end{equation}
where $d\in \{1, 2\}$.
The alternative regression target $\mathbf{y}_0$ remains open for future research.

The refinement process $p$ is the reverse direction of the diffusion process. 
Given the prediction of $\epsilon_t$ by $\varphi_\theta$, \autoref{eq_sr3yt} can be reformulated to approximate $\mathbf{y}_0$:
\begin{equation}
\begin{split}
    & \mathbf{y}_t = \sqrt{\gamma_t} \cdot \mathbf{\hat{y}}_0  + \sqrt{1-\gamma_t} \cdot \varphi_\theta \left( \mathbf{x}, \mathbf{y}_t, \gamma_t \right) \\
    \iff & \mathbf{\hat{y}}_0 = \frac{1}{\sqrt{\gamma_t}} \cdot \left( \mathbf{y}_t - \sqrt{1-\gamma_t} \cdot \varphi_\theta \left( \mathbf{x}, \mathbf{y}_t, \gamma_t \right) \right)
\end{split}
\end{equation}
Substituting $\mathbf{\hat{y}}_0$ into the posterior distribution to parameterize the mean of $p_\theta \left( \mathbf{y}_{t-1} | \mathbf{y}_t, \mathbf{x}\right)$, details shown in supplementary material, yields 
\begin{equation}
\label{eq:sr3_parameterizedMean}
      \mu_\theta \left( \mathbf{x}, \mathbf{y}_t, \gamma_t \right)
      =  \frac{1}{\sqrt{\alpha_t}} \left[ \mathbf{y}_t - \frac{1 - \alpha_t}{\sqrt{1 - \gamma_t}} \cdot \varphi_\theta \left( \mathbf{x}, \mathbf{y}_t, \gamma_t \right)   \right]
\end{equation}
The authors of SR3 set the variance of $p_\theta \left( \mathbf{y}_{t-1} | \mathbf{y}_t, \mathbf{x}\right)$ to $\left(1-\alpha_t\right)$ for the sake of simplicity. 
As a result, each refinement step with $\epsilon_t \sim \mathcal{N} \left( \mathbf{0}, \mathbf{I} \right)$ is realized as
\begin{equation}
    \mathbf{y}_{t-1} \leftarrow \frac{1}{\sqrt{\alpha_t}} \left[ \mathbf{y}_t - \frac{1 - \alpha_t}{\sqrt{1 - \gamma_t}} \cdot \varphi_\theta \left( \mathbf{x}, \mathbf{y}_t, \gamma_t \right)   \right] + \sqrt{1-\alpha_t} \cdot \epsilon_t
\end{equation}

And indeed, SR3 generates sharp images with more details than its regression-based counterparts on natural and face images.
The authors also strengthened the results by positively testing the generated images with 50 human subjects.
Thus, the outcome of SR3 is a promising direction for future investigations.
Nevertheless, SR3 seems prone to bias, e.g., applied to a facial dataset; the images result in too smooth skin texture by dropping moles, pimples, and piercings.
More extensive ablation studies could help in establishing the practicality for real-world SR applications.

\section{Upsampling}
\label{sec:upsampling_methods}
A critical aspect is how to increase the spatial size of a given feature map. 
This section gives an overview of upsampling methods, either interpolation-based (nearest-neighbor, bilinear, and bicubic interpolation) or learning-based (transposed convolution, sub-pixel layer, and meta-upscale).
Various visualizations can be found in the supplementary material.

\subsection{Interpolation-based Upsampling}
Many DL-based SR models use image interpolation methods because of their simplicity.
The most known methods are nearest-neighbor, bilinear, and bicubic interpolation. 

Nearest-neighbor interpolation is the most straightforward algorithm because the interpolated value is based on its nearest pixel values. 
Nearest-neighbor is swift since no calculations are needed, which is why it is favorable for SR models \cite{sajjadi2017enhancenet}. 
However, there are no interpolated transitions, which results in blocky artifacts.
 
In contrast, bilinear interpolation bypasses blocky artifacts by producing smoother transitions with linear interpolation on both axes.
It needs a receptive field of $2 \times 2$, making it relatively fast and easily applicable \cite{li2019feedback}. 

However, SR methods typically do not use it because bicubic interpolation delivers much smoother results. But it demands much more computation time due to a receptive field size of $4 \times 4$ and makes it the slowest approach among all three methods.
Nonetheless, it is generally used by SR models if an interpolation-based upsampling is applied \cite{dong2015image, tai2017image, lai2017deep} since the overall time consumption is unessential for GPU-based models, which are common in SR research.

\subsection{Learning-based Upsampling}
Learning-based upsampling introduces modules that upsample a given feature map within a learnable setup.
The most standard learning-based upsampling methods are the transposed convolution and sub-pixel layer. 
Promising alternatives are meta-upscaling, decomposed upsampling, attention-based upsampling, and upsampling via Look-Up Tables, which are the last discussed methods in this section.

\subsubsection{Transposed Convolution}
Transposed convolution expands the spatial size of a given feature map and subsequently applies a convolution operation. 
In general, the expansion is realized by adding zeros between given values. 
However, some approaches differ by first applying nearest-neighbor interpolation and then applying zero-padding, e.g., FSRCNN \cite{dong2016accelerating}. 
The receptive field of the transposed convolution layer can be arbitrary (often set to $3 \times 3$).
The added values depend on the kernel size of the subsequent convolution operation.
This procedure is widely known in literature and is also called deconvolution layer, although it does not apply deconvolution \cite{hui2018fast, dong2015image, tong2017image}.
In practice, transposed convolution layers tend to produce crosshatch artifacts due to zero-padding (further discussed in the supplementary material). 
Also, the upsampled feature values are fixed and redundant.
The sub-pixel layer was proposed to circumvent this problem.
    
\subsubsection{Sub-Pixel Layer}
Introduced with ESPCN \cite{shi2016real}, it uses a convolution layer to extract a deep feature map and rearranges it to return an upsampled output. 
Thus, the expansion is carried out in the channel dimension, which can be more efficient for smaller kernel sizes than transposed convolution.
However, assume zero values are used instead of nearest-neighbor interpolation in the transposed convolution. In that case, a transposed convolution can be simplified to a sub-pixel layer \cite{shi2016deconvolution}.
Given a scaling factor $s$ and input channel size $c$, it produces a feature map with $s^2 \cdot c$ channels. 
Next, a convolution operation with zero padding is applied, such that the spatial size of the input and the resulting feature map remains the same. 
Finally, it reshapes the feature map to produce a spatially upsampled output. 
The receptive field of the convolutional layer can be arbitrary but is often $1 \times 1$ or $3 \times 3$. 
This procedure is widely known in the literature as pixel shuffle layer \cite{saharia2021image, wang2018fully, lim2017enhanced}.
In practice, the sub-pixel layer produces repeating artifacts that can be difficult to unlearn for deeper layers, which is also further explained in the supplementary material.
    
\subsubsection{Decomposed Upsampling}
An extension to the above approaches is decomposed transposed convolution \cite{wojna2019devil}. 
Using 1D convolutions instead of 2D convolutions reduces the number of operations and parameters for the component $k^2$ to $2 \cdot k$. 
Note that the applied decomposition is not exclusively bound to transposed convolution and can also be used in sub-pixel layers.
    
\subsubsection{Attention-based Upsampling}
Another alternative to transposed convolution is attention-based upsampling \cite{kundu2020attention}. 
It follows the definition of attention-based convolution (or scaled dot product attention) and replaces the 1x1 convolutions with upsampling methods. 
In more detail, it replaces the convolution for the query matrix with bilinear interpolation and the convolution for the key and value matrix with zero-padding upsampling similar to transposed convolution. 
It needs fewer parameters than transposed convolution but a slightly higher number of operations.
Also, it trains considerably slower: a shared issue across attention mechanisms.
    
\subsubsection{Upsampling with Look-Up Tables}
A recently proposed alternative is to use a Look-Up Table (LUT) to upsample \cite{jo2021practical}.
Before generating the LUT, a small-scale SR model is trained to upscale small patches of a LR image to target HR patches.
Subsequently, the LUT is created by saving the results of the trained SR model applied on a uniformly distributed input space. 
It reduces the upsampling runtime to the time necessary for memory access while achieving better quality than bicubic interpolation. 
On the other hand, it requires additional training to create the LUT.
Also, the size of the small patches is crucial since the size of the LUT increases exponentially. 
However, it is an exciting approach that is worth further exploration.
    
\subsubsection{Flexible Upsampling}
Most existing SR methods only consider fixed, integer scale factors $s \in \mathbb{N}$. 
However, most real-world scenarios require flexible zooming, which demands SR methods that can be applied with arbitrary scale factors $s \in \mathbb{R}^+$. 
It raises a significant problem for models that use those learning-based upsampling methods: one has to train and save various models for different scales, which limits the use of such methods for real-world scenarios \cite{shi2016real, zhang2018image}.
In order to overcome this limitation, a meta-upscale module was proposed \cite{hu2019meta}. 
It predicts a set of filters for each position in a feature map that is later applied to a location in a lower-resolution feature map.
A visualization of the upscale module and the aforementioned steps can be found in the supplementary material.
This approach is exciting for tasks that require arbitrary magnification levels, e.g., zooming. 
Nevertheless, its downsides unfold if huge scaling factors are required because it has to predict the filter weights $W \left( i,j \right)$ independently for each position.
There are only a few works \cite{peng2020saint} that involve this approach in the literature. 
Besides existing SR methods, one can treat each image as a continuous function and generate patches at a fixed pixel resolution around a center like AnyRes-GAN (2022) \cite{chai2022any}, which could be an interesting path to follow for future SR research. 

\section{Attention Mechanisms for SR}
\label{sec:attentionMechanisms}
Attention revolutionized Natural Language Processing (NLP) and plays an essential role for DL in several applications \cite{vaswani2017attention}.
This section presents how SR methods employ attention. 
There are two categories: Channel and spatial attention, which can be applied simultaneously.

\subsection{Channel-Attention}
\begin{figure}
    \begin{center}
        \includegraphics[width=.27\textwidth]{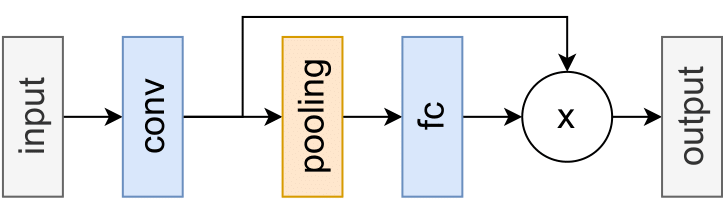}
        \caption{\label{fig:ChannelAttention}
        Channel-attention mechanism \cite{hu2018squeeze}. It reduces a feature map in the spatial dimensions and extracts weighting values by using several FC layers that are element-wise multiplied to the initial feature map.
        }
    \end{center}
\end{figure}
Feature maps generated by CNNs are not equally important. 
Therefore, essential channels should be weighted higher than counterpart channels, which is the goal of channel attention.
It focuses on ``which'' (channels) carry crucial details.
Hu et al. \cite{hu2018squeeze} introduced the channel-attention mechanism for CV as an add-on module for any CNN architecture.
\autoref{fig:ChannelAttention} illustrates the principle.
Upon this, the Residual Channel Attention Network (RCAN) \cite{zhang2018image} firstly applied the channel-attention to SR.
They used a second-order feature metric to reduce the spatial size and applied two fully connected layers to extract the importance weighting.
Because of its simple add-on installation, it is still in use for current research \cite{yang2021image}.
One interesting future direction of channel-attention would be to combine it with pruning in order to not only amplify quality but also to save computation time.

\begin{figure}
    \begin{center}
        \includegraphics[width=.49\textwidth]{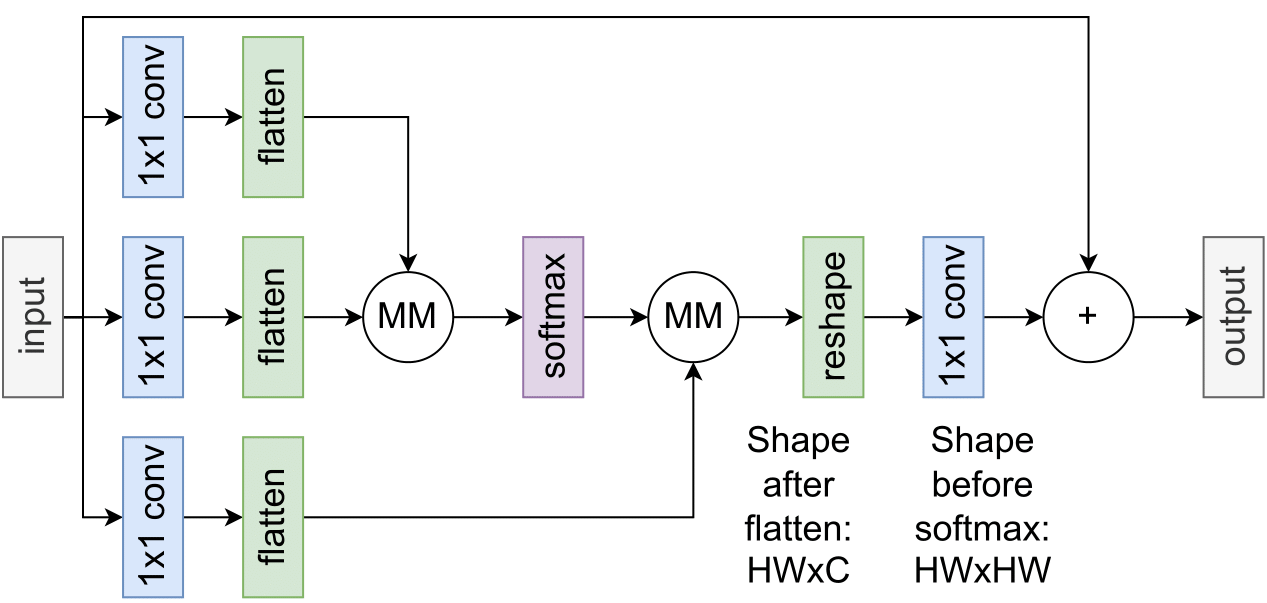}
        \caption{\label{fig:selfAttention}
        Spatial-attention mechanism \cite{wang2018non}. It extracts informations by inspecting the relationship between two positions (first matrix multiplication) and returns the importance of each position as a feature map (second matrix multiplication). MM denotes the matrix multiplication.
        }
    \end{center}
\end{figure}

\subsection{Spatial-Attention}
In contrast to channel attention, spatial attention focuses on ``where'' the input feature maps carry important details, which requires extracting global information from the input.
Therefore, Wang et al. \cite{wang2018non} proposed non-local attention, similar to self-attention for NLP, following the idea of Query-Key-Value-paradigm known from transformers \cite{vaswani2017attention}.

The principle is shown in \autoref{fig:selfAttention}.
Intuitively, the resulting feature map contains each position's weighted sum of features.
It captures spatially separated information by a long range in the feature map.
In SR, the Residual Non-local Attention Network (RNAN) \cite{zhang2019residual} was the first architecture using non-local spatial attention to improve performance.

A known architecture is SwinIR \cite{liang2021swinir}, which reaches state-of-the-art performance and follows the idea of transformers (i.e., multi-head self-attention \cite{vaswani2017attention}) more strictly by applying a sequence of multiple swin transformer layers. SwinFIR \cite{zhang2022swinfir} develops this idea further with fast fourier convolutions. Since transformers originate from NLP, their natural application is restricted to a sequence of patches, and therefore, it requires a patch-wise formulation of an image. This is a significant drawback of transformers applied to SR. SwinIR diminishes the problem by shifting between attention blocks to transfer knowledge between patches.

Exploring its full potential is an exciting topic for the future \cite{dai2019second, zhao2020efficient}. 
However, one major drawback is the difficulty of measuring the importance at a global spatial scale since it often requires vector multiplications (along rows and columns), which is slow for large-scale images.
Bottleneck tricks and patch-wise applications are used to circumvent this problem but are also sub-optimal for importance weighting.
Future research should further optimize computation speed without forfeiting global information.

\subsection{Mixed Attention}
Since both attention types can be applied easily, merging them into one framework is natural. 
Thus, the model focuses on ``which'' (channel) is essential and ``where'' (spatially) to extract the most valuable features.
This combines the benefits of both approaches and introduces an exciting field of research, especially in SR \cite{niu2020single, wang2021bam}.
One potential future direction would be to introduce attention mechanisms incorporating both concerns in one module.

A well-known architecture for mixed attention is the Holistic Attention Network (HAN) \cite{niu2020single}, which uses hierarchical features (from residual blocks) and obtains dependencies between features of different depths while allocating attention weights to them. Thus, it weighs the importance of depth layers in the residual blocks instead of spatial locations within one feature map. Finally, the network uses the layer attention weighted feature map and combines it via addition with channel attention before the final prediction.

\section{Additional Learning Strategies}
\label{sec:additionalStrategies}
Additional learning strategies are blueprints that can be used in addition to regular training.
This section presents the most common methods used in SR, which can immensely impact an SR model's overall performance.

\subsection{Curriculum Learning}
\label{sec:currLearning}
Curriculum learning follows the idea of training a model under easy conditions and gradually involving more complexity \cite{bengio2009curriculum}, i.e., additional scaling sizes. 
For instance, a model trains with a scale of two and gradually higher scaling factors, e.g., ProSR \cite{wang2018fully}. 
ADRSR \cite{bei2018new} concatenates all previously computed HR outputs. 
CARN \cite{ahn2018image} updates the HR image in sequential order, where additional layers extend the network for four times upsampling, which uses previously learned layers, but ignores previously computed HR images \cite{wang2018fully}.
Another way of using curriculum learning is gradually increasing the noise in LR images, e.g., in SRFBN \cite{li2019feedback}.
Curriculum learning can shorten the training time by reducing the difficulty of large scaling factors.

\subsection{Enhanced Predictions}
Instead of enhancing simple input-output pairs, one can use data augmentation techniques like rotation and flipping for final prediction. 
More specifically, create a set of images via data augmentation of one image, e.g., rotation.
Next, let the SR model reconstruct images of the set.
Finally, inverse the data augmentation via transformations and derive a final prediction, i.e., the mean \cite{tong2017image}, or the median \cite{shocher2018zero}.

\subsection{Learned Degradation}
So far, this work assumed learning processes from LR to HR. The Content Adaptive Resampler (CAR) \cite{sun2020learned} introduced a resampler for downscaling. It predicts kernels to produce downscaled images according to its HR input. Next, a SR model takes the LR image and predicts the SR image. Thus, it simultaneously learns the degradation mapping and upsampling task.
The authors demonstrate that the CAR framework trained jointly with SR networks improves state-of-the-art SR performance furthermore.

\subsection{Network Fusion}
Another way of improving performance is to apply multiple SR models. 
Network fusion uses the output of all additional SR models and applies a fusion layer to the outputs. 
Finally, it predicts the SR image used for the learning objective.
For instance, the Context-wise Network Fusion (CNF) \cite{ren2017image} uses this approach with SRCNN, a small SR model introduced in the next chapter.
This method adds a considerable amount of computation and memory, depending on the complexity of the SR models. 
However, it achieves a performance boost and is recommended if computational resources are sufficiently available.
A potential future direction would be to use network fusion to determine critical computation paths and to merge multiple networks into one by pruning or dropout mechanisms.

\subsection{Multi-Task Learning}
Multi-task learning is an exciting research area for SR with the core idea to train a given model to perform various tasks \cite{bei2018new, yuan2018unsupervised}.
E.g., one can assign a label to each image and use multiple datasets for training.
Next, a SR model can learn to reconstruct the SR image and predict its category (e.g., natural or manga image) \cite{urazoe2021multi}.
Another idea is to use datasets that have access to salient image boundaries or image segmentation and use features extracted from the model to predict these \cite{shi2017structure}.
Research on self-learning is an interesting starting point to combine ideas with SR.

\subsection{Normalization Techniques}
A slight change in the input distribution is a cause of many issues because layers need to continuously adapt to new distributions, which is known as covariate shift and can be alleviated with BatchNorm \cite{ioffe2015batch}.
While BatchNorm helps in tasks like image classification, it is frequently used for SR \cite{ledig2017photo, yuan2018unsupervised, tai2017memnet}.
However, BatchNorm does not apply well in SR because it removes network range flexibility by normalizing features \cite{lim2017enhanced}.
The performance of a SR model can be increased substantially without BatchNorm \cite{nah2017deep}.

Recently, the authors of Adaptive Deviation Modulator (AdaDM) \cite{liu2021adadm} studied this phenomenon and were able to show that the standard deviation of residual features shrinks a lot after normalization layers (including BatchNorm, layer, instance, and group normalization), which causes the performance degradation in SR.
The standard deviation resembles the amount of variation of pixel values, which primarily affects edges in images. 
They proposed a module within a feature extraction block to address this problem and amplify the pixel deviation of normalized features. 
It calculates a modulated output $\widetilde{\mathbf{y}}$ by
\begin{equation}
    \widetilde{\mathbf{y}} = \mathbf{y} \cdot \mathrm{e}^{\varphi \left[ \log \left( \sigma \left[ \mathbf{x}\right]\right)\right]},
\end{equation}
where $\mathbf{x}$ is the residually propagated input, $\sigma \left[ \mathbf{x}\right]$ its standard deviation along all three axes, and $\varphi$ a learnable feature extraction module.
The logarithmic scaling guarantees stable training.
As a result, it is possible to apply normalization layers to state-of-the-art SR models and to achieve substantial performance improvements.
One limitation is the additional GPU memory during training, roughly two times the size.
Therefore, lightweight normalization techniques similar to AdaDM, but with less GPU memory consumption are of high interest for future research.

\section{SR Models}
\label{sec:srmodels}
This section presents the most technical part: How to construct an SR model?
In the beginning, we introduce different frameworks that all architectures need to apply, which is the location of the upsampling itself.
After that, it chronologically examines different architecture types. 
Additional visualizations can be found in the supplementary material and an overview benchmark in \autoref{tab:benchmark}.

\subsection{Upsampling Location}
\begin{figure}
\begin{center}
    \includegraphics[width=.49\textwidth]{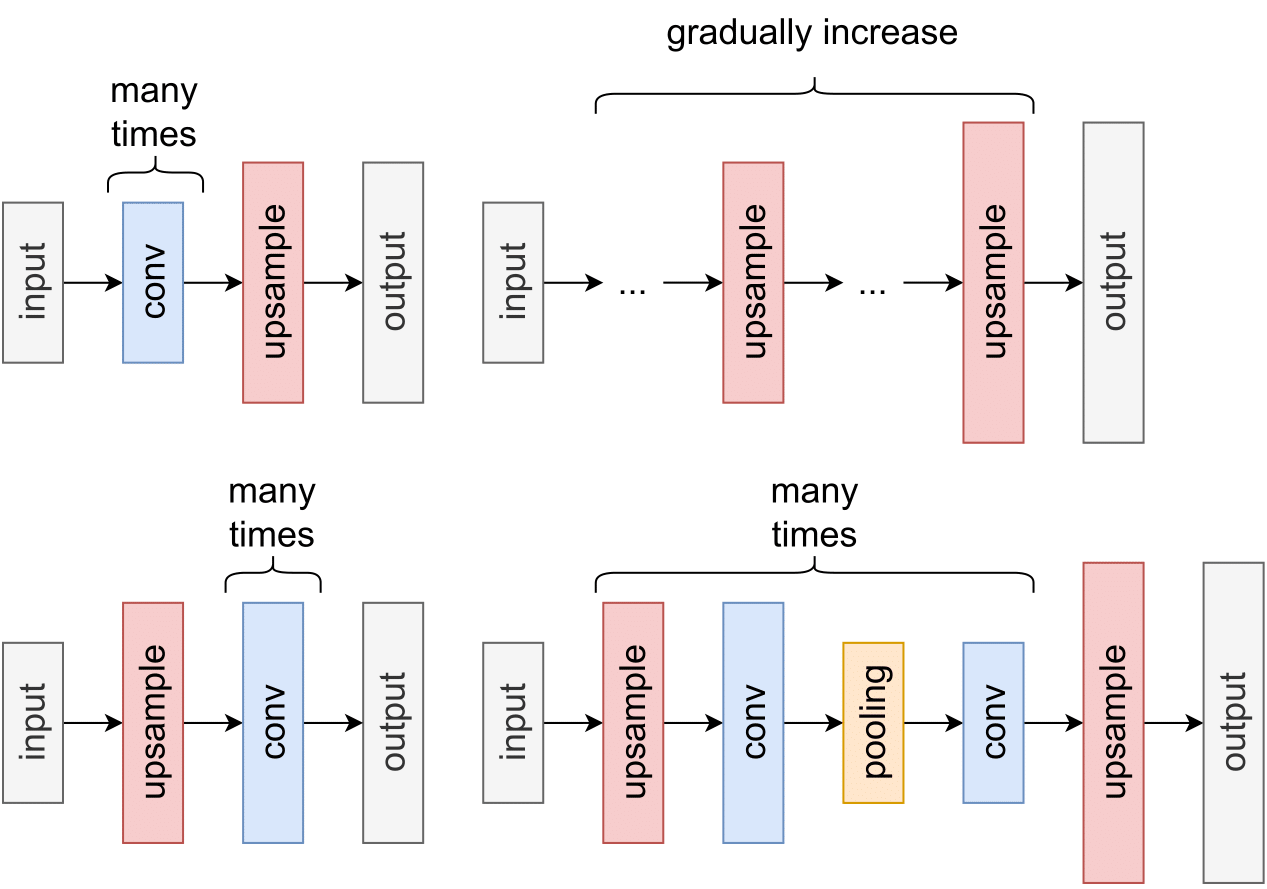}
    \caption{
    Visualization of different upsampling locations within a neural network: (upper left) post-upsampling, (bottom left) pre-upsampling, (upper right) progressive upsampling, and (bottom right) iterative up-and-down upsampling.}
    \label{fig:frameworks}
\end{center}
\end{figure}
Besides the choice of the upsampling method, one crucial decision in designing SR models is where to place the upsampling in the architecture.
There are four variants explored in SR literature \cite{bashir2021comprehensive}: Pre-, post-, progressive, and iterative up-and-down upsampling.
They are shown in \autoref{fig:frameworks} and discussed next.
An additional overview table can be found in the supplementary material.

\textbf{Pre-Upsampling}
\label{sec:preUpsampling}
In a pre-upsampling framework, upsampling is applied in the beginning.
Hereafter, it uses a composition of convolution layers to extract features.
Since the feature maps match the spatial size of the HR image early, the convolutional layers focus on refining the upsampled input.
Thus, it eases learning difficulty but introduces blur and noise amplification in the upsampled image, which impacts the final result.
Also, models trained with pre-upsampling have less performance or application issues than other frameworks since they use DL to refine independently of the scaling in the beginning.
However, it also comes with high memory and computational costs compared to other frameworks like post-upsampling because of the larger input for feature extraction.

\textbf{Post-Upsampling}
\label{sec:postUpsampling}
Post-upsampling lowers memory and computation costs by upsampling at the end, which results in feature extractions in lower dimensional space.
Due to this advantage, this framework decreases the complexity of SR models, e.g., FSRCNN \cite{dong2016accelerating}.
However, the computational burden advances for multi-scaling, where approximations at different scales are required.
A step-wise upsampling during the feed-forward process was proposed to alleviate this issue, the so-called progressive upsampling.

\textbf{Progressive Upsampling}
\label{sec:progressiveUpsampling}
Unlike pre- and post-upsampling, progressive upsampling gradually increases the feature map size within the architecture.
Cascaded CNN-based modules typically enhance to a single scaling factor. Its output is the input for the next module, typically built similarly.
Thus, it segregates the upscaling problem into small tasks, which is a perfect fit for multi-scale SR tasks.
Also, it reduces the learning difficulty for convolutional layers. 
However, the benefit is limited to the highest set scale, and higher scaling factors also require deeper architectures.

\textbf{Iterative Up-and-Down Upsampling}
\label{sec:iterUpAndDown}
Learning the mapping from HR to LR helps to understand the relation from LR to HR \cite{irani1991improving}.
Iterative up-and-down upsampling uses this insight by also downsampling within the architecture.
It is found mainly within recurrent-based network designs \cite{tan2022image, han2018image} and demonstrated significant improvement compared to the previous upsampling techniques.
However, the proper use requires further exploration.
It can be applied with other frameworks, an interesting avenue for future research.

\subsection{Simple Networks}
\label{sec:simpNets}
\begin{figure}
    \includegraphics[width=.49\textwidth]{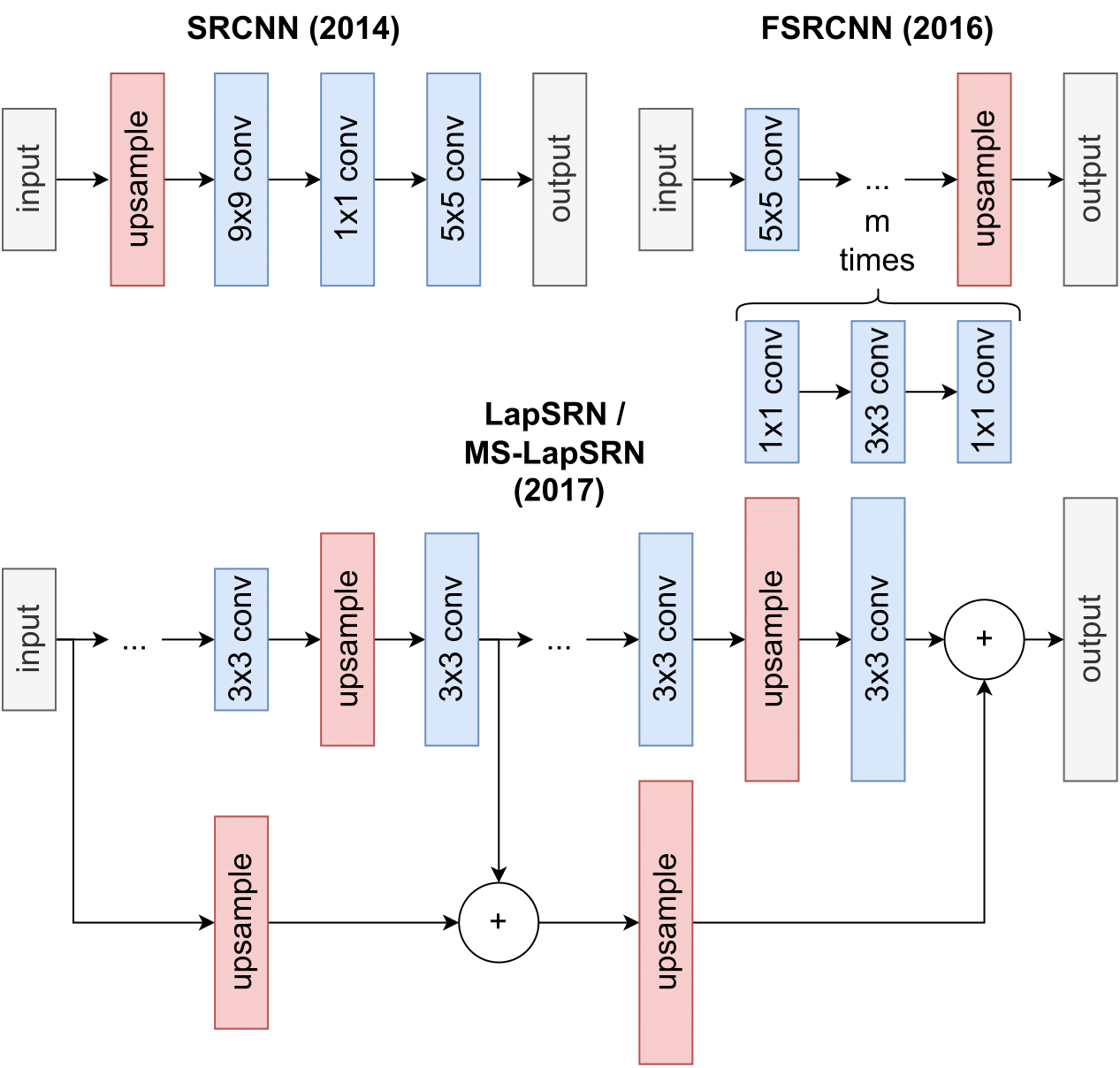}
    \caption{
    Architecture designs of SRCNN \cite{dong2015image}, FSRCNN \cite{dong2016accelerating}, ESPCN \cite{shi2016real} and LapSRN/MS-LapSRN \cite{lai2017deep}. They are grouped together in this work as simple network designs, because they use only convolutional operations and upsampling methods. }
    \label{fig:simple_cnns}
\end{figure}

Simple networks are architectures that mainly apply a chain of convolutions. 
They are easy to understand and typically use a bare minimum of computational resources due to their size. 
Most of these architectures can be found in the early days of DL-based SR because their performances are below state-of-the-art. 
Also, the ``the deeper, the better'' paradigm of DL does not apply well for simple networks because of vanishing/exploding gradients \cite{kim2016accurate}.
\autoref{fig:simple_cnns} shows the different simple network designs. 
The first CNN introduced for SR datasets was SRCNN (2014) by Dong et al. \cite{dong2015image}. 
It uses bicubic pre-upsampling to match the ground-truth spatial size (see \autoref{sec:preUpsampling}).
Subsequently, it consists of three convolution layers, which followed a popular strategy in image restoration: patch extraction, non-linear mapping, and reconstruction.
The authors of SRCNN claimed that applying more layers hurts the performance, which contradicts the DL paradigm ``the deeper, the better'' \cite{he2015convolutional}. 
As seen in the following, this observation was false and required more advanced building blocks to work correctly, e.g., residual connections like in VDSR \cite{kim2016accurate}.
In their follow-up paper, the authors explored various ways to speed up SRCNN, resulting in FSRCNN (2016) \cite{dong2016accelerating} that utilized three major tricks: 
First, they reduced the kernel size of convolution layers. 
Second, they used a 1x1 convolution layer to enhance and reduce the channel dimension before and after a feature processing with 3x3 convolutions. 
Third, they applied post-upsampling with transposed convolution, which is the main reason for the speed up (see \autoref{sec:postUpsampling}).
Surprisingly, they outperformed SRCNN while obtaining faster computation.

A year later, LapSRN (2017) \cite{lai2017deep} was proposed with a key contribution of a Laplacian pyramid structure \cite{ghiasi2016laplacian} that enables progressive upsampling (see \autoref{sec:progressiveUpsampling}). 
It takes coarse-resolution feature maps as input and predicts high-frequency residuals that progressively refine the SR reconstruction at each pyramid level.
To this end, predicting multi-scale images in one feed-forward pass is feasible, thereby facilitating resource-aware applications. 

Simple network architecture designs are found primarily in the early days of DL-based SR because of their limited capacity to learn complex structures due to their size.
Recently, researchers focused on networks with more depth, either with residual networks or synthetic depth with recurrent-based networks. The following sections introduce both possibilities.
\begin{figure}
    \begin{center}
        \includegraphics[width=.303\textwidth]{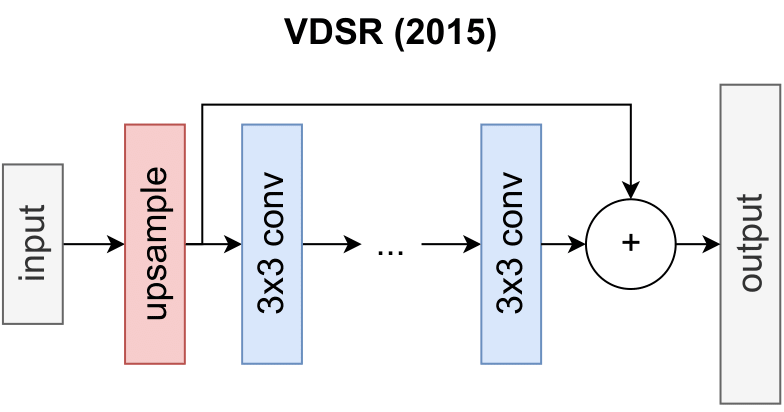}
        \caption{\label{fig:VDSR}
        Architecture design of VDSR \cite{kim2016accurate}. It applies a deep convolutional network as feature extractor but also uses a residual from the input to the final prediction. Therefore, the feature extractor can concentrate to add high-frequency details onto the interpolated image.
        }
    \end{center}
\end{figure}
\subsection{Residual Networks}
Residual networks use skip connections to jump over layers. The primary reason behind adding skip connections is two-fold: To avoid vanishing gradients and mitigate the accuracy saturation problem \cite{he2015convolutional}.
For SR, introducing skip connections unlocked the world of deep-constructed models.
The main advantage is that deep architectures substitute convolutions with large receptive fields, which are crucial for capturing important features.
The authors of SRCNN stated that the ``the deeper, the better'' paradigm does not hold to SR. 
In contrast, Kim et al. refuted this statement with VDSR (2015) \cite{kim2016accurate} and showed that very deep networks could significantly improve SR, visualized in \autoref{fig:VDSR}.

\begin{figure}
    \begin{center}
        \includegraphics[width=.499\textwidth]{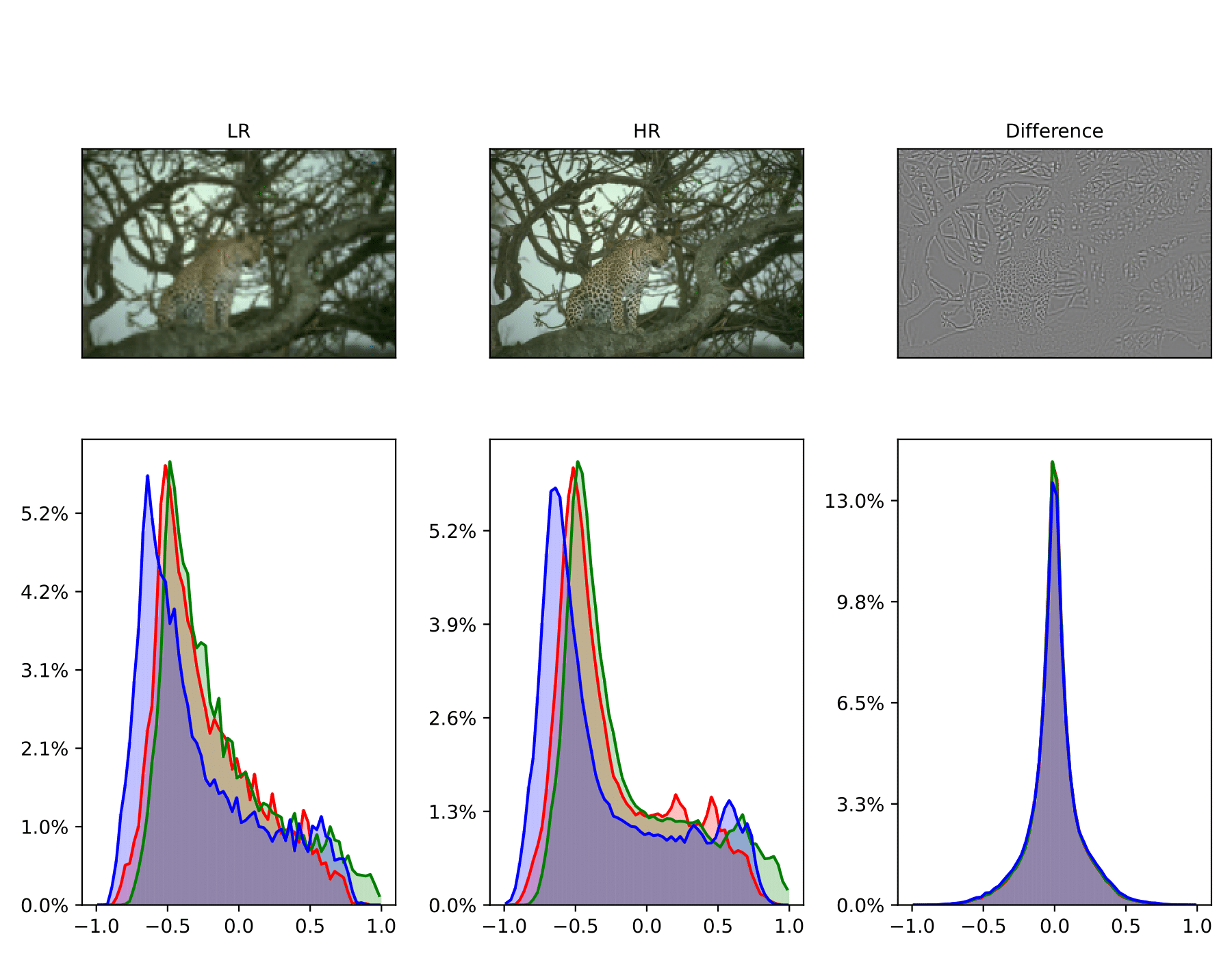}
        \caption{\label{fig:color_dist}
        Pixel value distribution of an example image from BSDS100 \cite{martin2001database}. It can be observed that the distribution of the LR and HR image are skewed. In contrary, the distribution of the difference between LR and HR looks normally distributed, which is easier learned by an SR model.
        }
    \end{center}
\end{figure}

They used two insights from other DL approaches:
First, they applied a famous architecture VGG-19 \cite{simonyan2014very} as a feature extraction block.
Second, they used a residual connection from the interpolation layer to the last layer.
As a result, the VGG-19 feature extraction block adds high-frequency details to the interpolation, resulting in a target distribution that is normal and reduces the learning difficulty immensely, as shown in \autoref{fig:color_dist}. 
Also, it diminishes vanishing/exploding gradients due to the sparse representation of the high-frequency added to the interpolation.
This merit emitted a trend for following residual networks that enhance the number of residuals used.

One example is RED-Net (2016) \cite{mao2016image}, which adapts the U-Net \cite{ronneberger2015u} architecture to SR. 
It incorporates a downsampling encoder and an upsampling decoder network, which downsamples the given output to extract features and then upsamples the feature maps to a target spatial size.
During this process, RED-Net extends the upsample operation with residual information acquired throughout the downsample procedure, which reduces vanishing gradients.
As a result, it outperforms SRCNN for several scaling factors.

Another example adapted ResNet \cite{he2016deep}, which consists of multiple residual units, and DenseNet \cite{huang2017densely}, which sends residual information to all later appearing convolution operations, in the publication of SRGAN (2016) \cite{ledig2017photo}. Moreover, the authors compared these architectures applied with pixel and adversarial loss (see \autoref{sec:learning_objectives}).
SRResNet consists of multiple stacked residual units that allow high-level feature extractions to access low-level feature information through numerous summation operations.
Therefore, it eases optimization by providing an easy back-propagation path to early layers.
Like SRResNet, SRDenseNet \cite{tong2017image} applies dense residual blocks, which utilize even more residual connections to allow direct paths to earlier layers.
In comparison, SRResNet outperforms SRCNN, DRCN, and ESPCN by a large margin.
An extension to SRDenseNet is the Residual Dense Network \cite{zhang2018residual}, proposed in 2018, which incorporates an additional residual connection upon the dense block.

The Densely Residual Laplacian Network (DRLN) \cite{anwar2020densely} is an extension of SRDenseNet and a post-upsampling, channel attention-based residual network and achieves competitive results regarding state-of-the-art. 
Each dense block is followed by a module based on Laplacian pyramid attention, which learns inter and intra-level dependencies between feature maps.
It weights sub-band features progressively in each DRLM, similar to HAN (concatenating various feature maps of different depths).

Another variant of residual blocks is the Information Distillation Network (IDN) \cite{hui2018fast}. 
It employs residual connections to accumulate a portion of a feature map to later layers. 
Given six convolution layers, it divides the feature map into two parts in the middle. Then, one part is processed further by the last three layers and added to the concatenation of the input and the other part.
In brief, networks that utilize residual connections are state-of-the-art. 
Their ability to efficiently propagate information helps fight vanishing/exploding gradients, resulting in excellent performances.
Sometimes, residual block usage is combined with other architectures, such as recurrent-based networks.

\subsection{Recurrent-Based Networks}
\label{sec:recurrentNets}
\begin{figure}
    \begin{center}
        \includegraphics[width=.358\textwidth]{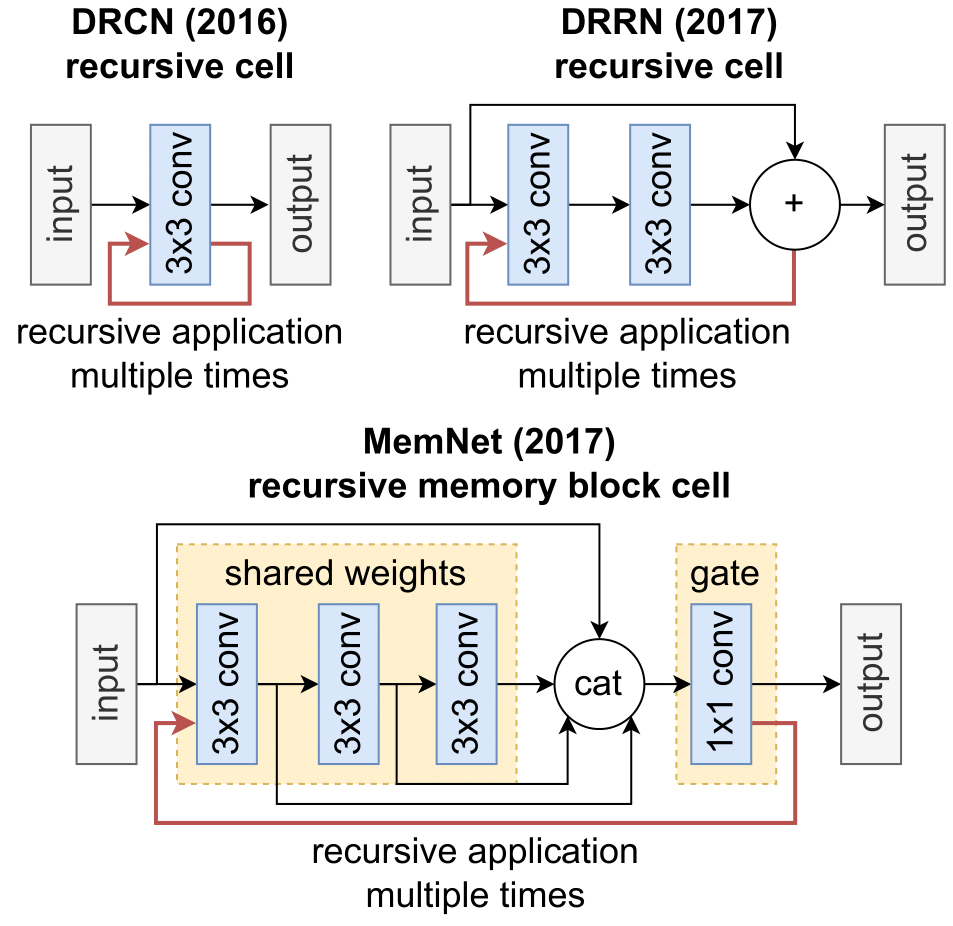}
        \caption{\label{fig:recurrentCNNs}
        Recurrent-based architecture designs of DRCN \cite{kim2016deeply} (upper left), DRRN \cite{tai2017image} (upper right), and MemNet \cite{tai2017memnet} (bottom). The DRCN applies the same convolutional layer multiple times. The DRNN extends this idea to recursive application of a residual block. The MemNet introduces a gate unit to distill most important features from all intermediate convolution results in a recursive manner.
        }
    \end{center}
\end{figure}
Artificial depth can be accomplished with recurrence, where the receptive field, crucial for capturing important information, is enlarged by repeating the same operation.
Also, recurrence reduces the number of parameters, which helps to combat overfitting and memory consumption for small devices. 
It is achieved by applying a convolution layer multiple times without introducing new parameters.

The first recurrent-based network for SR was introduced with DRCN (2015) by Kim et al. \cite{kim2016deeply}. 
It uses the same convolution layer up to 16 times, and a subsequent reconstruction layer considers all recursive outputs for final estimation.
However, they observed that their deeply-recursive network is hard to train but eased it with skip connections and recursive supervision, essentially auxiliary training. \autoref{fig:recurrentCNNs} visualizes DRCN.

Combining the core ideas of DRCN \cite{kim2016deeply} and VDSR \cite{kim2016accurate}, DRRN (2017) \cite{tai2017image} utilizes several stacked residual units in a recursive block structure. 
In addition, it uses 6x and 14x fewer parameters than VDSR and DRCN, respectively, while obtaining better results.
Contrary to DRCN, DRRN shares weight sets among residual units instead of one shared weight for all recursively-applied convolution layers. 
DRNN trains more stable and with deeper recursions than DRCN (52 in total) by emphasizing multi-path. 

Inspired by DRCN, Tai et al. introduced MemNet (2017) \cite{tai2017memnet}. 
The main contribution is a memory block consisting of recursive and gate units to mine persistent memory.
The recursive unit is applied multiple times, similar to DRCN. 
The outputs are concatenated and sent to a gate unit, which is a simple 1x1 convolution layer.
The adaptive gate unit controls the amount of prior information and the current state reserved.
\autoref{fig:recurrentCNNs} visualizes MemNet.
The effect of introducing gates was groundbreaking for sequence-to-sequence tasks (e.g., LSTM \cite{hochreiter1997long}) but deeply understanding its effect on SR tasks marks an open research question.

Inspired by DRRN, the authors of DSRN (2018) \cite{han2018image} explored a dual-state design with a multi-path network.
It introduces two states, one operating in HR and one in LR space, which jointly exploits both LR and HR signals.
The signals are exchanged recurrently between both spaces, LR-to-HR and HR-to-LR, via delayed feedback \cite{chung2015gated}.
From LR-to-HR, it uses a transposed convolution layer to upsample. 
HR-to-LR is performed via strided convolutions.
The final approximation uses the average of all estimations done in the HR space.
Thus, it applies an extended formulation of iterative up-and-down upsampling (see \autoref{sec:iterUpAndDown}).
The two states use more parameters than DRRN, but less than DRCN.
However, exploiting the dual-state design appropriately requires more exploration in the future.

The Super-Resolution Feedback Network (SRFBN, 2019) \cite{li2019feedback}is also using feedback \cite{zamir2017feedback}.
The essential contribution is the feedback block (FB) as an actual recurrent cell. 
The FB uses multiple iterative up-and-downsamplings with dense skip connections to produce high-level discriminant features.
The SRFBN generates a SR image for each iteration, and the FB block receives the previous iteration's output.
It tries to generate the same SR image for single degradation tasks in each iteration. 
For more complex cases, it trains to return better and better quality images with each iteration via curriculum learning (see \autoref{sec:currLearning}).
SRFBN has shown signiﬁcant improvement over the other frameworks but requires more research in the future.

Liu et al. proposed NLRN (2018) \cite{liu2018non} that provides a non-local module to produce feature correlation for self-similarity.
Each position in the image measures the feature correlation of each location in its neighborhood.
NRLN utilizes adjacent recurrent stages between the feature correlation messages.
And indeed, NLRN achieved slightly better performances than DRCN, DRCN, and MemNet.

Nevertheless, primary research on RNNs in SR is lately driven for MISR, such as video SR \cite{isobe2020video} or meta-learning \cite{park2020fast} related tasks.
In general, recurrent-based networks are interesting for saving parameters, but the major drawback is their computational overhead by repeatedly applying the same operation.
Also, they are not parallelizable due to the time dependency.
Alternatives are lightweight architectures, which are introduced next.

\begin{figure}
    \begin{center}
        \includegraphics[width=.49\textwidth]{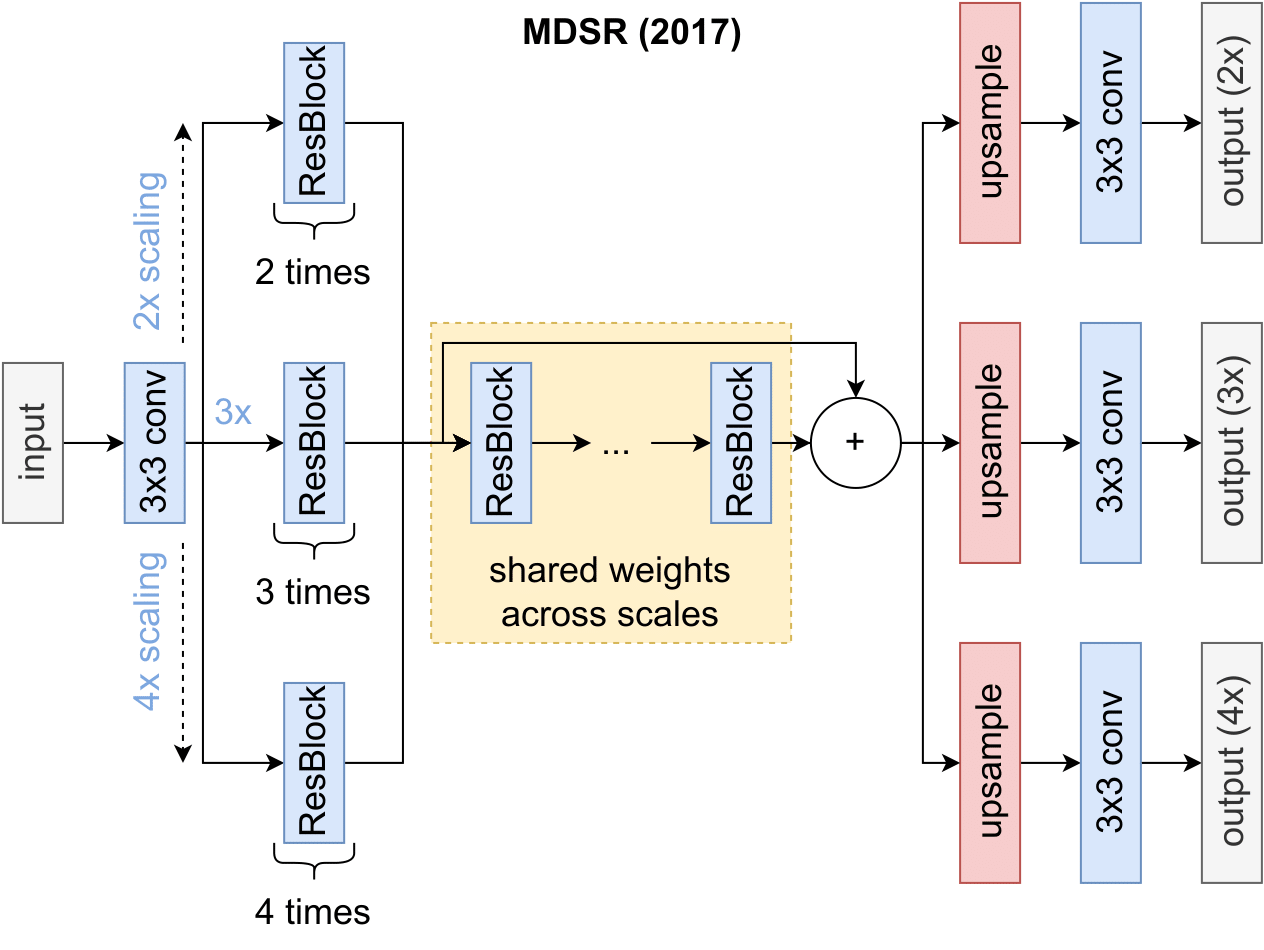}
        \caption{\label{fig:MDSR}
        Architecture design of MDSR \cite{lim2017enhanced}. It utilizes multi-path learning to select between multiple scaling factor paths (denoted by 2x, 3x, 4x on the bottom right of the blue boxes). Also, all paths share an intermediate feature extraction block to save parameters.
        }
    \end{center}
\end{figure}

\subsection{Lightweight Networks}
\label{sec:lightweight}
So far, we have introduced models that increase the quality of SR images and a few that try to do the same, but with less computational effort. 
For instance, FSRCNN \cite{dong2016accelerating} was built to be faster than SRCNN \cite{dong2015image} by utilizing smaller kernel sizes, post-upsampling, and 1x1 convolution layers for enhancing/reducing the channel dimension (see \autoref{sec:simpNets}). 
Another example in this work was recurrent-based networks that reduce redundant parameters as described in \autoref{sec:recurrentNets}. 
The downside of those lean recurrent networks is that parameter reduction comes at the expense of increased operations and inference time, an essential aspect of real-world scenarios.
E.g., SR on mobile devices is limited by battery capacity, which depends on the computation power needed.
Therefore, lightweight architectures explicitly focus on both execution speed and memory usage.
Supplementary materials include a parameter comparison, and a fair comparison of execution speed is in welcome demand.

The MDSR (2017) \cite{lim2017enhanced} uses the multi-path approach to learn multiple scaling factors with shared parameters.
It has three non-identical paths as pre-processing steps and three paths for upsampling. 
For a given scaling factor $s \in \{ 2, 3, 4\}$, MDSR is choosing deterministic between the three paths. 
The paths for larger scales are built deeper than those for lower scaling factors.
Between the pre-processing and upsampling steps is a shared module consisting of multiple residual blocks. 
This feature extraction block is trained and commonly used for all scaling factors, visualized in \autoref{fig:MDSR}.
The main advantage is that one model is sufficient to train on multiple scales, which saves parameters and memory.
In contrast, other SR models must be trained independently on different scales and saved independently for multi-scale applications.
Nevertheless, adding a new scaling factor requires training from scratch.

Other lightweight architectures adapt this idea to enable parameter-efficient multi-scale training, such as CARN/CARN-M (2018) \cite{ahn2018fast}.
\begin{figure}
    \begin{center}
        \includegraphics[width=.49\textwidth]{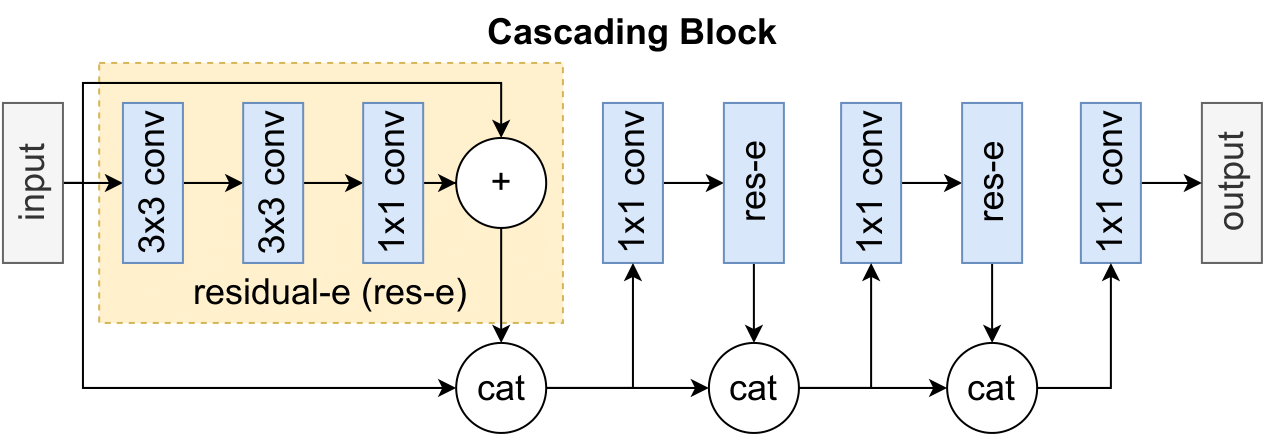}
        \caption{\label{fig:Cascading_Blocks}
        Design of cascading blocks \cite{ahn2018fast}. It is a densely connected block, which consists of Residual-E blocks \cite{ahn2018fast} and 1x1 convolutions that take as input the Residual-E block's output and the residual connections from the layers before. The Residual-E block itself is built like a residual unit but performs group convolution, which is efficient in inference time, depending on the group size.
        }
    \end{center}
\end{figure}
Moreover, it implements a cascading mechanism upon the residual network \cite{he2016deep}. 
CARN consists of multiple cascading blocks (see \autoref{fig:Cascading_Blocks}) and 1x1 convolutions between them. 
The output of the cascading blocks is sent to all subsequent 1x1 convolutions like in the cascading block itself. 
Thus, the local cascading is almost identical to a global one.
It allows multi-level representations and stable training like for residual networks.
Ultimately, it chooses between three paths, which upsample the feature map to 2x, 3x, or 4x scaling factors via efficient sub-pixel layers similar to MDSR. 
Inspired by MobileNet \cite{howard2017mobilenets}, CARN also uses grouped convolution in each residual block component.
This allows configuration of the model's efficiency since choosing different group sizes and the resulting performances are in a trade-off relationship.
The residual blocks with group convolution can reduce the computation up to 14 times, depending on the group size.
They tested a variant of CARN, which sets the group size so that the computation reduction is maximized and called it CARN-Mobile (CARN-M).
Moreover, they further reduced CARN-M's parameters by enabling weight-sharing of their residual blocks within each cascading block (reduction by up to three times compared to non-shared).

\begin{figure}
    \begin{center}
        \includegraphics[width=.404\textwidth]{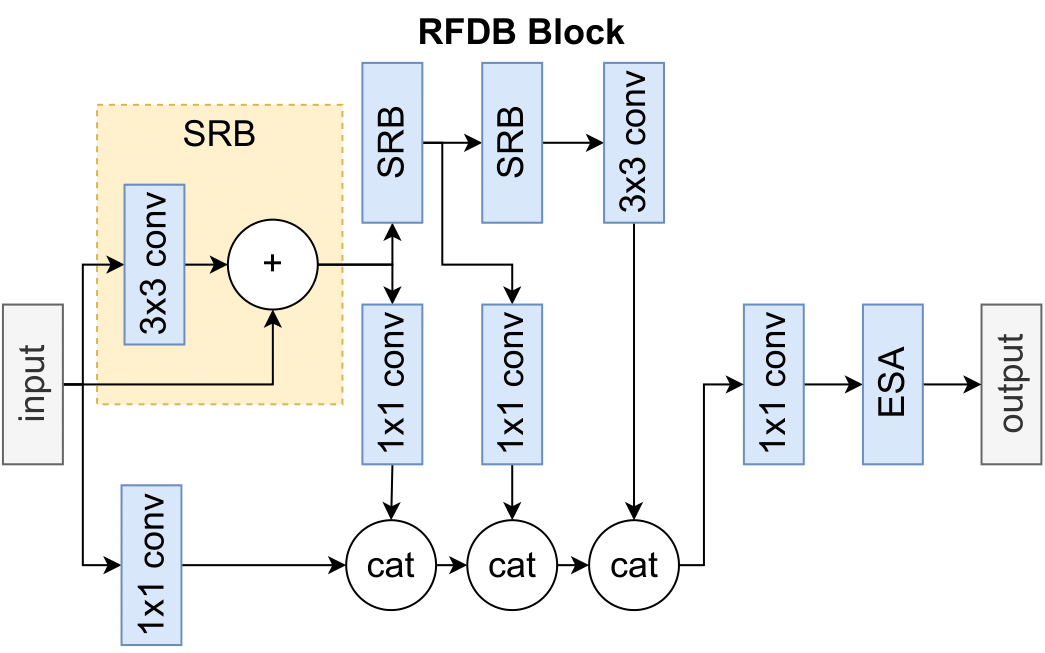}
        \caption{\label{fig:RFDB}
        Visualization of the RFDB block \cite{liu2020residual}. It uses a multi-path approach to distill features with 1x1 convolutions and introduces a shallow representation of the residual block, which consists of only one 3x3 convolution. In the end, all intermediate results are combined with one 1x1 convolution layer and an enhanced spatial attention \cite{liu2020residual} layer.
        }
    \end{center}
\end{figure}
Inspired by IDN and IMDB \cite{hui2019lightweight}, the RFDN (2020) \cite{liu2020residual} rethinks the IMDB architecture by using RFDB blocks as shown in \autoref{fig:RFDB}.
RFDB blocks consist of feature distillation connections, which cascade 1x1 convolutions towards a final layer.
Moreover, it uses shallow residual blocks (SRBs), consisting of only one 3x3 convolution, to process a given input further.
The final layer is a 1x1 convolution layer that combines all intermediate results. 
In the end, it applies the enhanced spatial attention \cite{liu2020residual}, designed specifically for lightweight models.
The RFDN architecture comprises subsequent RFDB blocks and uses a post-upsampling framework with a final sub-pixel layer.

A very hardware-aware and quantization-friendly network is XLSR (2021) \cite{ayazoglu2021extremely}.
It applies multi-paths to ease the burden of the convolution operations and employs 1x1 convolution to combine them pixel-wise. 
Each convolution layer has a small filter size (8, 16, 27). 
After the combination, it splits the feature map and applies multi-paths again. 
A core aspect of XLSR is the end activation layer, which exploits quantization benefits.
Quantization is useful since it can save parameters by using more miniature bit representations \cite{hubara2017quantized}. 
Unfortunately, many mobile devices support 8-bit. Therefore, applying uint8 quantization on SR models that performed well in float32 or float16 does not work. Clipped ReLU (constrained to a max value of 1) as the last activation layer instead of the typical ReLU can dimish this issue.
Nevertheless, the authors recommend searching for other maximum values with further experiments.

Generally, there are plenty of ideas to make SR models lightweight yet to be discovered. 
They include simplifications, quantization, and pruning of existing architectures. 
Also, more resource-constrained devices and applications utilizing SR are a growing field of interest \cite{sarker2021deep}.

\subsection{Wavelet Transform-based Networks}
\begin{figure}
    \begin{center}
        \includegraphics[width=.397\textwidth]{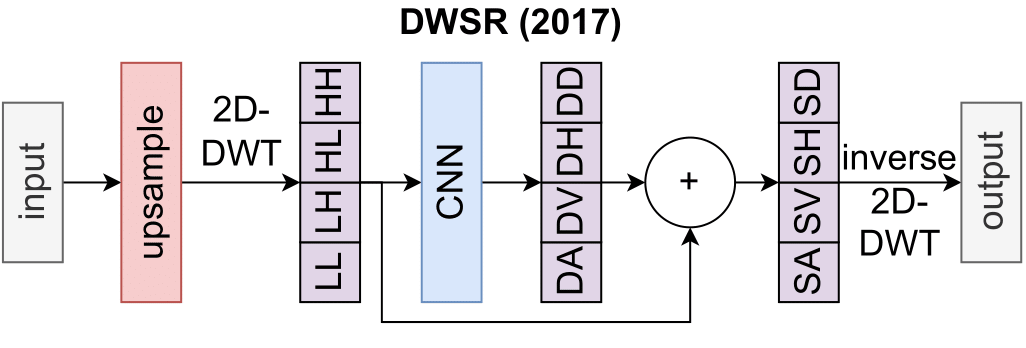}
        \caption{\label{fig:DWSR}
        Visualization of DWSR \cite{guo2017deep}. It calculates wavelet coefficients  and uses a CNN to calculate the difference in the average (DA), vertical (DV), horizontal (DH), and diagonal (DD) sub-bands.
        Next, it adds the initially calculated coefficients via residual connection to the difference to derive the coefficients of the predicted SR image (SA, SV, SH, and SD), which is obtained by using the reverse 2D-DWT.
        }
    \end{center}
\end{figure}

Different representations of images can bring benefits, such as computational speed up.
The wavelet theory gives a stable mathematical foundation to represent and store multi-resolution images, depicting contextual and textural information \cite{stephane1999wavelet}.
Discrete Wavelet Transform (DWT) decomposes an image into a sequence of wavelet coefficients.
The most frequent wavelet in SR is the Haar wavelet, computed via 2D Fast Wavelet Transform.
The wavelet coefficients are calculated by repeating the decomposition to each output coefficient iteratively. 
It captures image details in four sub-bands: average (LL), vertical (HL), horizontal (LH), and diagonal (HH) information. 
One of the first networks that worked with wavelet prediction was DWSR (2017) \cite{guo2017deep}. 
It uses a simple network architecture to refine the differences between the LR and HR image wavelet decompositions in a pre-upsampling framework.
First, it calculates the wavelet coefficients of the enlarged (with bicubic interpolation) LR image. 
Then it processes the wavelet coefficients with convolution layers. 
Next, it adds the initially calculated wavelet coefficients, which are beared with a residual connection. 
Thus, the convolution layers learn additional details of the coefficients.
Finally, it applies the reverse process of 2D-DWT to obtain the SR image, as depicted in \autoref{fig:DWSR}.

Another approach was proposed with WIDN (2019) \cite{sahito2019wavelet}, which uses stationary wavelet transform instead of DWT to perform better.
A more sophisticated model was proposed around the same time as DWSR with Wavelet-SRNet (2017) \cite{huang2017wavelet}.
It provides an embedding network, consisting of residual blocks, to generate feature maps from the LR image. 
Next, it applies DWT several times and utilizes multiple wavelet prediction networks.
Finally, it applies the reverse process and uses a transposed convolution for upsampling.
The coefficients are employed to a wavelet-based loss function, while the SR images are used for a traditional texture and MSE loss function.
As a result, their network applies to different input resolutions with various magnifications and shows robustness toward unknown Gaussian blur, poses, and occlusions for MS-SR.
The idea of multi-level wavelet CNNS can also be found in later publications, i.e., MWCNN (2018) \cite{liu2018multi}.

Following works apply a hybrid approach by mixing wavelet transform with other well-known SR methods. 
I.e., Zhang et al. proposed a wavelet-based SRGAN (2019) framework \cite{zhang2019super}, which merged the advantages of SRGAN and wavelet decomposition.
The generator uses an embedding network to process the input into feature maps, similar to Wavelet-SRNet. 
Next, it uses a wavelet prediction network to refine the coefficients, similar to DWSR.
In 2020, Xue et al. \cite{xue2020wavelet} combined wavelets with residual attention blocks that contain channel attention and spatial attention modules (mixed attention, see the following \autoref{sec:attentionMechanisms}) and called their network WRAN.
Over the last several years, the application of wavelets also found its way to Video SR \cite{zhu2021video}.

In general, wavelet transformations enable an efficient representation of images.
As a result, SR models using this strategy often reduce the overall model size and computational costs while reaching similar performances to state-of-the-art architectures.
However, this research area needs more exploration. 
For example, suitable normalization techniques since the distribution of high-frequency sub-bands and low-frequency sub-bands differ significantly or alternatives to convolution operations since they might be inappropriate due to the sparse representation of the high-frequency sub-bands.

\section{Unsupervised Super-Resolution}
\label{sec:unsupervised}
The astonishing performance of supervised SR is imputed to their ability to learn natural image mainly from many LR-HR image pairs, mostly with known degradation mapping, which is often unknown in practice.
Thus, supervised trained SR models are sometimes unreliable for practicable use-cases. 
For instance, when the training dataset has LR images generated without anti-aliasing (high-frequency is preserved), then the SR model trained on that dataset is not adequately suitable for real-world LR images generated with anti-aliasing (smoothed images).
In addition, some specialized application areas lack LR-HR image pair datasets.
Therefore, there is a growing interest in unsupervised SR.
We examine briefly this field, for further reading inclusive flow-based methods (density estimation of degradation kernels) we refer to the survey of Liu et al. \cite{liu2022blind}.

\subsection{Weakly-Supervised}
Weakly-supervised methods use unpaired LR and HR images like WESPE (2018) \cite{ignatov2018wespe}. 
WESPE consists of two generators and two discriminators.
The first generator takes a LR image and super-resolves it. 
The output of the first generator constitutes a SR image, but is also regularized with TV loss \cite{vella2019single}.
The second generator takes the prediction of the first generator and performs the inverse mapping.
The result of the second generator is optimized via content loss \cite{sajjadi2017enhancenet} with the original input, the LR image.
The two discriminators take the SR image of the first generator and are trained to distinguish between predictions and original HR images.
The first discriminator classifies inputs based on image colors into SR or HR images. 
The second discriminator uses image textures \cite{gatys2015texture} in order to classify.
\label{sec:weaklySR}

A similar approach is a cycle-in-cycle SR framework called CinCGAN (2018) \cite{yuan2018unsupervised}, based on CycleGAN \cite{zhu2017unpaired}.
It uses a total of four generators and two discriminators.
The first generator takes a noisy LR image and maps it to the clean version.
The first discriminator is trained to distinguish between clean LR images from the dataset and the predicted clean images.
The second generator trains the inverse function.
Thus, it generates the noisy image from the predicted clean version, which closes the first cycle of a CycleGan.
The third generator is of particular interest because it is the actual SR model which upsamples the LR image to HR.
The second discriminator is trained to distinguish between the predicted and the dataset's HR images.
The last generator maps the predicted HR image to the noisy LR image, which closes the second cycle of a CycleGAN.
Besides its promising results and similar approaches \cite{bulat2018learn}, it requires further research to decrease learning difficulty and computational cost.
\begin{figure}
    \begin{center}
        \includegraphics[width=.49\textwidth]{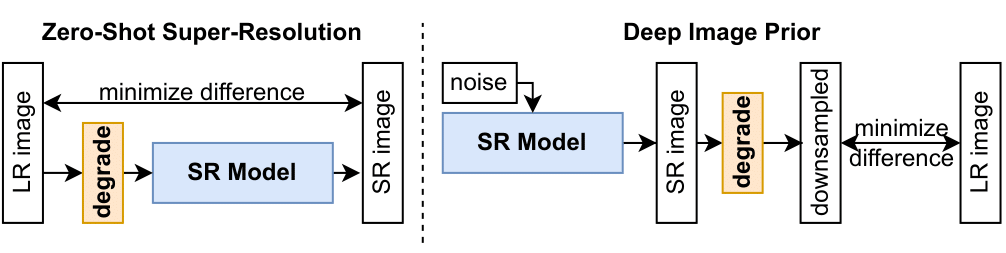}
        \caption{\label{fig:zssrdip}
        Zero-Shot Super-Resolution (ZSSR) \cite{shocher2018zero} and Deep Image Prior (DIP) \cite{ulyanov2018deep}. ZSSR uses the LR image and downsamples it and the SR model learns to reverse the downsampling. For the final prediction of the SR image of the LR image, it is applied to the LR image directly. DIP uses fixed noise as input and predicts the SR image, which is downsampled to optimize the difference between the downsampled image and a given LR image. The final prediction uses the SR model to predict the SR image but skips the degradation mapping.
        }
    \end{center}
\end{figure}
\subsection{Zero-Shot}
Zero-shot or one-shot learning is associated with training on objects and testing on entirely different objects from a different class that was never observed. 
Ideally, a classifier trained on horses should recognize zebras if the knowledge ``zebras look like striped horses'' is transferred \cite{xian2018zero}.
The first publication on Zero-Shot in SR  is ZSSR (2017) \cite{shocher2018zero}.
The goal was to train only on one image at hand, one of a kind.
The degradation mapping for ZSSR was chosen to be fixed, such as bicubic.
ZSSR downsamples the LR image and trains a CNN to reverse the degradation mapping.
The trained CNN is then finally used directly on the LR image.
Surprisingly, this method reached better results than SRCNN and was close to VDSR.

Upon this, a Degradation Simulation Network (DSN, 2020) \cite{cheng2020zero} based on depth information \cite{godard2019digging} was proposed to avoid a pre-defined degradation kernel.
It uses bi-cycle training to simultaneously learn the unknown degradation kernel and the reconstruction of SR images.
The MZSR \cite{soh2020meta} merged the ZSSR setting with meta-learning and used an external dataset to learn different blur kernels, which is called task distribution in the field of meta-learning.
The SR model is then trained on the downsampled image similar to ZSSR with the blur kernel returned from the meta-test phase.
The profit of this approach is that it conducts the SR model to learn specific information faster and performs better than pure ZSSR.
Zero-shot learning for SR marks an exciting area for further research because it is highly practical, especially for applications where application-specific datasets are rare or non-existent.

\subsection{Deep Image Prior}
\label{sec:DIP}
Ulyanov et al. \cite{ulyanov2018deep} proposed Deep Image Prior (DIP), which contradicts the conventional paradigm of training a CNN on large datasets. 
It uses a CNN to predict the LR image when downsampled, given some random noise instead of an actual image.
Therefore, it follows the strategy of ZSSR by using only the LR image.
However, it fixes the input to random noise and applies a fixed downsampling method to the prediction.
Moreover, it optimizes the difference between the downsampled prediction and the LR image.
The CNN then produces the SR image without using the fixed downsampling method.
Thus, it generates an SR image out of noise instead of transforming a raw image.
The difference between ZSSR and DIP is highlighted in \autoref{fig:zssrdip}.
Surprisingly, the results were close to LapSRN \cite{lai2018fast}.
Unfortunately, it is a theoretical publication about image priors, and the approach is too slow to be useful for most practical applications, as the authors stated themselves.
However, it does not exclude future ideas that could enhance the practicability of DIP concerning better image reconstruction quality and especially runtime.

\begin{figure}
    \begin{center}
        \includegraphics[width=.48\textwidth]{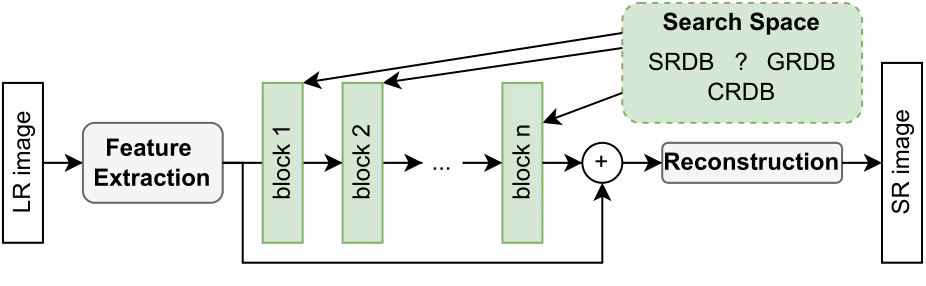}
        \caption{\label{fig:ESRN}
        Visualization of ESRN \cite{song2020efficient}. A controller derives an architecture by selecting from a pre-set of various efficient residual blocks.
        }
    \end{center}
\end{figure}
\section{Neural Architecture Search}
\label{sec:nas}
Recently, a new field called Neural Architecture Search (NAS) has gained popularity, which aims to automatically derive designs instead of hand-designed Neural Networks (NNs) created by human artistry \cite{chu2020multi}.
Fortunately, NAS can be applied to derive parts of SR models.

MoreMNAS (2019) \cite{chu2020multi} investigated NAS for SR and resource-aware mobile devices.
The goal is to derive cell-based designs similar to the idea of residual blocks.
A SR model usually consists of feature extraction, non-linear mapping, and reconstruction, like in SRCNN.
Their search space is constrained to non-linear mapping.
They modified the NAS algorithm NSGANet \cite{lu2019nsga} to SR, which is based on the evolution algorithm NSGA-II with multi-objective evolution \cite{1599245}. In the case of MoreMNAS, multi-objective fitness is defined by high PSNR, low FLOPS, and a few parameters.
Furthermore, they combine it with reinforcement learning, in which an LSTM \cite{hochreiter1997long} controller returns design choices based on NSGA-II, and the evaluator returns a MSE-based reward.
The evaluator trains and evaluates the design choices in a separate training phase.
As a result, MoreMNAS delivers rivaling models compared to methods (like FSRCNN, VDSR, DRRN, CARN-M, and DRCN) with fewer FLOPS.
FALSR (2019) \cite{chu2021fast} is a similar approach, which differs by using a hybrid controller and a cell-based elastic search space that enables both macro (the connections among
different cell blocks) and micro search (cell blocks).
A faster method with comparable performance to FALSR was proposed with ESRN (2019) \cite{song2020efficient}.
They search locations of pooling and upsampling operators and derive the architecture with the guidance of block credits, which weighs the sampling probability of
mutation to favor admirable blocks.
Moreover, they constrain the search space to efficient residual dense blocks from known lightweight architectures, such as RFDB from RFDN (see \autoref{sec:lightweight}).
\autoref{fig:ESRN} shows the latter.
DeCoNAS (2021) \cite{ahn2021neural} is a similar approach, which uses ENAS \cite{pham2018efficient} and densely connected network blocks.

HNAS (2020) \cite{guo2020hierarchical} introduces a hierarchical search space that consists of a cell-level search space and a network-level search space, similar to FALSR.
The cell-level search space identifies a series of blocks that increase model capacity.
The network-level search space determines the position of upsampling layers.
For cell-level search space, HNAS uses two LSTM controllers: one derives normal cells consisting of convolutions, and the other one the upsampling cells (bilinear interpolation, sub-pixel layer, transposed convolution, and more).
On top, a network-level controller is used to select the positions of the upsampling cells.

NAS-DIP (2020) \cite{chen2020dip} combines NAS with DIP \cite{ulyanov2018deep} (see \autoref{sec:DIP} and supplementary material for visualization).
It consists of two phases:
First, they apply reinforcement learning with a RNN controller (PSNR as a reward) to generate a network structure.
The second phase uses the network structure and optimizes the mapping from random noise to a degenerated image (as in DIP \cite{ulyanov2018deep}).
Next, it repeats the phases with the reward gained after the second phase.
Their search space is designed for two components.
The first component is the upsampling cell (e.g., bilinear, bicubic, nearest-neighbor).
The second component is feature extraction (e.g., convolution, depth-wise convolution, and others).
Moreover, they extend their search space to learn cross-level connection patterns across various feature levels in an autoencoder network close to the U-Net of DIP.

Another way of combining NAS with frameworks that worked well in SR was proposed by Lee et al. (2020) \cite{lee2020journey}. 
They combined NAS with GANs and defined search spaces for generators and discriminators. 
The GAN framework is then used to train the generator in the manner of SRGAN.
However, they faced substantial stability problems during their work, which needs further research.

One major drawback of reinforcement learning based NAS is that they need additional training to evaluate design choices. 
It imposes a significant time overhead on the search process.
Another problem is that reinforcement learning based NAS approaches rely on discrete choices performed and exploited by a controller.
One way to omit that is by using a gradient-based search like DARTS \cite{liu2018darts}.
HiNAS (2021) \cite{zhang2020memory} adopts gradient-based search and builds a flexible hierarchical search space. 
The search space is similar to HNAS, which refers to micro and macro search.
During cell search (micro), HiNAS considers all combinations of operations as a weighted sum. 
The highest weights derive the final cell design.
The macro search follows the same idea as micro search and conducts several supercells with different settings and derives the final architecture gradient-based. 
Wu et al. (2021) \cite{wu2021trilevel} proposed something similar. 
They adopt gradient-based search but extend the search space into three levels.
The first level describes the network level, which defines all candidate network paths. 
The second level determines possible candidate operations.
Lastly, the kernel-level is a subset of convolution kernel dimensions.

The main streams of NAS in SR are dominated by evolutionary algorithms paired with reinforcement learning.
They imply a significant search time, which is inevitable due to the repetitive action and reward processes.
Most recently, gradient-based methods were applied to facilitate the search time. 
Moreover, they enable continuous search by relaxing the search space, transforming discrete choices into a weighted sum of possible paths.
The trend points to various hierarchies within the search process, meaning that more aspects of determining a network design for SR are incorporated into NAS.
A well-designed architecture is critical for success on SR tasks. 
NAS approaches have yet to outperform hand-crafted state-of-the-art architectures.
More elaborate methods must be introduced to produce better-performing architectures concerning quality and inference speed.

\section{Discussion and Future Directions}
\label{sec:discussion}
This section gives a short overview of the topics discussed in this paper and shows potential future directions. 

\textbf{IQA}
Evaluating quality of generated images is difficult and still an open problem. PSNR and SSIM are valuable metrics because of fast calculation, but do not always match subjectively perceived quality.  
Developing a metric that overcomes this problem is of high interest. 
Future research should focus on such a metric since it would also be interesting besides SR, e.g., Text-to-Image Synthesis \cite{frolov2021adversarial}.

\textbf{Learning Objectives} The selection of learning objectives strongly depends on the data domain. 
A learning objective that fits all SR domains is not given and remains open for research.
Most SR papers use simple formulations, e.g., regression-based or combinations of them.
Exploring various probable loss functions is highly demanded.
Recent developments in uncertainty-driven loss functions can be a promising direction for new loss functions.

\textbf{Upsampling} One problem with the most commonly used upsampling layers is that they produce artifacts.
Also, the scale factor must be predefined.
Therefore, using them in an application like zooming is not feasible.
An alternative is meta-upscaling, which enables arbitrary scaling but comes with computational overhead and stability issues. 
Thus, new lightweight layers for arbitrary upscaling are highly interesting and should be focused on in the following years.
Moreover, bilateral upsampling and guided upsampling play a crucial role in nowadays applications, which should be explored more in the future concerning deep learning.

\textbf{Unsupervised Super-Resolution}
Data-rich datasets inherently dictate the overall performance of any SR method.
However, LR-HR image pairs are not always given.
In such a case, unsupervised SR methods are interesting.
It marks an exciting area for further research because it is highly practical, especially for applications where datasets are rare or non-existent.
Zero-shot learning and DIP are good starting points, but the execution speed and practicability is not enough for real-world applications.
New ideas are required to enhance practicability regarding better image reconstruction quality and especially execution time.

\textbf{Neural Network Architectures}
Concerning the statement of DIP, a well-designed architecture is critical for success on SR tasks.
Recently, many architectures were proposed to investigate certain aspects, such as SRResNet \cite{zhang2019super} (handling vanishing/exploding gradients), DSRN \cite{han2018image} (exploring recurrence for SR), XLSR \cite{ayazoglu2021extremely} (hardware-aware architecture) and SRGAN \cite{zhang2019super} (exploring GANS for SR).
Future work has to investigate which modules (like residual blocks or attention mechanisms) contribute the most regarding certain aspects to formulate a guideline for engineers on constructing a suitable architecture for a given problem: high quality vs. computational efficiency.

\section{Conclusion}
\label{sec:conclusion}
With the advent of DL, Super-Resolution (SR) has recently become a rapidly moving research area. However, despite promising results, the field continues to face challenges that call for more research, e.g., flexible upsampling. 
We reviewed the area of SR with recent advances and examined state-of-the-art models, such as transformer-based SR, and other architecture designs proposed lately (e.g., denoising diffusion probabilistic models).
We critically examined current strategies and identified new research areas. 
We complemented previous surveys by incorporating the latest developments and ideas, such as uncertainty-driven loss functions, new normalization techniques, wavelet networks, or neural architecture search. 
Also, we added various visualizations of models and methods in the chapters and the supplementary material to make navigation through this domain more accessible.
We hope this review helps researchers to push the boundaries of DL applied to SR further.

\ifCLASSOPTIONcompsoc
  \section*{Acknowledgments}
\else
  \section*{Acknowledgment}
\fi
This work was supported by the BMBF projects ExplAINN (Grant 01IS19074), EDeL (Grant 01IS19075), and  XAINES (Grant 01IW20005).

\vspace{-15 mm}

\begin{IEEEbiography}[{\includegraphics[width=1in,height=1in,clip,keepaspectratio]{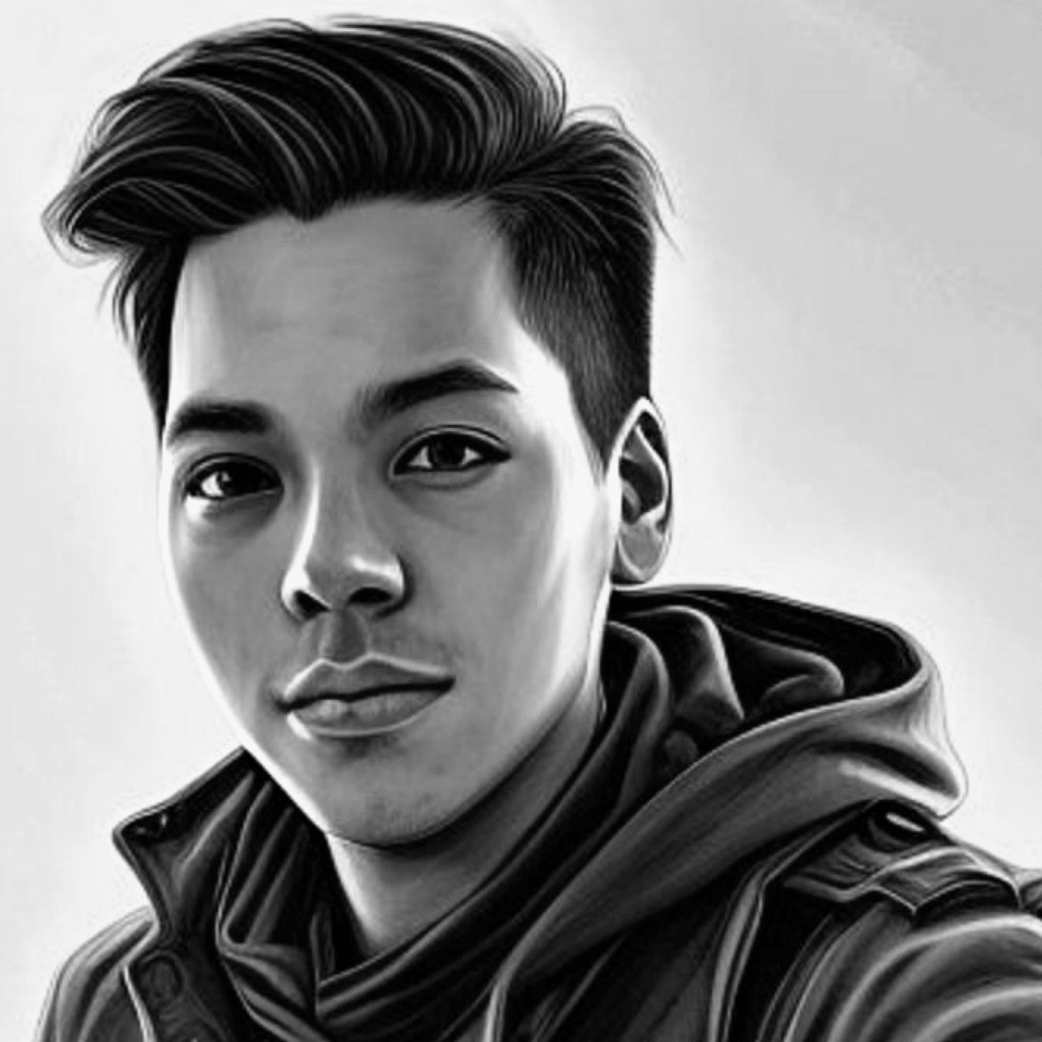}}]{Brian B. Moser}
 is a Ph.D. student at the TU Kaiserslautern and research assistant at the German Research Center for Artificial Intelligence (DFKI) in Kaiserslautern. He received the M.Sc. degree in computer science from the TU Kaiserslautern in 2021. His research interests include image super-resolution and deep learning.
\end{IEEEbiography}

\vskip -4.1\baselineskip plus -1fil

\begin{IEEEbiography}[{\includegraphics[width=1in,height=1in,clip,keepaspectratio]{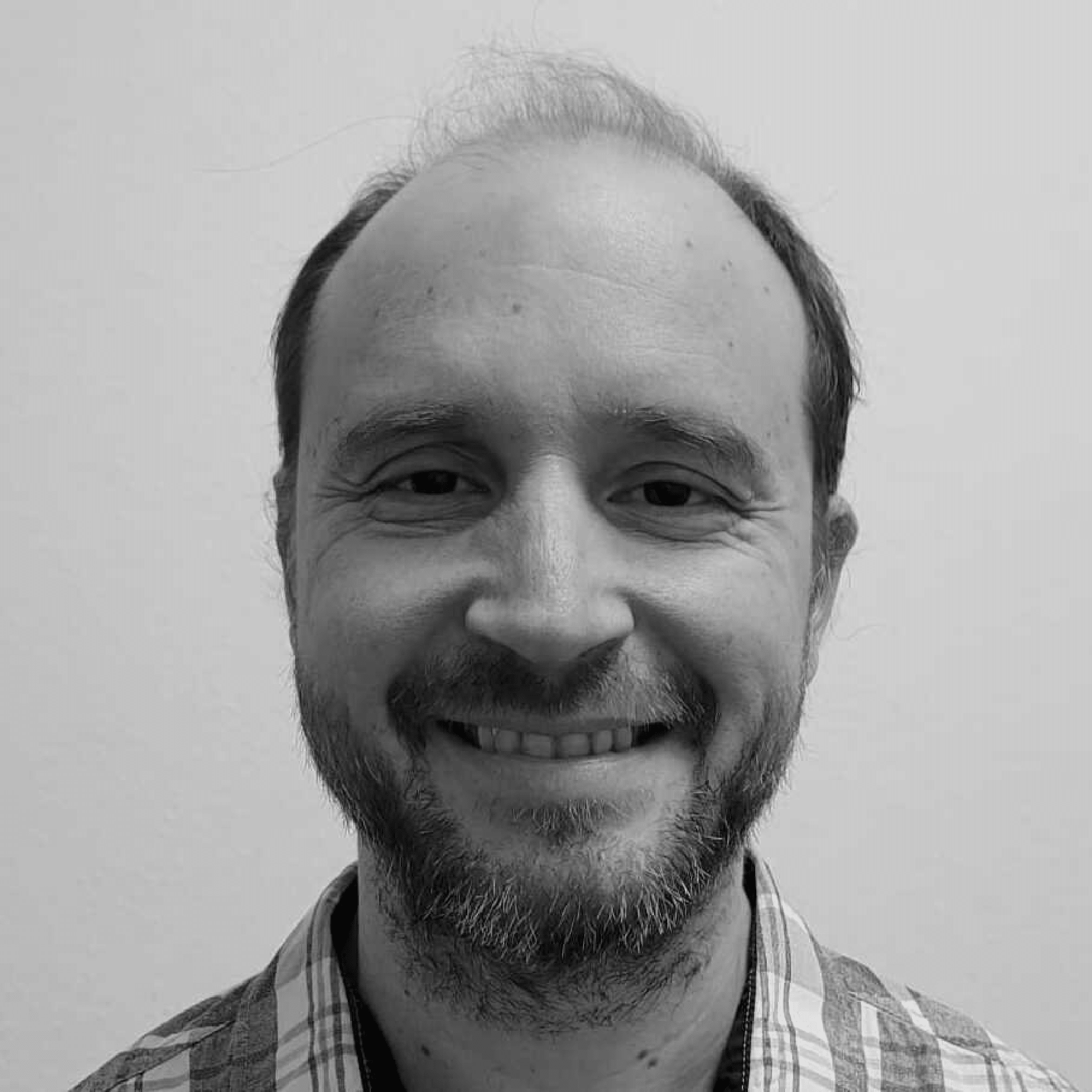}}]{Federico Raue}
 is a Senior Researcher at the German Research Center for Artificial Intelligence (DFKI) in Kaiserslautern. He received his PhD. degree at TU Kaiserslautern in 2018 and his M.Sc.  degree in Artificial Intelligence from Katholieke Universiteit Leuven in 2005. His research interests include meta-learning and multimodal machine learning.
\end{IEEEbiography}

\vskip -4.1\baselineskip plus -1fil

\begin{IEEEbiography}[{\includegraphics[width=1in,height=1in,clip,keepaspectratio]{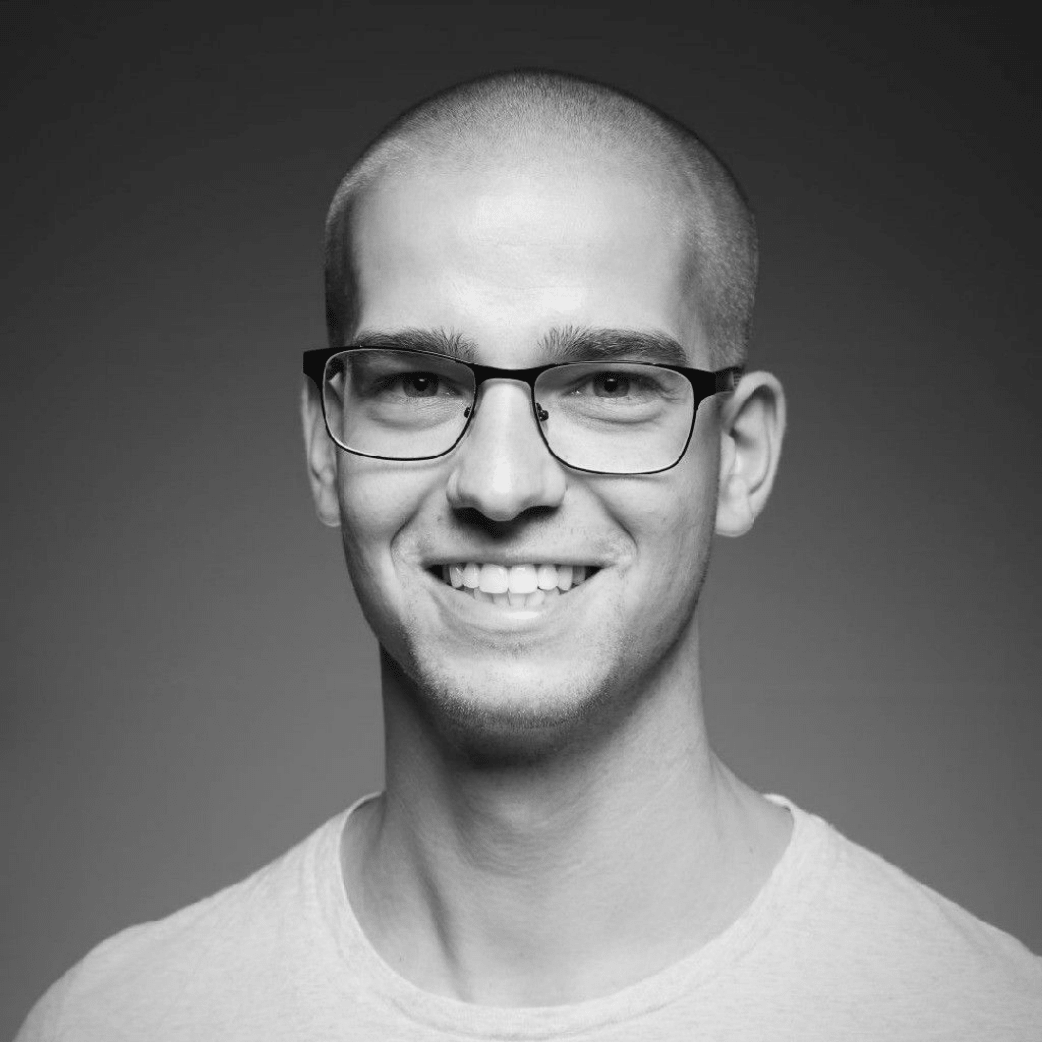}}]{Stanislav Frolov}
is a Ph.D. student at the TU Kaiserslautern and research assistant at the German Research Center for Artificial Intelligence (DFKI) in Kaiserslautern. He received the M.Sc. degree in electrical engineering from the Karlsruhe Institute of Technology in 2017. His research interests include generative models and deep learning.
\end{IEEEbiography}

\vskip -4.1\baselineskip plus -1fil

\begin{IEEEbiography}[{\includegraphics[width=1in,height=1in,clip,keepaspectratio]{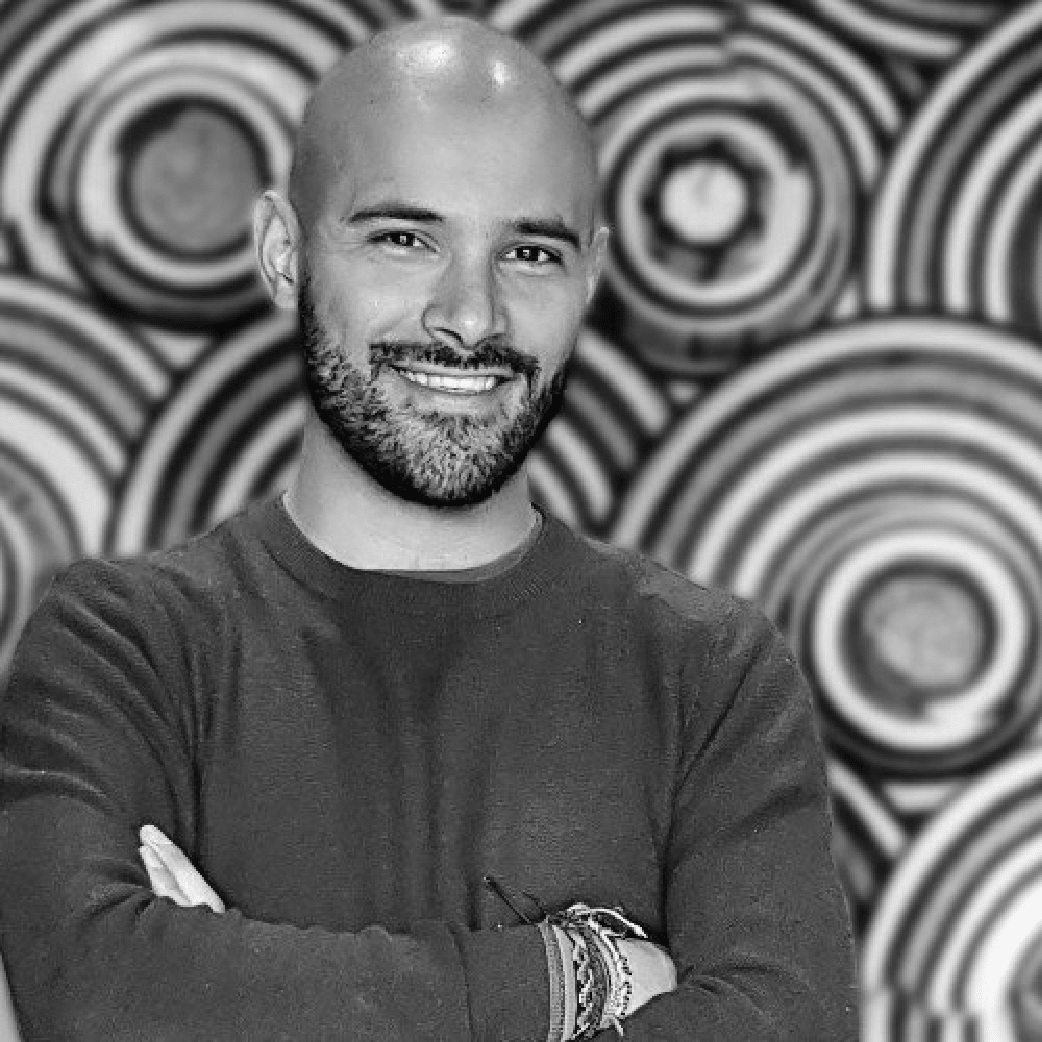}}]{Sebastian Palacio}
is a researcher in machine learning and head of the multimedia analysis and data mining group at the German Research Center for Artificial Intelligence (DFKI). His Ph.D. topic was about explainable AI with applications in computer vision. Other research interests include adversarial attacks, multi-task, curriculum and self-supervised learning.
\end{IEEEbiography}

\vskip -4.1\baselineskip plus -1fil

\begin{IEEEbiography}[{\includegraphics[width=1in,height=1in,clip,keepaspectratio]{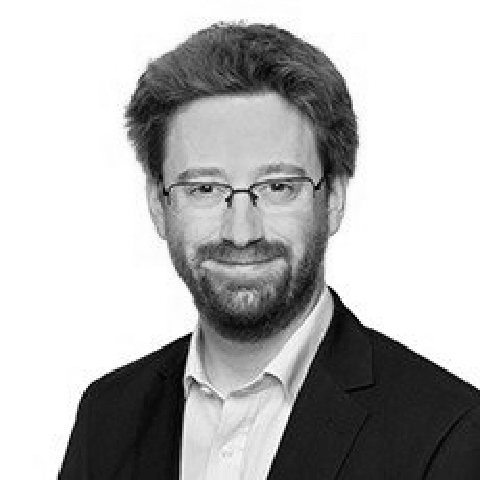}}]{Jörn Hees}
is a Professor for data science at the University of Applied Sciences Bonn-Rhein-Sieg. He received his Ph.D. from the TU Kaiserslautern in 2018 on the topic of simulating human associations with linked data. His research interests include machine learning, knowledge graphs, and multimedia analysis.
\end{IEEEbiography}

\vskip -4.1\baselineskip plus -1fil

\begin{IEEEbiography}[{\includegraphics[width=1in,height=1in,clip,keepaspectratio]{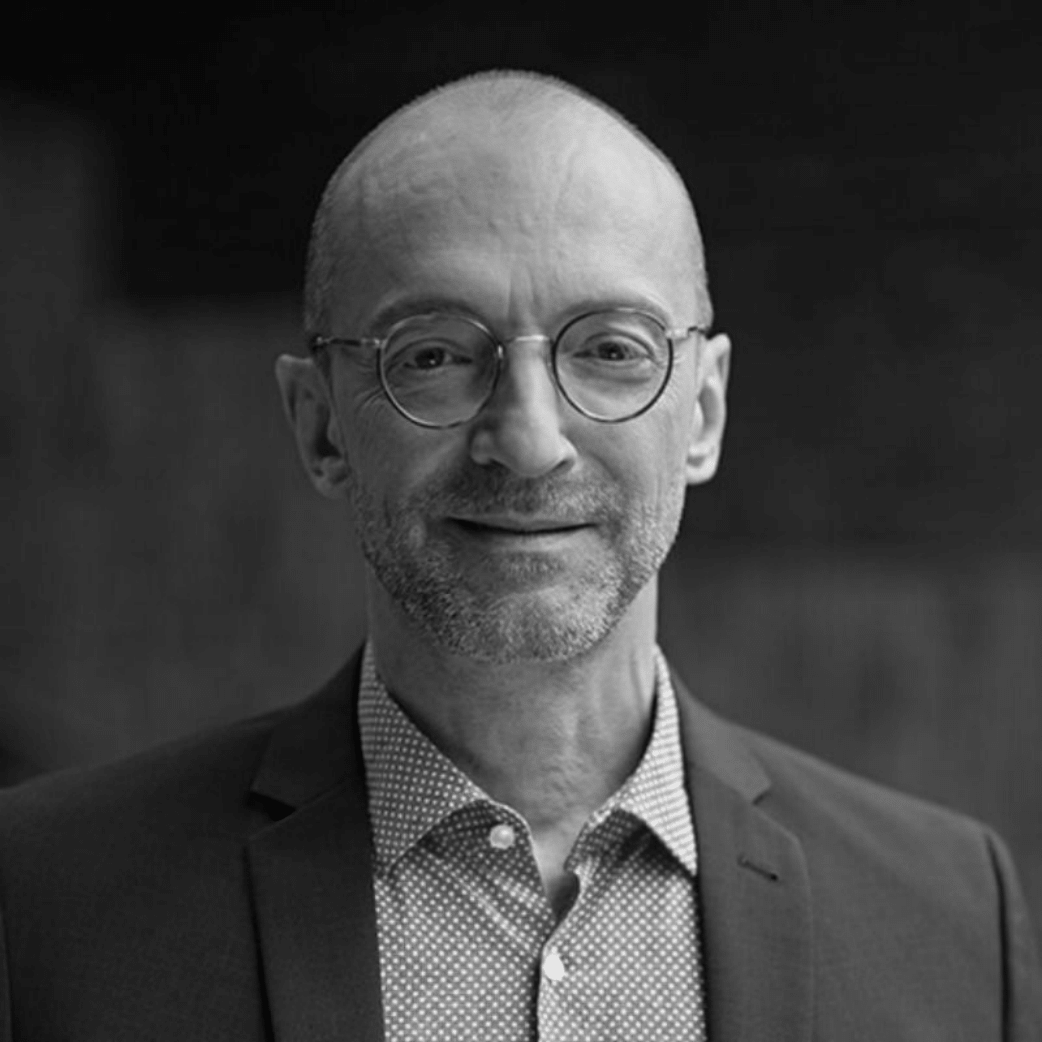}}]{Andreas Dengel}
is a Professor at the Department of Computer Science at TU Kaiserslautern and Executive Director of the German Research Center for Artificial Intelligence (DFKI) in Kaiserslautern, Head of the Smart Data and Knowledge  Services research area at DFKI and of the DFKI Deep Learning Competence Center. His research focuses on machine learning, pattern recognition, quantified learning, data mining, semantic technologies and document analysis.
\end{IEEEbiography}

\ifCLASSOPTIONcaptionsoff
  \newpage
\fi

\definecolor{simple}{HTML}{81ecec}
\definecolor{residual}{HTML}{55efc4}
\definecolor{recurrent}{HTML}{a29bfe}
\definecolor{attention}{HTML}{ffeaa7}
\definecolor{wavelet}{HTML}{74b9ff}
\definecolor{nas}{HTML}{fab1a0}
\definecolor{lightweight}{HTML}{dfe6e9}
\definecolor{unsupervised}{HTML}{fd79a8}

\addtolength{\tabcolsep}{-2.5pt}   
\begin{table*}
\begin{center}
    \caption{\label{tab:benchmark}Comparison (PSNR/SSIM) of SR approaches discussed in this work: \colorbox{simple}{Simple}, \colorbox{residual}{residual}, \colorbox{recurrent}{recurrent}, \colorbox{attention}{attention-based}, \colorbox{lightweight}{lightweight} and \colorbox{wavelet}{wavelet} networks, as well as \colorbox{nas}{NAS} derived networks and \colorbox{unsupervised}{unsupervised} trained models. The ordering reflects roughly the performance ranking. However, some comparisons are not fair due to different settings, e.g., NAS vs unsupervised results. Also, some publications do not report results on the commonly used datasets. Therefore, they are not listed (e.g. WESPE \cite{ignatov2018wespe}, DSN \cite{cheng2020zero}, CinCGan \cite{yuan2018unsupervised}).
    }
    \centerline{%
    \begin{tabular}{| c | c c c | c c c | c c c |}
    \hline
    \multirow{2}{*}{Model} & \multicolumn{3}{ c |}{Set5 \cite{bevilacqua2012low}} & \multicolumn{3}{ c |}{Set14 \cite{zeyde2010single}} & \multicolumn{3}{ c |}{BSDS100 \cite{martin2001database}}\\ 
    & x2 & x3 & x4 & x2 & x3 & x4 & x2 & x3 & x4 \\
    & PSNR/SSIM & PSNR/SSIM & PSNR/SSIM & PSNR/SSIM & PSNR/SSIM & PSNR/SSIM & PSNR/SSIM & PSNR/SSIM & PSNR/SSIM \\
    \hline
    Bicubic & 33.66/0.9542 & 30.39/0.8682 & 28.42/0.8104 & 30.24/0.8688 & 27.55/0.7742 & 26.00/0.7027 & 29.56/0.8431 & 27.21/0.7385 & 25.96/0.6675 \\
    \cellcolor[HTML]{fd79a8} DIP \cite{ulyanov2018deep}  & n.a. & n.a. & 28.90/n.a. & n.a. & n.a. & 27.00/n.a. & n.a. & n.a. & n.a. \\
    \cellcolor[HTML]{fab1a0} NAS-DIP \cite{chen2020dip} & 35.90/n.a. & n.a. & 31.09/n.a. & 31.89/n.a. & n.a. & 28.37/n.a. & n.a. & n.a. & n.a. \\
    \cellcolor[HTML]{81ecec} SRCNN \cite{dong2015image} & 36.66/0.9542 & 32.75/0.9090 & 30.48/0.8628 & 32.42/0.9063 & 29.28/0.8209 & 27.49/0.7503 & 31.36/0.8879 & 28.41/0.7863 & 26.90/0.7101 \\
    \cellcolor[HTML]{81ecec} FSRCNN \cite{dong2016accelerating} & 37.00/0.9558 & 33.16/0.9140 & 30.71/0.8657 & 32.63/0.9088 & 29.43/0.8242 & 27.59/0.7535 & 31.51/0.8910 & n.a. & 26.97/0.7140 \\
    \hline
    \cellcolor[HTML]{81ecec} ESPCN \cite{shi2016real} & n.a. & 33.13/n.a. & 30.90/n.a. & n.a. & 29.49/n.a. & 27.73/n.a. & n.a. & n.a. & n.a. \\
    \cellcolor[HTML]{fd79a8} MZSR \cite{soh2020meta}  & 37.25/0.9567 & n.a. & n.a. & n.a. & n.a. & n.a. & 31.64/0.8928 & n.a. & n.a. \\
    \cellcolor[HTML]{fd79a8} ZSSR \cite{shocher2018zero}  & 37.37/0.9570 & 33.42/0.9188 & 31.13/0.8796 & 33.00/0.9108 & 29.80/0.8304 & 28.01/0.7651 & 31.65/0.8920 & 28.67/0.7945 & 27.12/0.7211 \\
    \cellcolor[HTML]{81ecec} LapSRN \cite{hui2018fast} & 37.52/0.9591 & 33.81/0.9220 & 31.54/0.8852 & 32.99/0.9124 & 29.79/0.8325 & 28.09/0.7700 & 31.80/0.8952 & 28.82/0.7980 & 27.32/0.7275 \\  
    \cellcolor[HTML]{55efc4} VDSR \cite{kim2016accurate} & 37.53/0.9587 & 33.66/0.9213 & 31.35/0.8838 & 33.03/0.9124 & 29.77/0.8314 & 28.01/0.7674 & 31.90/0.8960 & 28.82/0.7976 & 27.29/0.7251 \\
    \hline
    \cellcolor[HTML]{74b9ff} DWSR \cite{guo2017deep} & 37.43/0.9568 & 33.82/0.9215 & 31.39/0.8833 & 33.07/0.9106 & 29.83/0.8308 & 28.04/0.7669 & 31.80/0.8940 & n.a. & 27.25/0.7240 \\
    \cellcolor[HTML]{a29bfe} DRCN \cite{kim2016deeply}& 37.63/0.9588 & 33.82/0.9226 & 31.53/0.8854 & 33.04/0.9118 & 29.76/0.8311 & 28.02/0.7670 & 31.85/0.8942 & 28.80/0.7963 & 28.23/0.7233 \\
    \cellcolor[HTML]{fab1a0} MoreMNAS \cite{chu2020multi} & 37.57/0.9584 & n.a. & n.a. & 33.25/0.9142 & n.a. & n.a. & 31.94/0.8966 & n.a. & n.a. \\
    \cellcolor[HTML]{55efc4} RED \cite{mao2016image} & 37.66/0.9599 & 33.82/0.9230 & 31.51/0.8869 & 32.94/0.9144 & 29.61/0.8341 & 27.86/0.7718 & 31.99/0.8974 & 28.93/0.7994 & 27.40/0.7290 \\
    \cellcolor[HTML]{a29bfe} DSRN \cite{han2018image}& 37.66/0.9590 & 33.88/0.9220 & 31.40/0.8830 & 33.15/0.9130 & 30.26/0.8370 & 28.07/0.7700 & 32.10/0.8970 & 28.81/0.7970 & 27.25/0.7240 \\
    \hline
    \cellcolor[HTML]{a29bfe} DRRN \cite{tai2017image} & 37.74/0.9591 & 34.03/0.9244 & 31.68/0.8888 & 33.23/0.9136 & 29.96/0.8349 & 28.21/0.7720 & 32.05/0.8973 & 28.95/0.8004 & 27.38/0.7284 \\
    \cellcolor[HTML]{dfe6e9} CARN-M \cite{ahn2018fast} & 37.53/0.9583 & 33.99/0.9236 & 31.92/0.8903 & 33.26/0.9141 & 30.08/0.8367 & 28.42/0.7762 & 31.92/0.8960 & 28.91/0.8000 & 27.44/0.7304 \\
    \cellcolor[HTML]{a29bfe} MemNet \cite{tai2017memnet} & 37.78/0.9597 & 34.09/0.9248 & 31.74/0.8893 & 33.28/0.9142 & 30.00/0.8350 & 28.26/0.7723 & 32.08/0.8978 & 28.96/0.8001 & 27.40/0.7281 \\
    \cellcolor[HTML]{55efc4} IDN \cite{hui2018fast} & 37.83/0.9600 & 34.11/0.9253 & 31.82/0.8903 & 33.30/0.9148 & 29.99/0.8354 & 28.25/0.7730 & 32.08/0.8985 & 28.95/0.8013 & 27.41/0.7297 \\
    \cellcolor[HTML]{fab1a0} FALSR \cite{chu2021fast} & 37.82/0.9595 & n.a. & n.a. & 33.55/0.9168 & n.a. & n.a. & 32.12/0.8987 & n.a. & n.a. \\
    \hline
    \cellcolor[HTML]{74b9ff} MWCNN \cite{liu2018multi} & 37.91/0.9600 & 34.17/0.9271 & 32.12/0.8941 & 33.70/0.9182 & 30.16/0.8414 & 28.41/0.7816 & 32.23/0.8999 & 29.12/0.8060 & 27.62/0.7355 \\
    \cellcolor[HTML]{a29bfe} NLRN \cite{liu2018non} & 38.00/0.9603 & 34.27/0.9266 & 31.92/0.8916 & 33.46/0.9159 & 30.16/0.8374 & 28.36/0.7745 & 32.19/0.8992 & 29.06/0.8026 & 27.48/0.7306 \\
    \cellcolor[HTML]{dfe6e9} CARN \cite{ahn2018fast} & 37.76/0.9590 & 34.29/0.9255 & 32.13/0.8937 & 33.52/0.9166 & 30.29/0.8407 & 28.60/0.7806 & 32.09/0.8978 & 29.06/0.8034 & 27.58/0.7349 \\
    \cellcolor[HTML]{dfe6e9} IMDN \cite{hui2019lightweight} & 38.00/0.9605 & 34.36/0.9270 & 32.21/0.8948 & 33.63/0.9177 & 30.32/0.8417 & 28.58/0.7811 & 32.19/0.8996 & 29.09/0.8046 & 27.56/0.7353 \\
    \cellcolor[HTML]{dfe6e9} RFDN \cite{liu2020residual} & 38.05/0.9606 & 34.41/0.9273 & 32.24/0.8952 & 33.68/0.9184 & 30.34/0.8420 & 28.61/0.7819 & 32.16/0.8994 & 29.09/0.8050 & 27.57/0.7360 \\
    \hline
    \cellcolor[HTML]{dfe6e9} RFDN-L \cite{liu2020residual} & 38.08/0.9606 & 34.47/0.9280 & 32.28/0.8957 & 33.67/0.9190 & 30.35/0.8421 & 28.61/0.7818 & 32.18/0.8996 & 29.11/0.8053 & 27.58/0.7363 \\
    \cellcolor[HTML]{fab1a0} DeCoNAS \cite{ahn2021neural} & 37.96/0.9594 & n.a. & n.a. & 33.63/0.9175 & n.a. & n.a. & 32.15/0.8986 & n.a. & n.a. \\
    \cellcolor[HTML]{fab1a0} HNAS \cite{guo2020hierarchical} & 38.11/0.9640 & n.a. & n.a. & 33.60/0.9200 & n.a. & n.a. & 32.17/0.9020 & n.a. & n.a. \\
    \cellcolor[HTML]{fab1a0} ESRN \cite{song2020efficient} & 38.04/0.9607 & 34.46/0.9281 & 32.26/0.8957 & 33.71/0.9185 & 30.43/0.8439 & 28.63/0.7818 & 32.23/0.9005 & 29.15/0.8072 & 27.62/0.7378 \\
    \cellcolor[HTML]{a29bfe} SRFBN \cite{li2019feedback} & 38.11/0.9609 & 34.70/0.9292 & 32.47/0.8983 & 33.82/0.9196 & 30.51/0.8461 & 28.81/0.7868 & 32.29/0.9010 & 29.24/0.8084 & 27.72/0.7409 \\
    \hline
    \cellcolor[HTML]{a29bfe} SRFBN+ \cite{li2019feedback} & 38.18/0.9611 & 34.77/0.9297 & 32.56/0.8992 & 33.90/0.9203 & 30.61/0.8473 & 28.87/0.7881 & 32.34/0.9015 & 29.29/0.8093 & 27.77/0.7419 \\
    \cellcolor[HTML]{dfe6e9} MDSR \cite{lim2017enhanced} & 38.17/0.9605 & 34.77/0.9288 & 32.60/0.8982 & 33.92/0.9203 & 30.53/0.8465 & 28.82/0.7876 & 32.34/0.9014 & 29.30/0.8101 & 27.78/0.7425 \\
    \cellcolor[HTML]{dfe6e9} EDSR \cite{lim2017enhanced} & 38.20/0.9606 & 34.76/0.9290 & 32.62/0.8984 & 34.02/0.9204 & 30.66/0.8481 & 28.94/0.7901 & 32.37/0.9018 & 29.32/0.8104 & 27.79/0.7437 \\
    \cellcolor[HTML]{74b9ff} WRAN \cite{xue2020wavelet} & 38.32/0.9630 & 34.79/0.9310 & 32.36/0.8990 & 34.21/0.9220 & 30.71/0.8520 & 28.60/0.7860 & 32.57/0.9070 & 29.36/0.8190 & 27.71/0.7420 \\
    \cellcolor[HTML]{ffeaa7} DRLN \cite{anwar2020densely} & 38.27/0.9616 & 34.78/0.9303 & 32.63/0.9002 & 34.28/0.9231 & 30.73/0.8488 & 28.94/0.7900 & 32.44/0.9028 & 29.36/0.8117 & 27.83/0.7444 \\
    \hline
    \cellcolor[HTML]{ffeaa7} HAN \cite{niu2020single} & 38.33/0.9617 & 34.85/0.9305 & 32.75/0.9016 & 34.24/0.9224 & 30.77/0.8495 & 28.99/0.7907 & 32.45/0.9030 & 29.39/0.8120 & 27.85/0.7454 \\
    \cellcolor[HTML]{ffeaa7} DRLN+ \cite{anwar2020densely} & 38.34/0.9619 & 34.86/0.9307 & 32.74/0.9013 & 34.43/0.9247 & 30.80/0.8498 & 29.02/0.7914 & 32.47/0.9032 & 29.40/0.8125 & 27.87/0.7453 \\
    \cellcolor[HTML]{ffeaa7} SwinIR \cite{liu2021swin} & 38.46/0.9624 & 35.04/0.9322 & 32.93/0.9043 & 33.07/0.9106 & 31.00/0.8542 & 29.15/0.7958 & 31.80/0.8940 & 29.49/0.8150 & 27.95/0.7494 \\
    \cellcolor[HTML]{74b9ff} WIDN \cite{sahito2019wavelet} & 39.60/0.9830 & 34.48/0.9430 & 32.85/0.9290 & 34.44/0.9800 & 30.95/0.9310 & 29.75/0.9090 & 33.52/0.9790 & 29.99/0.9280 & 28.10/0.9100 \\
    \cellcolor[HTML]{dfe6e9} CAR (EDSR) \cite{sun2020learned} & 38.94/0.9658 & n.a. & 33.88/0.9174 & 35.61/0.9404 & n.a. & 30.31/0.8382 & 33.83/0.9262 & n.a. & 29.15/0.8001 \\
    \hline
    \end{tabular}
    }
\end{center}
\end{table*}
\addtolength{\tabcolsep}{2.5pt} 

\bibliography{main}

\begin{thebibliography}{100}
\providecommand{\url}[1]{#1}
\csname url@samestyle\endcsname
\providecommand{\newblock}{\relax}
\providecommand{\bibinfo}[2]{#2}
\providecommand{\BIBentrySTDinterwordspacing}{\spaceskip=0pt\relax}
\providecommand{\BIBentryALTinterwordstretchfactor}{4}
\providecommand{\BIBentryALTinterwordspacing}{\spaceskip=\fontdimen2\font plus
\BIBentryALTinterwordstretchfactor\fontdimen3\font minus
  \fontdimen4\font\relax}
\providecommand{\BIBforeignlanguage}[2]{{%
\expandafter\ifx\csname l@#1\endcsname\relax
\typeout{** WARNING: IEEEtran.bst: No hyphenation pattern has been}%
\typeout{** loaded for the language `#1'. Using the pattern for}%
\typeout{** the default language instead.}%
\else
\language=\csname l@#1\endcsname
\fi
#2}}
\providecommand{\BIBdecl}{\relax}
\BIBdecl

\bibitem{zeyde2010single}
R.~Zeyde, M.~Elad, and M.~Protter, ``On single image scale-up using
  sparse-representations,'' in \emph{International conference on curves and
  surfaces}.\hskip 1em plus 0.5em minus 0.4em\relax Springer, 2010, pp.
  711--730.

\bibitem{martin2001database}
D.~Martin, C.~Fowlkes, D.~Tal, and J.~Malik, ``A database of human segmented
  natural images and its application to evaluating segmentation algorithms and
  measuring ecological statistics,'' in \emph{Proceedings Eighth IEEE
  International Conference on Computer Vision. ICCV 2001}, vol.~2.\hskip 1em
  plus 0.5em minus 0.4em\relax IEEE, 2001, pp. 416--423.

\bibitem{valsesia2021permutation}
D.~Valsesia and E.~Magli, ``Permutation invariance and uncertainty in
  multitemporal image super-resolution,'' \emph{arXiv preprint
  arXiv:2105.12409}, 2021.

\bibitem{bashir2021comprehensive}
S.~M.~A. Bashir, Y.~Wang, M.~Khan, and Y.~Niu, ``A comprehensive review of deep
  learning-based single image super-resolution,'' \emph{PeerJ Computer
  Science}, vol.~7, p. e621, 2021.

\bibitem{9044873}
Z.~Wang, J.~Chen, and S.~C.~H. Hoi, ``Deep learning for image super-resolution:
  A survey,'' \emph{IEEE Transactions on Pattern Analysis and Machine
  Intelligence}, vol.~43, no.~10, pp. 3365--3387, 2021.

\bibitem{anwar2020densely}
S.~Anwar and N.~Barnes, ``Densely residual laplacian super-resolution,''
  \emph{IEEE Transactions on Pattern Analysis and Machine Intelligence}, 2020.

\bibitem{sun2020learned}
W.~Sun and Z.~Chen, ``Learned image downscaling for upscaling using content
  adaptive resampler,'' \emph{IEEE Transactions on Image Processing}, vol.~29,
  pp. 4027--4040, 2020.

\bibitem{Yang2014SingleImageSA}
C.-Y. Yang, C.~Ma, and M.-H. Yang, ``Single-image super-resolution: A
  benchmark,'' in \emph{ECCV}, 2014.

\bibitem{Thapa2016APC}
D.~Thapa, K.~Raahemifar, W.~R. Bobier, and V.~Lakshminarayanan, ``A performance
  comparison among different super-resolution techniques,'' \emph{Comput.
  Electr. Eng.}, vol.~54, pp. 313--329, 2016.

\bibitem{chu2020multi}
X.~Chu, B.~Zhang, and R.~Xu, ``Multi-objective reinforced evolution in mobile
  neural architecture search,'' in \emph{European Conference on Computer
  Vision}.\hskip 1em plus 0.5em minus 0.4em\relax Springer, 2020, pp. 99--113.

\bibitem{kappeler2016video}
A.~Kappeler, S.~Yoo, Q.~Dai, and A.~K. Katsaggelos, ``Video super-resolution
  with convolutional neural networks,'' \emph{IEEE transactions on
  computational imaging}, vol.~2, no.~2, pp. 109--122, 2016.

\bibitem{sajjadi2017enhancenet}
M.~S. Sajjadi, B.~Scholkopf, and M.~Hirsch, ``Enhancenet: Single image
  super-resolution through automated texture synthesis,'' in \emph{Proceedings
  of the IEEE International Conference on Computer Vision}, 2017, pp.
  4491--4500.

\bibitem{ledig2017photo}
C.~Ledig, L.~Theis, F.~Husz{\'a}r, J.~Caballero, A.~Cunningham, A.~Acosta,
  A.~Aitken, A.~Tejani, J.~Totz, Z.~Wang \emph{et~al.}, ``Photo-realistic
  single image super-resolution using a generative adversarial network,'' in
  \emph{Proceedings of the IEEE conference on computer vision and pattern
  recognition}, 2017, pp. 4681--4690.

\bibitem{wang2004image}
Z.~Wang, A.~C. Bovik, H.~R. Sheikh, and E.~P. Simoncelli, ``Image quality
  assessment: from error visibility to structural similarity,'' \emph{IEEE
  transactions on image processing}, vol.~13, no.~4, pp. 600--612, 2004.

\bibitem{ponomarenko2015image}
N.~Ponomarenko, L.~Jin, O.~Ieremeiev, V.~Lukin, K.~Egiazarian, J.~Astola,
  B.~Vozel, K.~Chehdi, M.~Carli, F.~Battisti \emph{et~al.}, ``Image database
  tid2013: Peculiarities, results and perspectives,'' \emph{Signal processing:
  Image communication}, vol.~30, pp. 57--77, 2015.

\bibitem{kim2017deep}
J.~Kim and S.~Lee, ``Deep learning of human visual sensitivity in image quality
  assessment framework,'' in \emph{Proceedings of the IEEE conference on
  computer vision and pattern recognition}, 2017, pp. 1676--1684.

\bibitem{talebi2018nima}
H.~Talebi and P.~Milanfar, ``Nima: Neural image assessment,'' \emph{IEEE
  Transactions on Image Processing}, vol.~27, no.~8, pp. 3998--4011, 2018.

\bibitem{ma2017dipiq}
K.~Ma, W.~Liu, T.~Liu, Z.~Wang, and D.~Tao, ``dipiq: Blind image quality
  assessment by learning-to-rank discriminable image pairs,'' \emph{IEEE
  Transactions on Image Processing}, vol.~26, no.~8, pp. 3951--3964, 2017.

\bibitem{burges2005learning}
C.~Burges, T.~Shaked, E.~Renshaw, A.~Lazier, M.~Deeds, N.~Hamilton, and
  G.~Hullender, ``Learning to rank using gradient descent,'' in
  \emph{Proceedings of the 22nd international conference on Machine learning},
  2005, pp. 89--96.

\bibitem{liu2011learning}
T.-Y. Liu, ``Learning to rank for information retrieval,'' 2011.

\bibitem{liu2017rankiqa}
X.~Liu, J.~Van De~Weijer, and A.~D. Bagdanov, ``Rankiqa: Learning from rankings
  for no-reference image quality assessment,'' in \emph{Proceedings of the IEEE
  International Conference on Computer Vision}, 2017, pp. 1040--1049.

\bibitem{koch2015siamese}
G.~Koch, R.~Zemel, R.~Salakhutdinov \emph{et~al.}, ``Siamese neural networks
  for one-shot image recognition,'' in \emph{ICML deep learning workshop},
  vol.~2.\hskip 1em plus 0.5em minus 0.4em\relax Lille, 2015.

\bibitem{zhang2018unreasonable}
R.~Zhang, P.~Isola, A.~A. Efros, E.~Shechtman, and O.~Wang, ``The unreasonable
  effectiveness of deep features as a perceptual metric,'' in \emph{Proceedings
  of the IEEE conference on computer vision and pattern recognition}, 2018, pp.
  586--595.

\bibitem{simonyan2014very}
K.~Simonyan and A.~Zisserman, ``Very deep convolutional networks for
  large-scale image recognition,'' \emph{arXiv preprint arXiv:1409.1556}, 2014.

\bibitem{ding2020image}
K.~Ding, K.~Ma, S.~Wang, and E.~P. Simoncelli, ``Image quality assessment:
  Unifying structure and texture similarity,'' \emph{IEEE transactions on
  pattern analysis and machine intelligence}, 2020.

\bibitem{xue2013gradient}
W.~Xue, L.~Zhang, X.~Mou, and A.~C. Bovik, ``Gradient magnitude similarity
  deviation: A highly efficient perceptual image quality index,'' \emph{IEEE
  Transactions on Image Processing}, vol.~23, no.~2, pp. 684--695, 2013.

\bibitem{zhang2011fsim}
L.~Zhang, L.~Zhang, X.~Mou, and D.~Zhang, ``Fsim: A feature similarity index
  for image quality assessment,'' \emph{IEEE transactions on Image Processing},
  vol.~20, no.~8, pp. 2378--2386, 2011.

\bibitem{reisenhofer2018haar}
R.~Reisenhofer, S.~Bosse, G.~Kutyniok, and T.~Wiegand, ``A haar wavelet-based
  perceptual similarity index for image quality assessment,'' \emph{Signal
  Processing: Image Communication}, vol.~61, pp. 33--43, 2018.

\bibitem{wang2003multiscale}
Z.~Wang, E.~P. Simoncelli, and A.~C. Bovik, ``Multiscale structural similarity
  for image quality assessment,'' in \emph{The Thrity-Seventh Asilomar
  Conference on Signals, Systems \& Computers, 2003}, vol.~2.\hskip 1em plus
  0.5em minus 0.4em\relax Ieee, 2003, pp. 1398--1402.

\bibitem{zhang2017gradient}
B.~Zhang, P.~V. Sander, and A.~Bermak, ``Gradient magnitude similarity
  deviation on multiple scales for color image quality assessment,'' in
  \emph{2017 IEEE International Conference on Acoustics, Speech and Signal
  Processing (ICASSP)}.\hskip 1em plus 0.5em minus 0.4em\relax IEEE, 2017, pp.
  1253--1257.

\bibitem{bevilacqua2012low}
M.~Bevilacqua, A.~Roumy, C.~Guillemot, and M.~L. Alberi-Morel, ``Low-complexity
  single-image super-resolution based on nonnegative neighbor embedding,''
  2012.

\bibitem{matsui2017sketch}
Y.~Matsui, K.~Ito, Y.~Aramaki, A.~Fujimoto, T.~Ogawa, T.~Yamasaki, and
  K.~Aizawa, ``Sketch-based manga retrieval using manga109 dataset,''
  \emph{Multimedia Tools and Applications}, vol.~76, no.~20, pp.
  21\,811--21\,838, 2017.

\bibitem{dong2016accelerating}
C.~Dong, C.~C. Loy, and X.~Tang, ``Accelerating the super-resolution
  convolutional neural network,'' in \emph{European conference on computer
  vision}.\hskip 1em plus 0.5em minus 0.4em\relax Springer, 2016, pp. 391--407.

\bibitem{MATLAB:2017b}
\emph{{MATLAB version 9.3.0.713579 (R2017b)}}, The Mathworks, Inc., Natick,
  Massachusetts, 2017.

\bibitem{lugmayr2020ntire}
A.~Lugmayr, M.~Danelljan, and R.~Timofte, ``Ntire 2020 challenge on real-world
  image super-resolution: Methods and results,'' in \emph{Proceedings of the
  IEEE/CVF Conference on Computer Vision and Pattern Recognition Workshops},
  2020, pp. 494--495.

\bibitem{blau20182018}
Y.~Blau, R.~Mechrez, R.~Timofte, T.~Michaeli, and L.~Zelnik-Manor, ``The 2018
  pirm challenge on perceptual image super-resolution,'' in \emph{Proceedings
  of the European Conference on Computer Vision (ECCV) Workshops}, 2018, pp.
  0--0.

\bibitem{zhang2018image}
Y.~Zhang, K.~Li, K.~Li, L.~Wang, B.~Zhong, and Y.~Fu, ``Image super-resolution
  using very deep residual channel attention networks,'' in \emph{Proceedings
  of the European conference on computer vision (ECCV)}, 2018, pp. 286--301.

\bibitem{lim2017enhanced}
B.~Lim, S.~Son, H.~Kim, S.~Nah, and K.~Mu~Lee, ``Enhanced deep residual
  networks for single image super-resolution,'' in \emph{Proceedings of the
  IEEE conference on computer vision and pattern recognition workshops}, 2017,
  pp. 136--144.

\bibitem{dong2015image}
C.~Dong, C.~C. Loy, K.~He, and X.~Tang, ``Image super-resolution using deep
  convolutional networks,'' \emph{IEEE transactions on pattern analysis and
  machine intelligence}, vol.~38, no.~2, pp. 295--307, 2015.

\bibitem{tong2017image}
T.~Tong, G.~Li, X.~Liu, and Q.~Gao, ``Image super-resolution using dense skip
  connections,'' in \emph{Proceedings of the IEEE international conference on
  computer vision}, 2017, pp. 4799--4807.

\bibitem{hu2019meta}
X.~Hu, H.~Mu, X.~Zhang, Z.~Wang, T.~Tan, and J.~Sun, ``Meta-sr: A
  magnification-arbitrary network for super-resolution,'' in \emph{Proceedings
  of the IEEE/CVF Conference on Computer Vision and Pattern Recognition}, 2019,
  pp. 1575--1584.

\bibitem{zhang2018residual}
Y.~Zhang, Y.~Tian, Y.~Kong, B.~Zhong, and Y.~Fu, ``Residual dense network for
  image super-resolution,'' in \emph{Proceedings of the IEEE conference on
  computer vision and pattern recognition}, 2018, pp. 2472--2481.

\bibitem{hui2018fast}
Z.~Hui, X.~Wang, and X.~Gao, ``Fast and accurate single image super-resolution
  via information distillation network,'' in \emph{Proceedings of the IEEE
  conference on computer vision and pattern recognition}, 2018, pp. 723--731.

\bibitem{lai2017deep}
W.-S. Lai, J.-B. Huang, N.~Ahuja, and M.-H. Yang, ``Deep laplacian pyramid
  networks for fast and accurate super-resolution,'' in \emph{Proceedings of
  the IEEE conference on computer vision and pattern recognition}, 2017, pp.
  624--632.

\bibitem{kendall2017uncertainties}
A.~Kendall and Y.~Gal, ``What uncertainties do we need in bayesian deep
  learning for computer vision?'' \emph{arXiv preprint arXiv:1703.04977}, 2017.

\bibitem{ning2021uncertainty}
Q.~Ning, W.~Dong, X.~Li, J.~Wu, and G.~Shi, ``Uncertainty-driven loss for
  single image super-resolution,'' \emph{Advances in Neural Information
  Processing Systems}, vol.~34, 2021.

\bibitem{kingma2013auto}
D.~P. Kingma and M.~Welling, ``Auto-encoding variational bayes,'' \emph{arXiv
  preprint arXiv:1312.6114}, 2013.

\bibitem{Lim_2017_CVPR_Workshops}
B.~Lim, S.~Son, H.~Kim, S.~Nah, and K.~M. Lee, ``Enhanced deep residual
  networks for single image super-resolution,'' in \emph{The IEEE Conference on
  Computer Vision and Pattern Recognition (CVPR) Workshops}, July 2017.

\bibitem{dong2018denoising}
W.~Dong, P.~Wang, W.~Yin, G.~Shi, F.~Wu, and X.~Lu, ``Denoising prior driven
  deep neural network for image restoration,'' \emph{IEEE transactions on
  pattern analysis and machine intelligence}, vol.~41, no.~10, pp. 2305--2318,
  2018.

\bibitem{goodfellow2014generative}
I.~Goodfellow, J.~Pouget-Abadie, M.~Mirza, B.~Xu, D.~Warde-Farley, S.~Ozair,
  A.~Courville, and Y.~Bengio, ``Generative adversarial nets,'' \emph{Advances
  in neural information processing systems}, vol.~27, 2014.

\bibitem{frolov2021adversarial}
S.~Frolov, T.~Hinz, F.~Raue, J.~Hees, and A.~Dengel, ``Adversarial
  text-to-image synthesis: A review,'' \emph{arXiv preprint arXiv:2101.09983},
  2021.

\bibitem{yuan2018unsupervised}
Y.~Yuan, S.~Liu, J.~Zhang, Y.~Zhang, C.~Dong, and L.~Lin, ``Unsupervised image
  super-resolution using cycle-in-cycle generative adversarial networks,'' in
  \emph{Proceedings of the IEEE Conference on Computer Vision and Pattern
  Recognition Workshops}, 2018, pp. 701--710.

\bibitem{wang2018fully}
Y.~Wang, F.~Perazzi, B.~McWilliams, A.~Sorkine-Hornung, O.~Sorkine-Hornung, and
  C.~Schroers, ``A fully progressive approach to single-image
  super-resolution,'' in \emph{Proceedings of the IEEE conference on computer
  vision and pattern recognition workshops}, 2018, pp. 864--873.

\bibitem{yu2018single}
L.~Yu, X.~Long, and C.~Tong, ``Single image super-resolution based on improved
  wgan,'' in \emph{Proceedings of the 2018 International Conference on Advanced
  Control, Automation and Artificial Intelligence (ACAAI 2018), Shenzhen,
  China}, 2018, pp. 21--22.

\bibitem{gulrajani2017improved}
I.~Gulrajani, F.~Ahmed, M.~Arjovsky, V.~Dumoulin, and A.~Courville, ``Improved
  training of wasserstein gans,'' \emph{arXiv preprint arXiv:1704.00028}, 2017.

\bibitem{lucic2017gans}
M.~Lucic, K.~Kurach, M.~Michalski, S.~Gelly, and O.~Bousquet, ``Are gans
  created equal? a large-scale study,'' \emph{arXiv preprint arXiv:1711.10337},
  2017.

\bibitem{singla2022review}
K.~Singla, R.~Pandey, and U.~Ghanekar, ``A review on single image super
  resolution techniques using generative adversarial network,'' \emph{Optik},
  p. 169607, 2022.

\bibitem{rudin1992nonlinear}
L.~I. Rudin, S.~Osher, and E.~Fatemi, ``Nonlinear total variation based noise
  removal algorithms,'' \emph{Physica D: nonlinear phenomena}, vol.~60, no.
  1-4, pp. 259--268, 1992.

\bibitem{vella2019single}
M.~Vella and J.~F. Mota, ``Single image super-resolution via cnn architectures
  and tv-tv minimization,'' \emph{arXiv preprint arXiv:1907.05380}, 2019.

\bibitem{portilla2000parametric}
J.~Portilla and E.~P. Simoncelli, ``A parametric texture model based on joint
  statistics of complex wavelet coefficients,'' \emph{International journal of
  computer vision}, vol.~40, no.~1, pp. 49--70, 2000.

\bibitem{gatys2015texture}
L.~Gatys, A.~S. Ecker, and M.~Bethge, ``Texture synthesis using convolutional
  neural networks,'' \emph{Advances in neural information processing systems},
  vol.~28, pp. 262--270, 2015.

\bibitem{ho2020denoising}
J.~Ho, A.~Jain, and P.~Abbeel, ``Denoising diffusion probabilistic models,''
  \emph{arXiv preprint arXiv:2006.11239}, 2020.

\bibitem{saharia2021image}
C.~Saharia, J.~Ho, W.~Chan, T.~Salimans, D.~J. Fleet, and M.~Norouzi, ``Image
  super-resolution via iterative refinement,'' \emph{arXiv preprint
  arXiv:2104.07636}, 2021.

\bibitem{li2019feedback}
Z.~Li, J.~Yang, Z.~Liu, X.~Yang, G.~Jeon, and W.~Wu, ``Feedback network for
  image super-resolution,'' in \emph{Proceedings of the IEEE/CVF Conference on
  Computer Vision and Pattern Recognition}, 2019, pp. 3867--3876.

\bibitem{tai2017image}
Y.~Tai, J.~Yang, and X.~Liu, ``Image super-resolution via deep recursive
  residual network,'' in \emph{Proceedings of the IEEE conference on computer
  vision and pattern recognition}, 2017, pp. 3147--3155.

\bibitem{shi2016real}
W.~Shi, J.~Caballero, F.~Husz{\'a}r, J.~Totz, A.~P. Aitken, R.~Bishop,
  D.~Rueckert, and Z.~Wang, ``Real-time single image and video super-resolution
  using an efficient sub-pixel convolutional neural network,'' in
  \emph{Proceedings of the IEEE conference on computer vision and pattern
  recognition}, 2016, pp. 1874--1883.

\bibitem{shi2016deconvolution}
W.~Shi, J.~Caballero, L.~Theis, F.~Huszar, A.~Aitken, C.~Ledig, and Z.~Wang,
  ``Is the deconvolution layer the same as a convolutional layer?'' \emph{arXiv
  preprint arXiv:1609.07009}, 2016.

\bibitem{wojna2019devil}
Z.~Wojna, V.~Ferrari, S.~Guadarrama, N.~Silberman, L.-C. Chen, A.~Fathi, and
  J.~Uijlings, ``The devil is in the decoder: Classification, regression and
  gans,'' \emph{International Journal of Computer Vision}, vol. 127, no.~11,
  pp. 1694--1706, 2019.

\bibitem{kundu2020attention}
S.~Kundu, H.~Mostafa, S.~N. Sridhar, and S.~Sundaresan, ``Attention-based image
  upsampling,'' \emph{arXiv preprint arXiv:2012.09904}, 2020.

\bibitem{jo2021practical}
Y.~Jo and S.~J. Kim, ``Practical single-image super-resolution using look-up
  table,'' in \emph{Proceedings of the IEEE/CVF Conference on Computer Vision
  and Pattern Recognition}, 2021, pp. 691--700.

\bibitem{peng2020saint}
C.~Peng, W.-A. Lin, H.~Liao, R.~Chellappa, and S.~K. Zhou, ``Saint: spatially
  aware interpolation network for medical slice synthesis,'' in
  \emph{Proceedings of the IEEE/CVF Conference on Computer Vision and Pattern
  Recognition}, 2020, pp. 7750--7759.

\bibitem{chai2022any}
L.~Chai, M.~Gharbi, E.~Shechtman, P.~Isola, and R.~Zhang, ``Any-resolution
  training for high-resolution image synthesis,'' \emph{arXiv preprint
  arXiv:2204.07156}, 2022.

\bibitem{vaswani2017attention}
A.~Vaswani, N.~Shazeer, N.~Parmar, J.~Uszkoreit, L.~Jones, A.~N. Gomez,
  {\L}.~Kaiser, and I.~Polosukhin, ``Attention is all you need,'' in
  \emph{Advances in neural information processing systems}, 2017, pp.
  5998--6008.

\bibitem{hu2018squeeze}
J.~Hu, L.~Shen, and G.~Sun, ``Squeeze-and-excitation networks,'' in
  \emph{Proceedings of the IEEE conference on computer vision and pattern
  recognition}, 2018, pp. 7132--7141.

\bibitem{yang2021image}
Y.~Yang and Y.~Qi, ``Image super-resolution via channel attention and spatial
  graph convolutional network,'' \emph{Pattern Recognition}, vol. 112, p.
  107798, 2021.

\bibitem{wang2018non}
X.~Wang, R.~Girshick, A.~Gupta, and K.~He, ``Non-local neural networks,'' in
  \emph{Proceedings of the IEEE conference on computer vision and pattern
  recognition}, 2018, pp. 7794--7803.

\bibitem{zhang2019residual}
Y.~Zhang, K.~Li, K.~Li, B.~Zhong, and Y.~Fu, ``Residual non-local attention
  networks for image restoration,'' \emph{arXiv preprint arXiv:1903.10082},
  2019.

\bibitem{liang2021swinir}
J.~Liang, J.~Cao, G.~Sun, K.~Zhang, L.~Van~Gool, and R.~Timofte, ``Swinir:
  Image restoration using swin transformer,'' in \emph{Proceedings of the
  IEEE/CVF International Conference on Computer Vision}, 2021, pp. 1833--1844.

\bibitem{zhang2022swinfir}
D.~Zhang, F.~Huang, S.~Liu, X.~Wang, and Z.~Jin, ``Swinfir: Revisiting the
  swinir with fast fourier convolution and improved training for image
  super-resolution,'' \emph{arXiv preprint arXiv:2208.11247}, 2022.

\bibitem{dai2019second}
T.~Dai, J.~Cai, Y.~Zhang, S.-T. Xia, and L.~Zhang, ``Second-order attention
  network for single image super-resolution,'' in \emph{Proceedings of the
  IEEE/CVF Conference on Computer Vision and Pattern Recognition}, 2019, pp.
  11\,065--11\,074.

\bibitem{zhao2020efficient}
H.~Zhao, X.~Kong, J.~He, Y.~Qiao, and C.~Dong, ``Efficient image
  super-resolution using pixel attention,'' in \emph{European Conference on
  Computer Vision}.\hskip 1em plus 0.5em minus 0.4em\relax Springer, 2020, pp.
  56--72.

\bibitem{niu2020single}
B.~Niu, W.~Wen, W.~Ren, X.~Zhang, L.~Yang, S.~Wang, K.~Zhang, X.~Cao, and
  H.~Shen, ``Single image super-resolution via a holistic attention network,''
  in \emph{European Conference on Computer Vision}.\hskip 1em plus 0.5em minus
  0.4em\relax Springer, 2020, pp. 191--207.

\bibitem{wang2021bam}
F.~Wang, H.~Hu, and C.~Shen, ``Bam: A lightweight and efficient balanced
  attention mechanism for single image super resolution,'' \emph{arXiv preprint
  arXiv:2104.07566}, 2021.

\bibitem{bengio2009curriculum}
Y.~Bengio, J.~Louradour, R.~Collobert, and J.~Weston, ``Curriculum learning,''
  in \emph{Proceedings of the 26th annual international conference on machine
  learning}, 2009, pp. 41--48.

\bibitem{bei2018new}
Y.~Bei, A.~Damian, S.~Hu, S.~Menon, N.~Ravi, and C.~Rudin, ``New techniques for
  preserving global structure and denoising with low information loss in
  single-image super-resolution,'' in \emph{Proceedings of the IEEE Conference
  on Computer Vision and Pattern Recognition Workshops}, 2018, pp. 874--881.

\bibitem{ahn2018image}
N.~Ahn, B.~Kang, and K.-A. Sohn, ``Image super-resolution via progressive
  cascading residual network,'' in \emph{Proceedings of the IEEE Conference on
  Computer Vision and Pattern Recognition Workshops}, 2018, pp. 791--799.

\bibitem{shocher2018zero}
A.~Shocher, N.~Cohen, and M.~Irani, ``“zero-shot” super-resolution using
  deep internal learning,'' in \emph{Proceedings of the IEEE conference on
  computer vision and pattern recognition}, 2018, pp. 3118--3126.

\bibitem{ren2017image}
H.~Ren, M.~El-Khamy, and J.~Lee, ``Image super resolution based on fusing
  multiple convolution neural networks,'' in \emph{Proceedings of the IEEE
  Conference on Computer Vision and Pattern Recognition Workshops}, 2017, pp.
  54--61.

\bibitem{urazoe2021multi}
K.~Urazoe, N.~Kuroki, Y.~Kato, S.~Ohtani, T.~Hirose, and M.~Numa,
  ``Multi-category image super-resolution with convolutional neural network and
  multi-task learning,'' \emph{IEICE TRANSACTIONS on Information and Systems},
  vol. 104, no.~1, pp. 183--193, 2021.

\bibitem{shi2017structure}
Y.~Shi, K.~Wang, C.~Chen, L.~Xu, and L.~Lin, ``Structure-preserving image
  super-resolution via contextualized multitask learning,'' \emph{IEEE
  transactions on multimedia}, vol.~19, no.~12, pp. 2804--2815, 2017.

\bibitem{ioffe2015batch}
S.~Ioffe and C.~Szegedy, ``Batch normalization: Accelerating deep network
  training by reducing internal covariate shift,'' in \emph{International
  conference on machine learning}.\hskip 1em plus 0.5em minus 0.4em\relax PMLR,
  2015, pp. 448--456.

\bibitem{tai2017memnet}
Y.~Tai, J.~Yang, X.~Liu, and C.~Xu, ``Memnet: A persistent memory network for
  image restoration,'' in \emph{Proceedings of the IEEE international
  conference on computer vision}, 2017, pp. 4539--4547.

\bibitem{nah2017deep}
S.~Nah, T.~Hyun~Kim, and K.~Mu~Lee, ``Deep multi-scale convolutional neural
  network for dynamic scene deblurring,'' in \emph{Proceedings of the IEEE
  conference on computer vision and pattern recognition}, 2017, pp. 3883--3891.

\bibitem{liu2021adadm}
J.~Liu, J.~Tang, and G.~Wu, ``Adadm: Enabling normalization for image
  super-resolution,'' \emph{arXiv preprint arXiv:2111.13905}, 2021.

\bibitem{irani1991improving}
M.~Irani and S.~Peleg, ``Improving resolution by image registration,''
  \emph{CVGIP: Graphical models and image processing}, vol.~53, no.~3, pp.
  231--239, 1991.

\bibitem{tan2022image}
C.~Tan, L.~Wang, and S.~Cheng, ``Image super-resolution via dual-level
  recurrent residual networks,'' \emph{Sensors}, vol.~22, no.~8, p. 3058, 2022.

\bibitem{han2018image}
W.~Han, S.~Chang, D.~Liu, M.~Yu, M.~Witbrock, and T.~S. Huang, ``Image
  super-resolution via dual-state recurrent networks,'' in \emph{Proceedings of
  the IEEE conference on computer vision and pattern recognition}, 2018, pp.
  1654--1663.

\bibitem{kim2016accurate}
J.~Kim, J.~K. Lee, and K.~M. Lee, ``Accurate image super-resolution using very
  deep convolutional networks,'' in \emph{Proceedings of the IEEE conference on
  computer vision and pattern recognition}, 2016, pp. 1646--1654.

\bibitem{he2015convolutional}
K.~He and J.~Sun, ``Convolutional neural networks at constrained time cost,''
  in \emph{Proceedings of the IEEE conference on computer vision and pattern
  recognition}, 2015, pp. 5353--5360.

\bibitem{ghiasi2016laplacian}
G.~Ghiasi and C.~C. Fowlkes, ``Laplacian pyramid reconstruction and refinement
  for semantic segmentation,'' in \emph{European conference on computer
  vision}.\hskip 1em plus 0.5em minus 0.4em\relax Springer, 2016, pp. 519--534.

\bibitem{mao2016image}
X.~Mao, C.~Shen, and Y.-B. Yang, ``Image restoration using very deep
  convolutional encoder-decoder networks with symmetric skip connections,''
  \emph{Advances in neural information processing systems}, vol.~29, pp.
  2802--2810, 2016.

\bibitem{ronneberger2015u}
O.~Ronneberger, P.~Fischer, and T.~Brox, ``U-net: Convolutional networks for
  biomedical image segmentation,'' in \emph{International Conference on Medical
  image computing and computer-assisted intervention}.\hskip 1em plus 0.5em
  minus 0.4em\relax Springer, 2015, pp. 234--241.

\bibitem{he2016deep}
K.~He, X.~Zhang, S.~Ren, and J.~Sun, ``Deep residual learning for image
  recognition,'' in \emph{Proceedings of the IEEE conference on computer vision
  and pattern recognition}, 2016, pp. 770--778.

\bibitem{huang2017densely}
G.~Huang, Z.~Liu, L.~Van Der~Maaten, and K.~Q. Weinberger, ``Densely connected
  convolutional networks,'' in \emph{Proceedings of the IEEE conference on
  computer vision and pattern recognition}, 2017, pp. 4700--4708.

\bibitem{kim2016deeply}
J.~Kim, J.~K. Lee, and K.~M. Lee, ``Deeply-recursive convolutional network for
  image super-resolution,'' in \emph{Proceedings of the IEEE conference on
  computer vision and pattern recognition}, 2016, pp. 1637--1645.

\bibitem{hochreiter1997long}
S.~Hochreiter and J.~Schmidhuber, ``Long short-term memory,'' \emph{Neural
  computation}, vol.~9, no.~8, pp. 1735--1780, 1997.

\bibitem{chung2015gated}
J.~Chung, C.~Gulcehre, K.~Cho, and Y.~Bengio, ``Gated feedback recurrent neural
  networks,'' in \emph{International conference on machine learning}.\hskip 1em
  plus 0.5em minus 0.4em\relax PMLR, 2015, pp. 2067--2075.

\bibitem{zamir2017feedback}
A.~R. Zamir, T.-L. Wu, L.~Sun, W.~B. Shen, B.~E. Shi, J.~Malik, and
  S.~Savarese, ``Feedback networks,'' in \emph{Proceedings of the IEEE
  conference on computer vision and pattern recognition}, 2017, pp. 1308--1317.

\bibitem{liu2018non}
D.~Liu, B.~Wen, Y.~Fan, C.~C. Loy, and T.~S. Huang, ``Non-local recurrent
  network for image restoration,'' \emph{arXiv preprint arXiv:1806.02919},
  2018.

\bibitem{isobe2020video}
T.~Isobe, X.~Jia, S.~Gu, S.~Li, S.~Wang, and Q.~Tian, ``Video super-resolution
  with recurrent structure-detail network,'' in \emph{European Conference on
  Computer Vision}.\hskip 1em plus 0.5em minus 0.4em\relax Springer, 2020, pp.
  645--660.

\bibitem{park2020fast}
S.~Park, J.~Yoo, D.~Cho, J.~Kim, and T.~H. Kim, ``Fast adaptation to
  super-resolution networks via meta-learning,'' in \emph{Computer Vision--ECCV
  2020: 16th European Conference, Glasgow, UK, August 23--28, 2020,
  Proceedings, Part XXVII 16}.\hskip 1em plus 0.5em minus 0.4em\relax Springer,
  2020, pp. 754--769.

\bibitem{ahn2018fast}
N.~Ahn, B.~Kang, and K.-A. Sohn, ``Fast, accurate, and lightweight
  super-resolution with cascading residual network,'' in \emph{Proceedings of
  the European Conference on Computer Vision (ECCV)}, 2018, pp. 252--268.

\bibitem{howard2017mobilenets}
A.~G. Howard, M.~Zhu, B.~Chen, D.~Kalenichenko, W.~Wang, T.~Weyand,
  M.~Andreetto, and H.~Adam, ``Mobilenets: Efficient convolutional neural
  networks for mobile vision applications,'' \emph{arXiv preprint
  arXiv:1704.04861}, 2017.

\bibitem{liu2020residual}
J.~Liu, J.~Tang, and G.~Wu, ``Residual feature distillation network for
  lightweight image super-resolution,'' in \emph{European Conference on
  Computer Vision}.\hskip 1em plus 0.5em minus 0.4em\relax Springer, 2020, pp.
  41--55.

\bibitem{hui2019lightweight}
Z.~Hui, X.~Gao, Y.~Yang, and X.~Wang, ``Lightweight image super-resolution with
  information multi-distillation network,'' in \emph{Proceedings of the 27th
  ACM International Conference on Multimedia}, 2019, pp. 2024--2032.

\bibitem{ayazoglu2021extremely}
M.~Ayazoglu, ``Extremely lightweight quantization robust real-time single-image
  super resolution for mobile devices,'' in \emph{Proceedings of the IEEE/CVF
  Conference on Computer Vision and Pattern Recognition}, 2021, pp. 2472--2479.

\bibitem{hubara2017quantized}
I.~Hubara, M.~Courbariaux, D.~Soudry, R.~El-Yaniv, and Y.~Bengio, ``Quantized
  neural networks: Training neural networks with low precision weights and
  activations,'' \emph{The Journal of Machine Learning Research}, vol.~18,
  no.~1, pp. 6869--6898, 2017.

\bibitem{sarker2021deep}
I.~H. Sarker, ``Deep learning: a comprehensive overview on techniques,
  taxonomy, applications and research directions,'' \emph{SN Computer Science},
  vol.~2, no.~6, pp. 1--20, 2021.

\bibitem{stephane1999wavelet}
M.~Stephane, ``A wavelet tour of signal processing,'' 1999.

\bibitem{guo2017deep}
T.~Guo, H.~Seyed~Mousavi, T.~Huu~Vu, and V.~Monga, ``Deep wavelet prediction
  for image super-resolution,'' in \emph{Proceedings of the IEEE Conference on
  Computer Vision and Pattern Recognition Workshops}, 2017, pp. 104--113.

\bibitem{sahito2019wavelet}
F.~Sahito, P.~Zhiwen, J.~Ahmed, and R.~A. Memon, ``Wavelet-integrated deep
  networks for single image super-resolution,'' \emph{Electronics}, vol.~8,
  no.~5, p. 553, 2019.

\bibitem{huang2017wavelet}
H.~Huang, R.~He, Z.~Sun, and T.~Tan, ``Wavelet-srnet: A wavelet-based cnn for
  multi-scale face super resolution,'' in \emph{Proceedings of the IEEE
  International Conference on Computer Vision}, 2017, pp. 1689--1697.

\bibitem{liu2018multi}
P.~Liu, H.~Zhang, K.~Zhang, L.~Lin, and W.~Zuo, ``Multi-level wavelet-cnn for
  image restoration,'' in \emph{Proceedings of the IEEE conference on computer
  vision and pattern recognition workshops}, 2018, pp. 773--782.

\bibitem{zhang2019super}
Q.~Zhang, H.~Wang, T.~Du, S.~Yang, Y.~Wang, Z.~Xing, W.~Bai, and Y.~Yi,
  ``Super-resolution reconstruction algorithms based on fusion of deep learning
  mechanism and wavelet,'' in \emph{Proceedings of the 2nd International
  Conference on Artificial Intelligence and Pattern Recognition}, 2019, pp.
  102--107.

\bibitem{xue2020wavelet}
S.~Xue, W.~Qiu, F.~Liu, and X.~Jin, ``Wavelet-based residual attention network
  for image super-resolution,'' \emph{Neurocomputing}, vol. 382, pp. 116--126,
  2020.

\bibitem{zhu2021video}
X.~Zhu, Z.~Li, J.~Lou, and Q.~Shen, ``Video super-resolution based on a
  spatio-temporal matching network,'' \emph{Pattern Recognition}, vol. 110, p.
  107619, 2021.

\bibitem{liu2022blind}
A.~Liu, Y.~Liu, J.~Gu, Y.~Qiao, and C.~Dong, ``Blind image super-resolution: A
  survey and beyond,'' \emph{IEEE Transactions on Pattern Analysis and Machine
  Intelligence}, 2022.

\bibitem{ignatov2018wespe}
A.~Ignatov, N.~Kobyshev, R.~Timofte, K.~Vanhoey, and L.~Van~Gool, ``Wespe:
  weakly supervised photo enhancer for digital cameras,'' in \emph{Proceedings
  of the IEEE Conference on Computer Vision and Pattern Recognition Workshops},
  2018, pp. 691--700.

\bibitem{zhu2017unpaired}
J.-Y. Zhu, T.~Park, P.~Isola, and A.~A. Efros, ``Unpaired image-to-image
  translation using cycle-consistent adversarial networks,'' in
  \emph{Proceedings of the IEEE international conference on computer vision},
  2017, pp. 2223--2232.

\bibitem{bulat2018learn}
A.~Bulat, J.~Yang, and G.~Tzimiropoulos, ``To learn image super-resolution, use
  a gan to learn how to do image degradation first,'' in \emph{Proceedings of
  the European conference on computer vision (ECCV)}, 2018, pp. 185--200.

\bibitem{ulyanov2018deep}
D.~Ulyanov, A.~Vedaldi, and V.~Lempitsky, ``Deep image prior,'' in
  \emph{Proceedings of the IEEE conference on computer vision and pattern
  recognition}, 2018, pp. 9446--9454.

\bibitem{xian2018zero}
Y.~Xian, C.~H. Lampert, B.~Schiele, and Z.~Akata, ``Zero-shot learning—a
  comprehensive evaluation of the good, the bad and the ugly,'' \emph{IEEE
  transactions on pattern analysis and machine intelligence}, vol.~41, no.~9,
  pp. 2251--2265, 2018.

\bibitem{cheng2020zero}
X.~Cheng, Z.~Fu, and J.~Yang, ``Zero-shot image super-resolution with depth
  guided internal degradation learning,'' in \emph{European Conference on
  Computer Vision}.\hskip 1em plus 0.5em minus 0.4em\relax Springer, 2020, pp.
  265--280.

\bibitem{godard2019digging}
C.~Godard, O.~Mac~Aodha, M.~Firman, and G.~J. Brostow, ``Digging into
  self-supervised monocular depth estimation,'' in \emph{Proceedings of the
  IEEE/CVF International Conference on Computer Vision}, 2019, pp. 3828--3838.

\bibitem{soh2020meta}
J.~W. Soh, S.~Cho, and N.~I. Cho, ``Meta-transfer learning for zero-shot
  super-resolution,'' in \emph{Proceedings of the IEEE/CVF Conference on
  Computer Vision and Pattern Recognition}, 2020, pp. 3516--3525.

\bibitem{lai2018fast}
W.-S. Lai, J.-B. Huang, N.~Ahuja, and M.-H. Yang, ``Fast and accurate image
  super-resolution with deep laplacian pyramid networks,'' \emph{IEEE
  transactions on pattern analysis and machine intelligence}, vol.~41, no.~11,
  pp. 2599--2613, 2018.

\bibitem{song2020efficient}
D.~Song, C.~Xu, X.~Jia, Y.~Chen, C.~Xu, and Y.~Wang, ``Efficient residual dense
  block search for image super-resolution,'' in \emph{Proceedings of the AAAI
  Conference on Artificial Intelligence}, vol.~34, no.~07, 2020, pp.
  12\,007--12\,014.

\bibitem{lu2019nsga}
Z.~Lu, I.~Whalen, V.~Boddeti, Y.~Dhebar, K.~Deb, E.~Goodman, and W.~Banzhaf,
  ``Nsga-net: neural architecture search using multi-objective genetic
  algorithm,'' in \emph{Proceedings of the Genetic and Evolutionary Computation
  Conference}, 2019, pp. 419--427.

\bibitem{1599245}
P.~Ngatchou, A.~Zarei, and A.~El-Sharkawi, ``Pareto multi objective
  optimization,'' in \emph{Proceedings of the 13th International Conference on,
  Intelligent Systems Application to Power Systems}, 2005, pp. 84--91.

\bibitem{chu2021fast}
X.~Chu, B.~Zhang, H.~Ma, R.~Xu, and Q.~Li, ``Fast, accurate and lightweight
  super-resolution with neural architecture search,'' in \emph{2020 25th
  International Conference on Pattern Recognition (ICPR)}.\hskip 1em plus 0.5em
  minus 0.4em\relax IEEE, 2021, pp. 59--64.

\bibitem{ahn2021neural}
J.~Y. Ahn and N.~I. Cho, ``Neural architecture search for image
  super-resolution using densely constructed search space: Deconas,'' in
  \emph{2020 25th International Conference on Pattern Recognition
  (ICPR)}.\hskip 1em plus 0.5em minus 0.4em\relax IEEE, 2021, pp. 4829--4836.

\bibitem{pham2018efficient}
H.~Pham, M.~Guan, B.~Zoph, Q.~Le, and J.~Dean, ``Efficient neural architecture
  search via parameters sharing,'' in \emph{International Conference on Machine
  Learning}.\hskip 1em plus 0.5em minus 0.4em\relax PMLR, 2018, pp. 4095--4104.

\bibitem{guo2020hierarchical}
Y.~Guo, Y.~Luo, Z.~He, J.~Huang, and J.~Chen, ``Hierarchical neural
  architecture search for single image super-resolution,'' \emph{IEEE Signal
  Processing Letters}, vol.~27, pp. 1255--1259, 2020.

\bibitem{chen2020dip}
Y.-C. Chen, C.~Gao, E.~Robb, and J.-B. Huang, ``Nas-dip: Learning deep image
  prior with neural architecture search,'' in \emph{Computer Vision--ECCV 2020:
  16th European Conference, Glasgow, UK, August 23--28, 2020, Proceedings, Part
  XVIII 16}.\hskip 1em plus 0.5em minus 0.4em\relax Springer, 2020, pp.
  442--459.

\bibitem{lee2020journey}
R.~Lee, {\L}.~Dudziak, M.~Abdelfattah, S.~I. Venieris, H.~Kim, H.~Wen, and
  N.~D. Lane, ``Journey towards tiny perceptual super-resolution,'' in
  \emph{European Conference on Computer Vision}.\hskip 1em plus 0.5em minus
  0.4em\relax Springer, 2020, pp. 85--102.

\bibitem{liu2018darts}
H.~Liu, K.~Simonyan, and Y.~Yang, ``Darts: Differentiable architecture
  search,'' \emph{arXiv preprint arXiv:1806.09055}, 2018.

\bibitem{zhang2020memory}
H.~Zhang, Y.~Li, C.~Gong, H.~Chen, Z.~Bai, and C.~Shen, ``Memory-efficient
  hierarchical neural architecture search for image restoration,'' \emph{arXiv
  preprint arXiv:2012.13212}, 2020.

\bibitem{wu2021trilevel}
Y.~Wu, Z.~Huang, S.~Kumar, R.~S. Sukthanker, R.~Timofte, and L.~Van~Gool,
  ``Trilevel neural architecture search for efficient single image
  super-resolution,'' \emph{arXiv preprint arXiv:2101.06658}, 2021.

\bibitem{liu2021swin}
Z.~Liu, Y.~Lin, Y.~Cao, H.~Hu, Y.~Wei, Z.~Zhang, S.~Lin, and B.~Guo, ``Swin
  transformer: Hierarchical vision transformer using shifted windows,'' in
  \emph{Proceedings of the IEEE/CVF International Conference on Computer
  Vision}, 2021, pp. 10\,012--10\,022.

\bibitem{cai2019toward}
J.~Cai, H.~Zeng, H.~Yong, Z.~Cao, and L.~Zhang, ``Toward real-world single
  image super-resolution: A new benchmark and a new model,'' in
  \emph{Proceedings of the IEEE/CVF International Conference on Computer
  Vision}, 2019, pp. 3086--3095.

\bibitem{yang2021real}
X.~Yang, W.~Xiang, H.~Zeng, and L.~Zhang, ``Real-world video super-resolution:
  A benchmark dataset and a decomposition based learning scheme,'' in
  \emph{Proceedings of the IEEE/CVF International Conference on Computer
  Vision}, 2021, pp. 4781--4790.

\bibitem{yang2010image}
J.~Yang, J.~Wright, T.~S. Huang, and Y.~Ma, ``Image super-resolution via sparse
  representation,'' \emph{IEEE transactions on image processing}, vol.~19,
  no.~11, pp. 2861--2873, 2010.

\bibitem{huang2015single}
J.-B. Huang, A.~Singh, and N.~Ahuja, ``Single image super-resolution from
  transformed self-exemplars,'' in \emph{Proceedings of the IEEE conference on
  computer vision and pattern recognition}, 2015, pp. 5197--5206.

\bibitem{karras2017progressive}
T.~Karras, T.~Aila, S.~Laine, and J.~Lehtinen, ``Progressive growing of gans
  for improved quality, stability, and variation,'' \emph{arXiv preprint
  arXiv:1710.10196}, 2017.

\bibitem{karras2019style}
T.~Karras, S.~Laine, and T.~Aila, ``A style-based generator architecture for
  generative adversarial networks,'' in \emph{Proceedings of the IEEE/CVF
  Conference on Computer Vision and Pattern Recognition}, 2019, pp. 4401--4410.

\bibitem{wang2019flickr1024}
Y.~Wang, L.~Wang, J.~Yang, W.~An, and Y.~Guo, ``Flickr1024: A large-scale
  dataset for stereo image super-resolution,'' in \emph{Proceedings of the
  IEEE/CVF International Conference on Computer Vision Workshops}, 2019, pp.
  0--0.

\bibitem{agustsson2017ntire}
E.~Agustsson and R.~Timofte, ``Ntire 2017 challenge on single image
  super-resolution: Dataset and study,'' in \emph{Proceedings of the IEEE
  conference on computer vision and pattern recognition workshops}, 2017, pp.
  126--135.

\bibitem{zhang2021benchmarking}
K.~Zhang, D.~Li, W.~Luo, W.~Ren, B.~Stenger, W.~Liu, H.~Li, and M.-H. Yang,
  ``Benchmarking ultra-high-definition image super-resolution,'' in
  \emph{Proceedings of the IEEE/CVF International Conference on Computer
  Vision}, 2021, pp. 14\,769--14\,778.

\bibitem{gu2019div8k}
S.~Gu, A.~Lugmayr, M.~Danelljan, M.~Fritsche, J.~Lamour, and R.~Timofte,
  ``Div8k: Diverse 8k resolution image dataset,'' in \emph{2019 IEEE/CVF
  International Conference on Computer Vision Workshop (ICCVW)}.\hskip 1em plus
  0.5em minus 0.4em\relax IEEE, 2019, pp. 3512--3516.

\bibitem{arbelaez2010contour}
P.~Arbelaez, M.~Maire, C.~Fowlkes, and J.~Malik, ``Contour detection and
  hierarchical image segmentation,'' \emph{IEEE transactions on pattern
  analysis and machine intelligence}, vol.~33, no.~5, pp. 898--916, 2010.

\bibitem{geiger2012we}
A.~Geiger, P.~Lenz, and R.~Urtasun, ``Are we ready for autonomous driving? the
  kitti vision benchmark suite,'' in \emph{2012 IEEE conference on computer
  vision and pattern recognition}.\hskip 1em plus 0.5em minus 0.4em\relax IEEE,
  2012, pp. 3354--3361.

\bibitem{yi2014learning}
D.~Yi, Z.~Lei, S.~Liao, and S.~Z. Li, ``Learning face representation from
  scratch,'' \emph{arXiv preprint arXiv:1411.7923}, 2014.

\bibitem{liu2015deep}
Z.~Liu, P.~Luo, X.~Wang, and X.~Tang, ``Deep learning face attributes in the
  wild,'' in \emph{Proceedings of the IEEE international conference on computer
  vision}, 2015, pp. 3730--3738.

\bibitem{cao2018vggface2}
Q.~Cao, L.~Shen, W.~Xie, O.~M. Parkhi, and A.~Zisserman, ``Vggface2: A dataset
  for recognising faces across pose and age,'' in \emph{2018 13th IEEE
  international conference on automatic face \& gesture recognition (FG
  2018)}.\hskip 1em plus 0.5em minus 0.4em\relax IEEE, 2018, pp. 67--74.

\bibitem{wang2018recovering}
X.~Wang, K.~Yu, C.~Dong, and C.~C. Loy, ``Recovering realistic texture in image
  super-resolution by deep spatial feature transform,'' in \emph{Proceedings of
  the IEEE conference on computer vision and pattern recognition}, 2018, pp.
  606--615.

\bibitem{Shen_2019_ICCV}
Z.~Shen, W.~Wang, X.~Lu, J.~Shen, H.~Ling, T.~Xu, and L.~Shao, ``Human-aware
  motion deblurring,'' in \emph{Proceedings of the IEEE/CVF International
  Conference on Computer Vision (ICCV)}, October 2019.

\bibitem{yang2021implicit}
J.~Yang, S.~Shen, H.~Yue, and K.~Li, ``Implicit transformer network for screen
  content image continuous super-resolution,'' \emph{Advances in Neural
  Information Processing Systems}, vol.~34, 2021.

\bibitem{liu2013bayesian}
C.~Liu and D.~Sun, ``On bayesian adaptive video super resolution,'' \emph{IEEE
  transactions on pattern analysis and machine intelligence}, vol.~36, no.~2,
  pp. 346--360, 2013.

\bibitem{lin2015mcl}
J.~Y. Lin, R.~Song, C.-H. Wu, T.~Liu, H.~Wang, and C.-C.~J. Kuo, ``Mcl-v: A
  streaming video quality assessment database,'' \emph{Journal of Visual
  Communication and Image Representation}, vol.~30, pp. 1--9, 2015.

\bibitem{Nah_2017_CVPR}
S.~Nah, T.~H. Kim, and K.~M. Lee, ``Deep multi-scale convolutional neural
  network for dynamic scene deblurring,'' in \emph{The IEEE Conference on
  Computer Vision and Pattern Recognition (CVPR)}, July 2017.

\bibitem{xue2019video}
T.~Xue, B.~Chen, J.~Wu, D.~Wei, and W.~T. Freeman, ``Video enhancement with
  task-oriented flow,'' \emph{International Journal of Computer Vision (IJCV)},
  vol. 127, no.~8, pp. 1106--1125, 2019.

\bibitem{odena2016deconvolution}
\BIBentryALTinterwordspacing
A.~Odena, V.~Dumoulin, and C.~Olah, ``Deconvolution and checkerboard
  artifacts,'' \emph{Distill}, 2016. [Online]. Available:
  \url{http://distill.pub/2016/deconv-checkerboard}
\BIBentrySTDinterwordspacing

\bibitem{kopf2007joint}
J.~Kopf, M.~F. Cohen, D.~Lischinski, and M.~Uyttendaele, ``Joint bilateral
  upsampling,'' \emph{ACM Transactions on Graphics (ToG)}, vol.~26, no.~3, pp.
  96--es, 2007.

\bibitem{chen2016bilateral}
J.~Chen, A.~Adams, N.~Wadhwa, and S.~W. Hasinoff, ``Bilateral guided
  upsampling,'' \emph{ACM Transactions on Graphics (TOG)}, vol.~35, no.~6, pp.
  1--8, 2016.

\bibitem{wang2018esrgan}
X.~Wang, K.~Yu, S.~Wu, J.~Gu, Y.~Liu, C.~Dong, Y.~Qiao, and C.~Change~Loy,
  ``Esrgan: Enhanced super-resolution generative adversarial networks,'' in
  \emph{Proceedings of the European conference on computer vision (ECCV)
  workshops}, 2018, pp. 0--0.

\end{thebibliography}

\section*{Appendix}
\renewcommand*{\thesection}{\alph{section}.}

\IEEEPARstart{I}{n} the main text, we complemented previous surveys by critically identifying current strategies and new research areas. The supplementary material hereby gives further information and visualizations on the topics discussed. It supports understanding the main concepts and ideas examined in the main text.

\section*{The SISR problem setting}
\begin{figure}[!h]
    \begin{center}
        \includegraphics[width=0.49\textwidth]{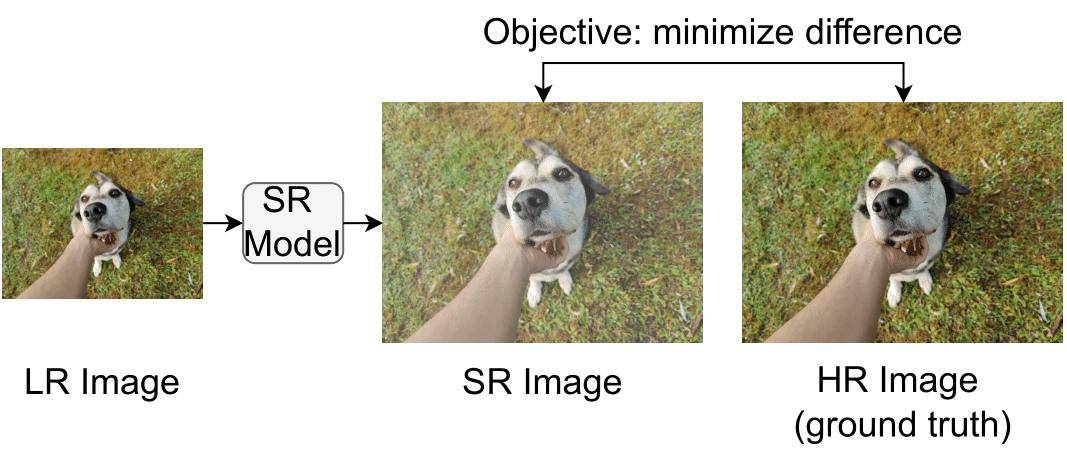}
        \caption{\label{fig:sr_framework}
        The SISR setting. The task is to super-resolve with a SR model a given LR image. The resulting SR image is then compared with the target HR image. The SR model learns to minimize the difference.}
    \end{center}
\end{figure}

The Single Image Super-Resolution (SISR) setting is about scaling up a given Low-Resolution (LR) image to a High-Resolution (HR) image. Typically, a SISR approach optimizes an objective that minimizes the difference between an estimated Super-Resolution (SR) image and the given ground-truth image as shown in \autoref{fig:sr_framework}.

\section*{About Datasets and their Limitations}
As mentioned in the paper, there is a vast amount of datasets available that meet different requirements. 
\autoref{tab:datasets} lists commonly used datasets.
Note that some datasets contain only single-resolution images. 
Thus, most researchers synthetically produce LR-HR images by considering the original images as HR images and by generating LR images with techniques like bicubic interpolation and anti-aliasing \cite{MATLAB:2017b}. 
However, the generation of LR images significantly affects the performance of the trained SR model, which is often observed when the model is applied to real-world scenarios.
In other words, LR images deviating from the assumed data distributions cause a trained SR model inevitably to produce much less pleasing results.
In the latter case, the reader should recall unsupervised SR techniques shown in this work that explicitly want to generalize the degradation process, which is often unknown in practice.
Current SR datasets that can meet real-world degradation (e.g., RealSR \cite{cai2019toward}) are sparse, rare, and some still synthesize LR inputs from HR images (e.g., RealVSR \cite{yang2021real}). 
We also refer to the survey of Liu et al. \cite{liu2022blind} that exclusively examines unsupervised SR in more detail.
In contrast to Liu et al., this work aims to cast a broader net in DL-based SR overall. 

\section*{Visualizations for Learning Objectives}
\begin{figure}[!h]
    \begin{center}
        \includegraphics[width=.47\textwidth]{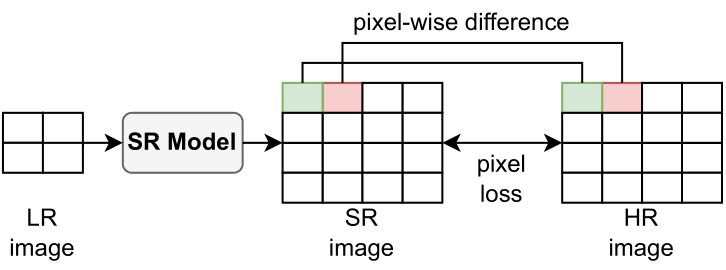}
        \caption{\label{fig:pixel_loss}
        Pixel loss concept. The SR model constructs the SR image and trains to minimize the pixel loss function based on the difference between the SR and the HR image. Prominent representatives of pixel loss are the $L1$-, $L2$-, and the Charbonnier-loss. Pixel loss functions favor a high PSNR because both formulations use pixel differences. In literature, $L1$-loss is favored chiefly.
        }
    \end{center}
\end{figure}

\begin{figure}[!h]
    \begin{center}
        \includegraphics[width=.47\textwidth]{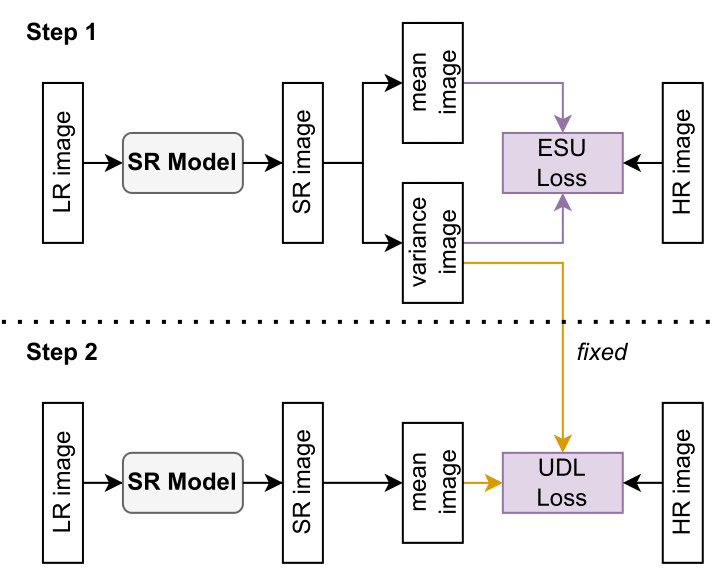}
        \caption{\label{fig:udl_loss}
        UDL based learning objective \cite{ning2021uncertainty}. It is a Dual Network approach with two steps. (1) The SR model approximates the HR image by learning a mean and a variance using the ESU loss. (2) The SR model approximates the HR image by predicting a mean and the fixed variance value learned from step (1) and using the UDL loss.
        }
    \end{center}
\end{figure}

\begin{figure}[!h]
    \begin{center}
        \includegraphics[width=.48\textwidth]{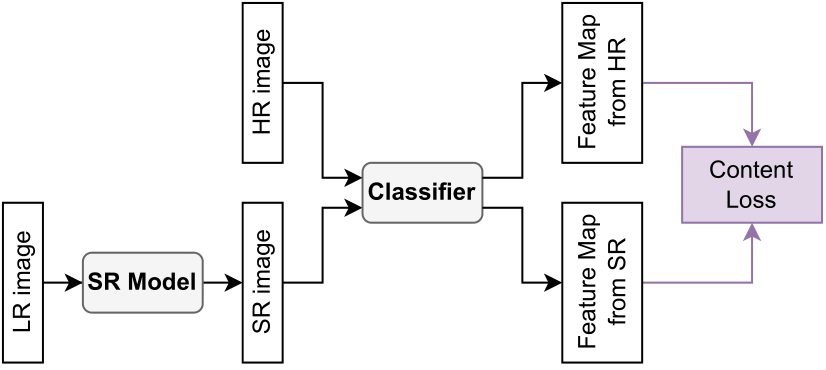}
        \caption{\label{fig:content_loss}
        Content loss based learning objective. A separate network, e.g. an image classifier CNN, is used to extract feature maps, which are then used for content loss. It determines the difference between the feature maps extracted from the approximated and the ground-truth HR image.
        }
    \end{center}
\end{figure}

\begin{figure}[!h]
    \begin{center}
        \includegraphics[width=.49\textwidth]{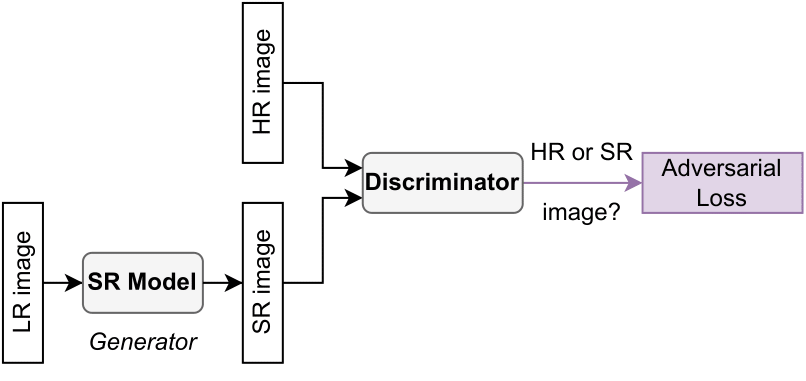}
        \caption{\label{fig:gan_loss}
        GAN based learning objective \cite{ning2021uncertainty}. It consists of a generator (SR model) and a discriminator network. The generator approximates HR images while the discriminator distinguishes between the HR and the approximated image. The generator's goal is to fool the discriminator.
        }
    \end{center}
\end{figure}

This section contains various visualizations that address the learning objectives discussed in the main text. \autoref{fig:pixel_loss} shows the pixel loss concept, which is mainly embodied with the $L1$-, $L2$-, and the Charbonnier-loss. \autoref{fig:udl_loss} shows the uncertainty-driven loss function, which incorporates variance. \autoref{fig:content_loss} shows the content loss, which is often used as a regularization term for GANs that optimize the adversarial loss visualized in \autoref{fig:gan_loss}.

\begin{table*}
\begin{center}
    \caption{\label{tab:datasets}Benchmark datasets for Super-Resolution (SISR and MISR). "S" denotes that the LR images are generated synthetically.}
    \begin{tabular}{| l l r c c l |}
    \hline
    Purpose & Dataset & Amount of & Year of & Type & Degradation\\
     &  name & images & release &&\\
    \hline
    \multirow{3}{*}{Classic SR Training} & BSDS200 \cite{martin2001database} & 500 & 2001 & natural&S: bicubic downscaling\\
    & T91 \cite{yang2010image} & 91 & 2010& flowers& only HR images\\
    & General-100 \cite{dong2016accelerating} & 100  & 2016& natural& only HR images\\
    \hline
    \multirow{5}{*}{Classic SR Testing} & BSDS100 \cite{martin2001database} & 100  & 2001& natural&S: bicubic downscaling\\
    & Set14 \cite{zeyde2010single} & 14 & 2010& natural &S: bicubic downscaling\\
    & Set5 \cite{bevilacqua2012low} & 5 & 2012& natural &S: bicubic downscaling\\
    & Urban100 \cite{huang2015single} & 100  & 2015&urban scenes&S: bicubic downscaling\\
    & Manga109 \cite{matsui2017sketch} & 109  & 2017& manga& only HR images\\
    \hline
    \multirow{3}{*}{1K images} & CelebA-HQ \cite{karras2017progressive} & 30,000 & 2017& face& only 1K HR images\\
    & Flickr-Faces-HQ \cite{karras2019style} & 70,000 & 2019& face & only 1K HR images \\
    & Flickr1024 \cite{wang2019flickr1024} & 1024 & 2019& natural & stereo image pairs\\
    \hline
    \multirow{4}{*}{2K images} & DIV2K \cite{agustsson2017ntire} & 900 & 2017& natural &S: Track 1 - bicubic downscaling\\ &&&&&S: Track 2 - unknown downscaling \\
    & Flickr2K \cite{Lim_2017_CVPR_Workshops} & 2650 & 2017& natural &S: Track 1 - bicubic downscaling\\ &&&&&S: Track 2 - unknown downscaling\\
    \hline
    \multirow{5}{*}{4K images} & UHDSR4K \cite{zhang2021benchmarking} & 8,099 & 2021& diverse & S: bicubic downscaling\\
    &&&&from& S: 3x blur+bicubic downscaling \\
    &&&&internet& S: 3x bicubic downscaling\\
    &&&&& + gaussian noise \\
    \hline
    \multirow{4}{*}{8K images} & DIV8K \cite{gu2019div8k} & 1,504 & 2019&natural&S: bicubic downscaling\\
    & UHDSR8K \cite{zhang2021benchmarking} & 2,966 & 2021 &diverse&S: bicubic downscaling\\
    &&&&from& S: 3x blur+bicubic downscaling \\
    &&&&internet& S: 3x bicubic downscaling\\
    &&&&& + gaussian noise\\
    \hline
    \multirow{13}{*}{other SR datasets} & BSDS500 \cite{arbelaez2010contour} & 500 & 2010&natural&S: bicubic downscaling\\
    & KITTI \cite{geiger2012we} & 3889 & 2012&traffic&diverse: HR images,\\
    &&&&& grayscale stereo image pairs, and \\
    &&&&& 3D laser scans \\
    & CASIA Webfaces \cite{yi2014learning} & 494,414 & 2014& faces& only face-detected HR images\\
    & CelebA \cite{liu2015deep} & 202,599 & 2015&faces&only HR images\\
    & VGGFace2 \cite{cao2018vggface2} & 3,310,000 & 2018 &faces&only HR images\\
    & PIRM \cite{blau20182018} & 200 & 2018 &natural&S: bicubic downscaling\\
    & OST300 \cite{wang2018recovering} & 300 & 2018 &outdoor&only HR images\\
    & HIDE \cite{Shen_2019_ICCV} & 8,422 & 2019 &natural& S: video sequences are averaged for\\
    &&&&&real and plausible motion blurred \\
    &&&&&LR images\\
    & RealSR \cite{cai2019toward} & 595 & 2019 & natural & LR-HR pairs with two cameras à  \\ &&&&&four focal lengths \\
    & SCI1k \cite{yang2021implicit} & 1,000 & 2021 &screen content&S: Variant 1 - bicubic downscaling\\
    &&&&& S: Variant 2 - JPEG compression \\
    \hline
    \multirow{7}{*}{Video SR datasets} & VID4 \cite{liu2013bayesian} & 4 & 2013 & natural& only HR image sequences\\
    & MCL-V \cite{lin2015mcl} & 12 & 2015 &natural&S: compressed and resized with \\&&&&&different settings\\
    & GOPRO \cite{Nah_2017_CVPR} & 33 & 2017 &natural&S: blurred LR and HR image pairs\\
    & Vimeo-90k \cite{xue2019video} & 90,000 & 2019&diverse& only HR image sequences\\
    & RealVSR \cite{yang2021real} & 500 & 2021 &natural&S: different scales by simulating two\\
    &&&&&cameras with different focal lengths\\
    \hline
    \end{tabular}
\end{center}
\end{table*}

\clearpage

\section*{Detailed Formula for SR3 Chapter}
In the chapter about Denoising Diffusion Probabilistic Models, a parameterized mean of $p_\theta \left( \mathbf{y}_{t-1} | \mathbf{y}_t, \mathbf{x}\right)$ was shown, which we do not find intuitive to derive. Therefore, we include the detailed formulation: 
\begin{equation*}
\label{eq_SR3Mean}
\begin{split}
      & \mu_\theta \left( \mathbf{x}, \mathbf{y}_t, \gamma_t \right) \\
      = & \frac{\sqrt{\gamma_{t-1}} \cdot \left( 1 - \alpha_t\right)}{1 - \gamma_t} \cdot \frac{1}{\sqrt{\gamma_t}} \cdot \left( \mathbf{y}_t - \sqrt{1-\gamma_t} \cdot \varphi_\theta \left( \mathbf{x}, \mathbf{y}_t, \gamma_t \right) \right) \\&+ \frac{\sqrt{\alpha_t} \cdot \left( 1 - \gamma_{t-1}\right)}{1-\gamma_t} \cdot \mathbf{y}_t \\
      = & \sqrt{\frac{\gamma_{t-1}}{\gamma_t}} \cdot \frac{1 - \alpha_t}{1 - \gamma_t} \cdot \left( \mathbf{y}_t - \sqrt{1-\gamma_t} \cdot \varphi_\theta \left( \mathbf{x}, \mathbf{y}_t, \gamma_t \right)\right) \\&+ \frac{\sqrt{\alpha_t} \cdot \left( \mathbf{y}_t - \gamma_{t-1} \cdot \mathbf{y}_t\right)}{1 - \gamma_t} \\
      = & \frac{1}{\sqrt{\alpha_t}} \cdot \frac{1 - \alpha_t}{1 - \gamma_t} \cdot \left( \mathbf{y}_t - \sqrt{1-\gamma_t} \cdot \varphi_\theta \left( \mathbf{x}, \mathbf{y}_t, \gamma_t \right)\right) \\&+ \frac{1}{\sqrt{\alpha_t}} \cdot\frac{\alpha_t \cdot \left( \mathbf{y}_t - \gamma_{t-1} \cdot \mathbf{y}_t\right)}{1 - \gamma_t}  \\
      = & \frac{1}{\sqrt{\alpha_t}} \cdot\frac{\mathbf{y}_t - \alpha_t \cdot\mathbf{y}_t}{1-\gamma_t} - \frac{\left( 1-\alpha_t\right) \cdot  \sqrt{1-\gamma_t} \cdot \varphi_\theta \left( \mathbf{x}, \mathbf{y}_t, \gamma_t \right)}{1-\gamma_t} \\&+\frac{1}{\sqrt{\alpha_t}}\cdot \frac{\alpha_t \cdot \mathbf{y}_t - \alpha_t \cdot \gamma_{t-1} \cdot \mathbf{y}_t }{1 - \gamma_t} \\
      = & \frac{1}{\sqrt{\alpha_t}} \left[ \frac{\mathbf{y}_t}{1-\gamma_t} - \frac{\left( 1-\alpha_t\right) \cdot \varphi_\theta \left( \mathbf{x}, \mathbf{y}_t, \gamma_t \right)}{\sqrt{1-\gamma_t}} - \frac{\gamma_{t} \cdot \mathbf{y}_t }{1 - \gamma_t}\right] \\
      = & \frac{1}{\sqrt{\alpha_t}} \left[ \mathbf{y}_t - \frac{1 - \alpha_t}{\sqrt{1 - \gamma_t}} \cdot \varphi_\theta \left( \mathbf{x}, \mathbf{y}_t, \gamma_t \right)   \right]
\end{split}
\end{equation*}

\begin{figure}[!h]
    \includegraphics[width=.16\textwidth]{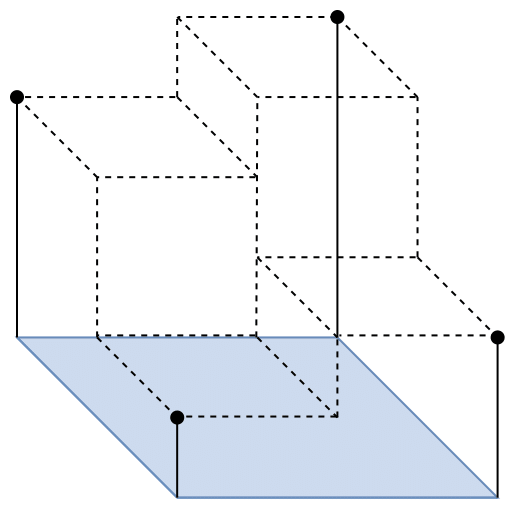}\hfill
    \includegraphics[width=.16\textwidth]{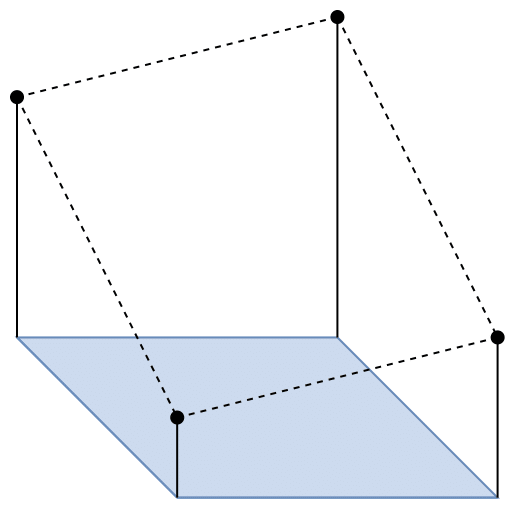}\hfill
    \includegraphics[width=.16\textwidth]{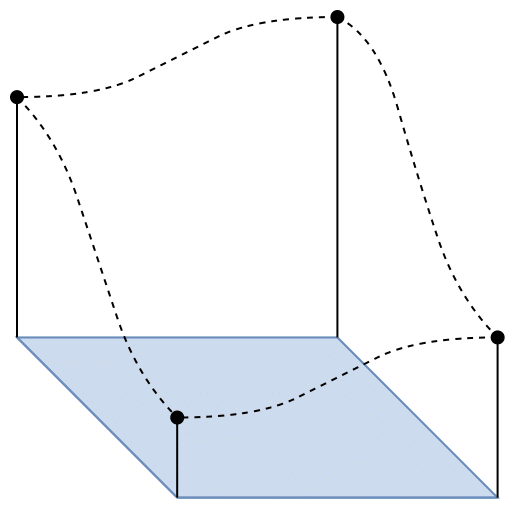}
    \caption{
    Visualization of discussed interpolation-based upsampling methods: Nearest-neighbor (left), bilinear (middle), and bicubic Interpolation (right). 
    It shows the spatial space of the image with a blue area and the ground-truth intensity values, which are the points connected to the edges with a solid line. 
    The dashed lines indicate the interpolated intensity values between the ground-truth intensity values.}
    \label{fig:intBasUps}
\end{figure}

\section*{Concerning Upsampling and Artifacts}
\label{sec:upsampling}
We examined standard upsampling methods in the main text, shown in \autoref{fig:intBasUps}.
Moreover, we introduced learning-based upsampling methods.
A well-known phenomenon with learning-based upsampling is artifacts like those investigated by Odena et al. (2016) \cite{odena2016deconvolution}.

\begin{figure}[!h]
    \begin{center}
        \includegraphics[width=.49\textwidth]{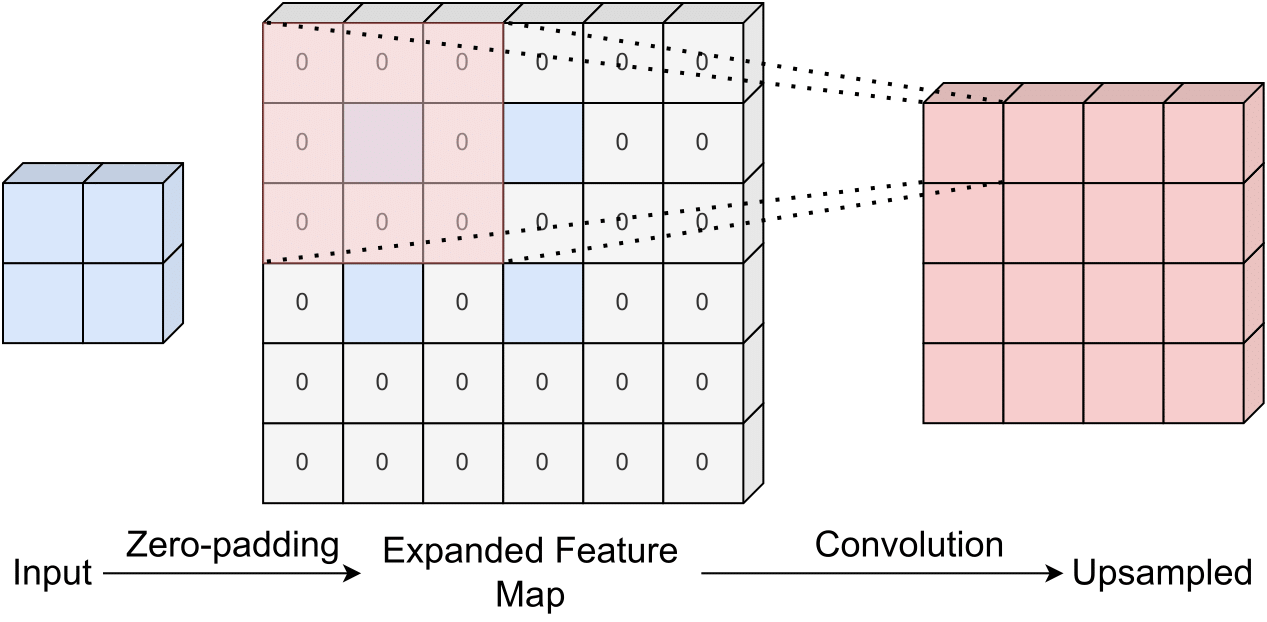}
        \caption{
        Visualization of the transposed convolution layer. It expands the given input with zero-padding and applies a convolution to process the features further.}
        \label{fig:transpose}
    \end{center}
\end{figure}

\begin{figure}[!h]
    \begin{center}
        \includegraphics[width=0.3\textwidth]{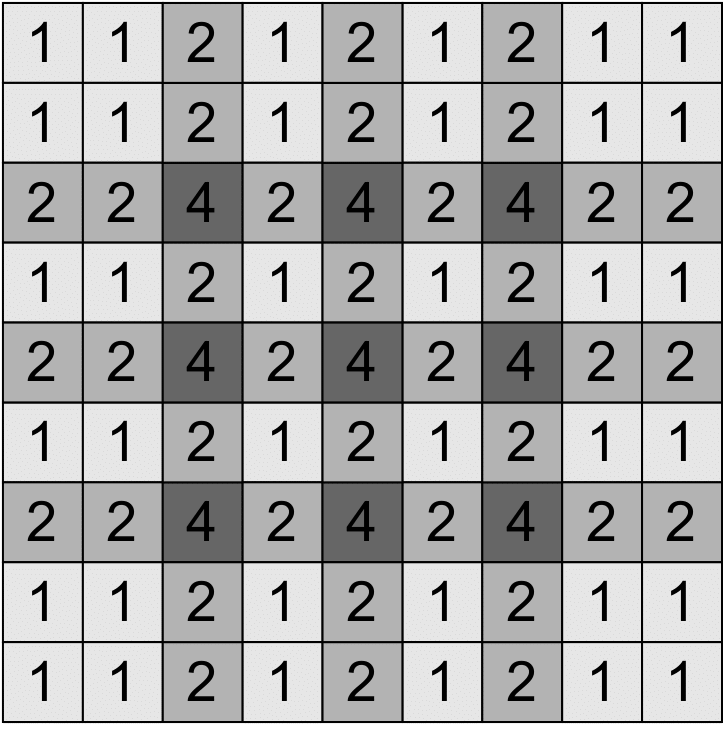}
        \caption{\label{fig:checkerboard}
        Visualization for artifacts. It shows how often a 3x3 kernel with a striding of two takes into account a position. As a result, it visits some positions multiple times, leading to the famous checkerboard pattern.}
    \end{center}
\end{figure}

They emerge when SR models use techniques like transposed convolution (as shown in \autoref{fig:transpose}) with specific settings.
In particular, there is an overlap if the output window size is not divisible by the stride. In other words, some positions in the input are visited multiple times, as shown in \autoref{fig:checkerboard}.

\begin{figure}[!h]
    \centering
    \includegraphics[width=.397\textwidth]{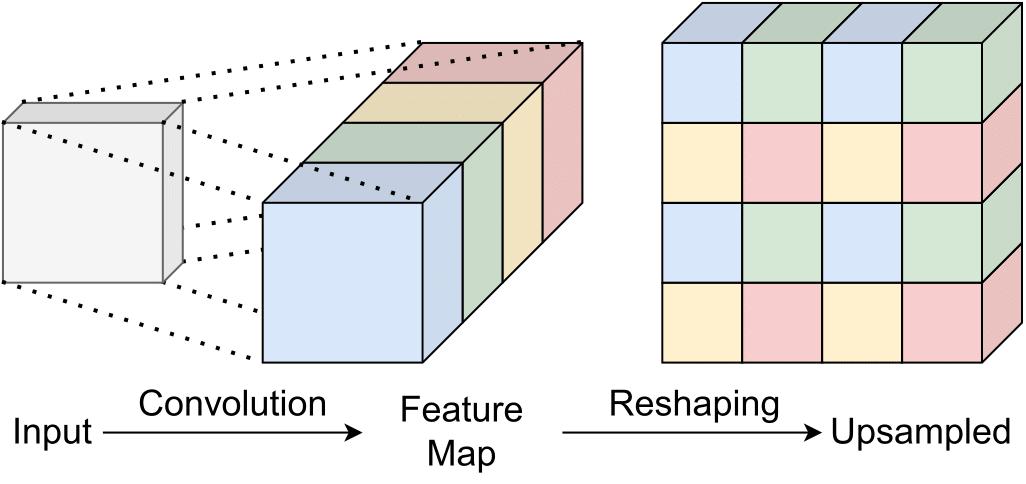}
    \caption{
    Visualization of the upsampling method Sub-pixel Layer, adapted from Shi et al. \cite{shi2016real}. 
    It generates a deep feature map by using convolutional operations. 
    Next, it rearranges the feature map to increase the spatial size while decreasing the channel dimension.}
    \label{fig:subpixel}
\end{figure}

In theory, the SR model should still be feasible to avoid the artifacts. 
But be that as it may, the model struggles to avoid these patterns in practice, and the kernels often learn this pattern.
One way to circumvent this problem is by applying a window size divisible by the stride. In that case, the transposed convolution is equal to a sub-pixel layer.
While sub-pixel layers help, it is still possible that this layer produces artifacts due to its shuffling, as shown in \autoref{fig:subpixel}.

\begin{figure}[!h]
    \begin{center}
        \includegraphics[width=.49\textwidth]{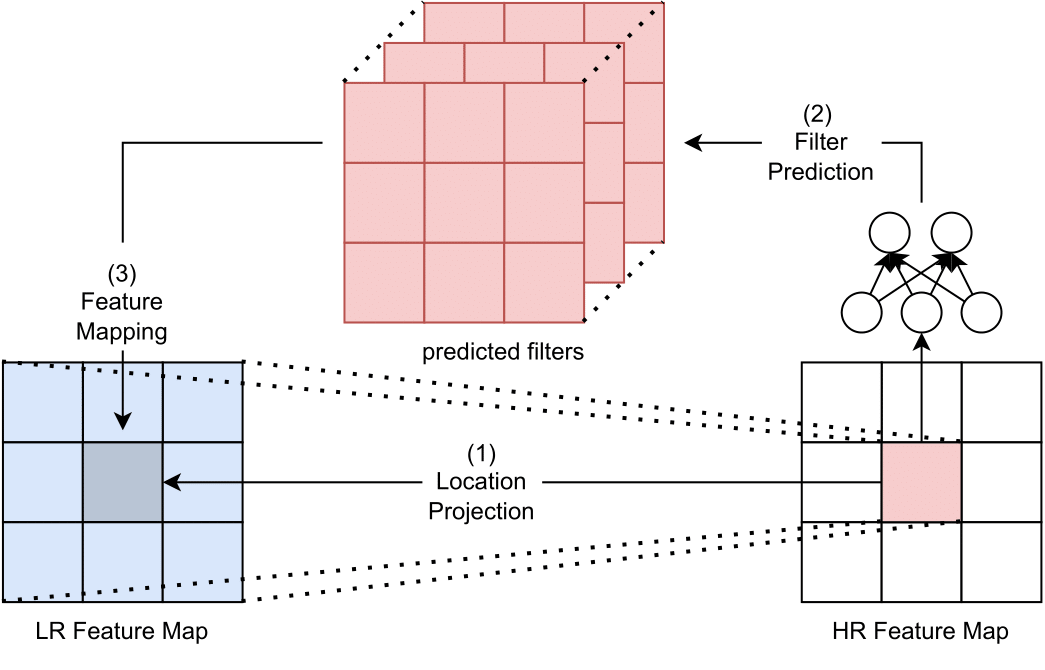}
        \caption{
        Visualization of meta-upscale, adapted from Hu et al. \cite{hu2019meta}. It can upscale a feature map with arbitrary scaling factors.
        It generates a higher resolution feature map iterative for each position in three steps: 
        (1) It locates the position in the original feature map via location projection. 
        The location in the HR feature map is divided component-wise by the scaling factor.
        (2) It generates a set of filter weights. 
        The prediction depends on the location in the HR feature map and its corresponding projection in the LR feature map.
        (3) It applies the weights derived by (2) to the LR feature map and returns the HR feature map value.}
        \label{fig:meta_upscale}
    \end{center}
\end{figure}

Additionally, we introduced the flexible and learning-based upsampling method meta-upscale, which is demonstrated visually in \autoref{fig:meta_upscale}. 
We also discussed the role of upsampling locations within a SR model. 
\autoref{tab:frameworkTab} shows the advantages and disadvantages of possible locations found in the literature.

\begin{table*}[!h]
\begin{center}
    \caption{\label{tab:frameworkTab}Comparison of upsampling frameworks (pre-, post-, progressive, and iterative up-and-down upsampling).}
    \begin{tabular}{|l l l |}
    \hline
    Framework & Pro & Contra\\
    \hline
    Pre- & \textbullet Reduced learning difficulty & \textbullet Slow feature extraction\\
    Upsampling & \enspace because it only needs to refine  & \enspace due to spatial space\\
    & \textbullet Can handle any random & \textbullet High memory consumption\\
    & \enspace size input without &  \enspace due to spatial space \\
    & \enspace forfeiting performance  & \\
    \hline
    Post- & \textbullet Fast in comparison & \textbullet Reduced complexity\\
    Upsampling & \enspace to pre-Upsampling  & \textbullet Harder to train due\\
     &  & \enspace to reduced spatial size\\
      &  & \enspace of feature extraction\\
       &  & \enspace maps\\
     &  & \textbullet Not suitable for\\
      &  & \enspace multi-scale\\
    \hline
    Iterative & \textbullet Improved performance & \textbullet Harder to interpret\\
    Up-and-down & \enspace in comparison to  & \textbullet Harder to train due\\
    Upsampling & \enspace pre- and post-upsampling  & to parallel learning\\
     & \textbullet Learns also the  & of up- and downsampling\\
      & \enspace degradation relationship  & \\
    \hline
    Progressive & \textbullet Improved performance & \textbullet Harder to integrate\\
    Upsampling & \textbullet Suitable for multi-scale  & \textbullet Needs modified learning\\
     &  & \enspace objective\\
    \hline
    \end{tabular}
\end{center}
\end{table*}

\newpage
~\newpage

\newpage

\begin{table}
\begin{center}
    \caption{\label{tab:exps_lightweightParams}Parameter and Multi-Adds comparison of models (for scaling 3).}
    \begin{tabular}{ |c | c | c | c |}
    \hline
    Method & Year & \# Parameters  & Multi-Adds\\
    & & [K] & [G]\\
    \hline
    SRCNN \cite{dong2015image} & 2014 & 57 & 52.7\\
    FSRCNN \cite{dong2016accelerating} & 2016 & 25& 5\\
    VDSR \cite{kim2016accurate} & 2016 & 665 & 612.6\\
    DRCN \cite{kim2016deeply} & 2016 & 1,774 & 17,974.3 \\
    MDSR \cite{lim2017enhanced}& 2017 & 8,000& - \\
    DRRN \cite{tai2017image} & 2017 & 297& 6,796.9 \\
    MemNet \cite{tai2017memnet} & 2017 & 677 & 2,662.4 \\
    CARN \cite{ahn2018fast}& 2018 & 1,592& 118.8\\
    CARN-M \cite{ahn2018fast}& 2018 & 412& 46.1\\
    IMDN \cite{hui2019lightweight} & 2019 & 703 & -\\
    RFDN \cite{liu2020residual}& 2020 & 550& 55.4\\
    XLSR \cite{ayazoglu2021extremely}& 2021 & 22& -\\
    \hline
    \end{tabular}
\end{center}
\end{table}

\section*{Beyond Classical Upsampling}

As mentioned in the previous section, we discussed typical upsampling methods (bicubic, transposed convolution, sub-pixel, and more) used in DL-based methods nowadays. 
We also want to mention essential alternatives that influence daily applications but still need to be met in current research areas with respect to deep learning. 

One example is Joint Bilateral Upsampling \cite{kopf2007joint} or JBU.
The idea is to have an edge-preserving upsampling method that operates in both domains, the LR and HR space. 
For example, given two filters, $f$ and $g$, $f$ is a spatial filter operating in the LR domain while $g$ is a range filter operating in the HR domain. 
This method is similar to meta-upscale, but the HR coordinates are not used to predict filters for the corresponding LR coordinates but to operate simultaneously at two different resolutions.
Further investigation of this concept for learning-based upsampling methods is an exciting avenue for future research.

Upon JBU, there is Bilateral Guided Upsampling \cite{chen2016bilateral} that fits local curves that map the
input to the output intensities within a patch.
After fitting, it is feasible to produce an HR output by evaluating local curves on the HR input.
In order to fit multiple local curves (since an image consists of many patches), the input is transformed into a bilateral grid, and the 3D array of affine matrices is used to obtain the best maps between the input and output color.
The authors showed that restricting their method makes it feasible to reduce this algorithm to JBU.

\newpage

\section*{Additional Figures: Residual Networks}

\begin{figure}[!h]
    \begin{center}
        \includegraphics[width=.49\textwidth]{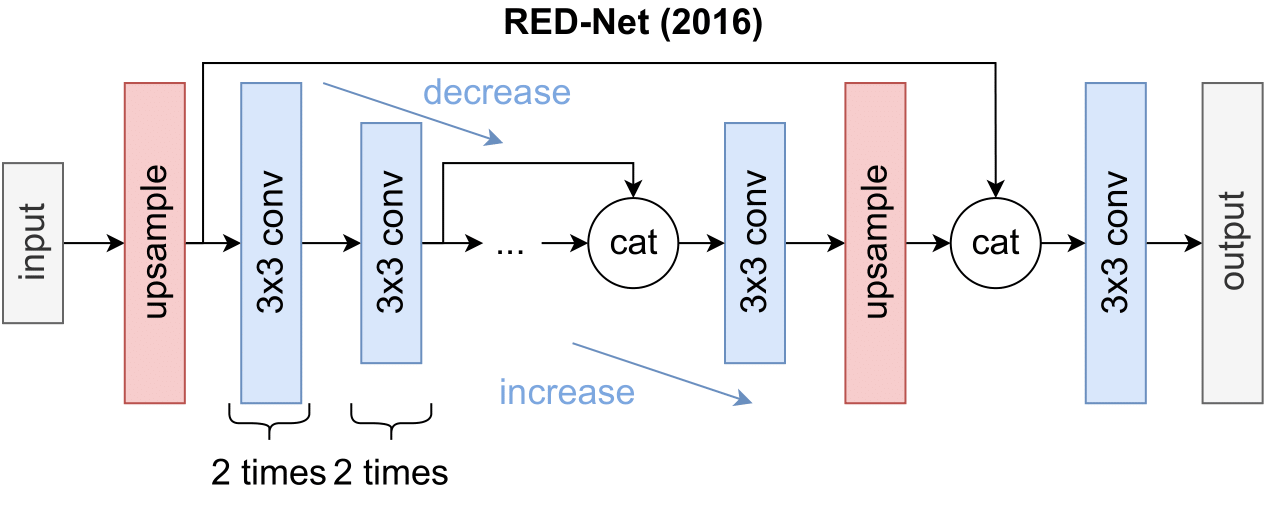}
        \caption{\label{fig:RED-Net}
        Architecture design of RED-Net \cite{mao2016image}. It is inspired by the U-Net \cite{ronneberger2015u} architecture and diminishes vanishing gradient effects by using residuals between the downsampling and upsampling paths.
        }
    \end{center}
\end{figure}

\begin{figure}[!h]
    \begin{center}
        \includegraphics[width=.358\textwidth]{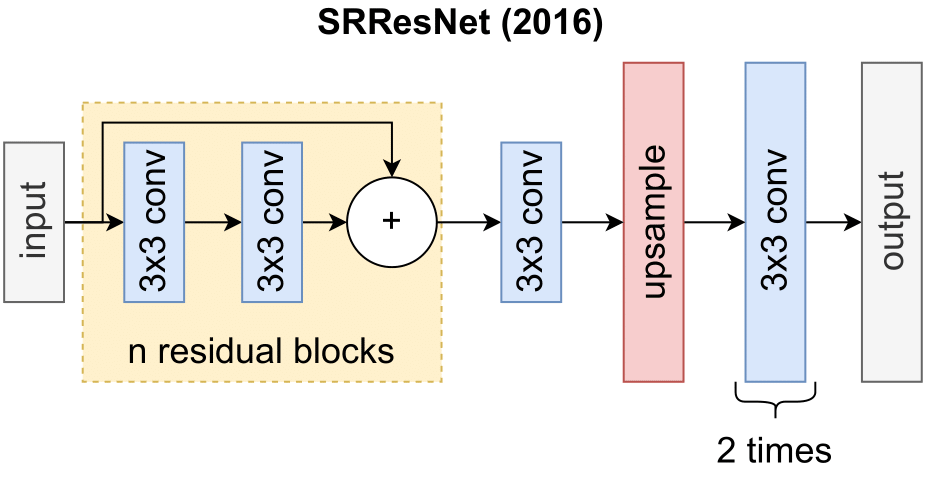}
        \caption{\label{fig:SRResNet}
        Architecture design of SRResNet \cite{ledig2017photo}. It is inspired by ResNet \cite{he2016deep} and applies multiple residual units to propagate information from early layers to later layers.
        }
    \end{center}
\end{figure}

\begin{figure}[!h]
    \begin{center}
        \includegraphics[width=.459\textwidth]{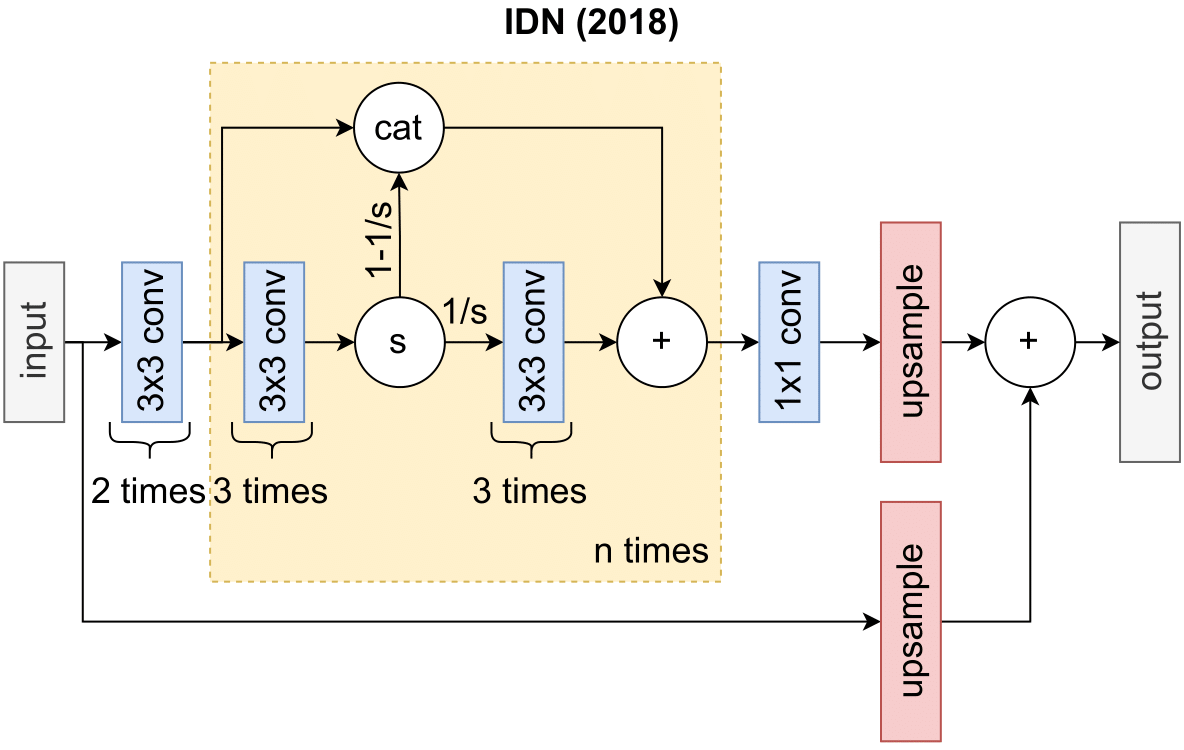}
        \caption{\label{fig:IDN}
        Architecture design of IDN \cite{hui2018fast}. It splits the feature maps in the middle part and processes one part further while the other part is concatened to the residual information path. The notation ``c'' denotes concatenation and ``s'' denotes the split operation.
        }
    \end{center}
\end{figure}

This section contains additional visualizations of residual networks presented in the main text, namely RED-Net \cite{mao2016image} (see \autoref{fig:RED-Net}), SRResNet \cite{ledig2017photo} (see \autoref{fig:SRResNet}), and IDN \cite{hui2018fast} (see \autoref{fig:IDN}).

\newpage

\section*{Additional Figures: Recurrent CNN Networks}

\begin{figure}[!h]
    \begin{center}
        \includegraphics[width=.35\textwidth]{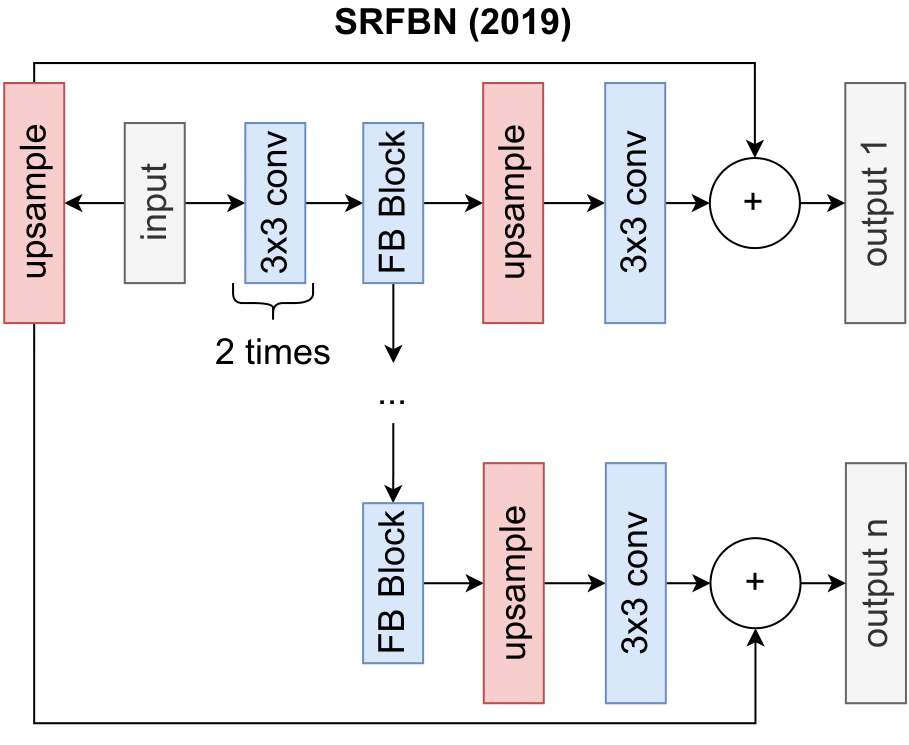}
        \caption{\label{fig:SRFBN}
        Architecture design of SRFBN \cite{li2019feedback}. It utilizes FeedBack (FB) blocks that uses multiple Iterative up-and-down upsamplings with dense connections. For each iteration, the SRFBN generates a SR image. They are used to train the network with curriculum learning.
        }
    \end{center}
\end{figure}

\begin{figure}[!h]
    \begin{center}
        \includegraphics[width=.49\textwidth]{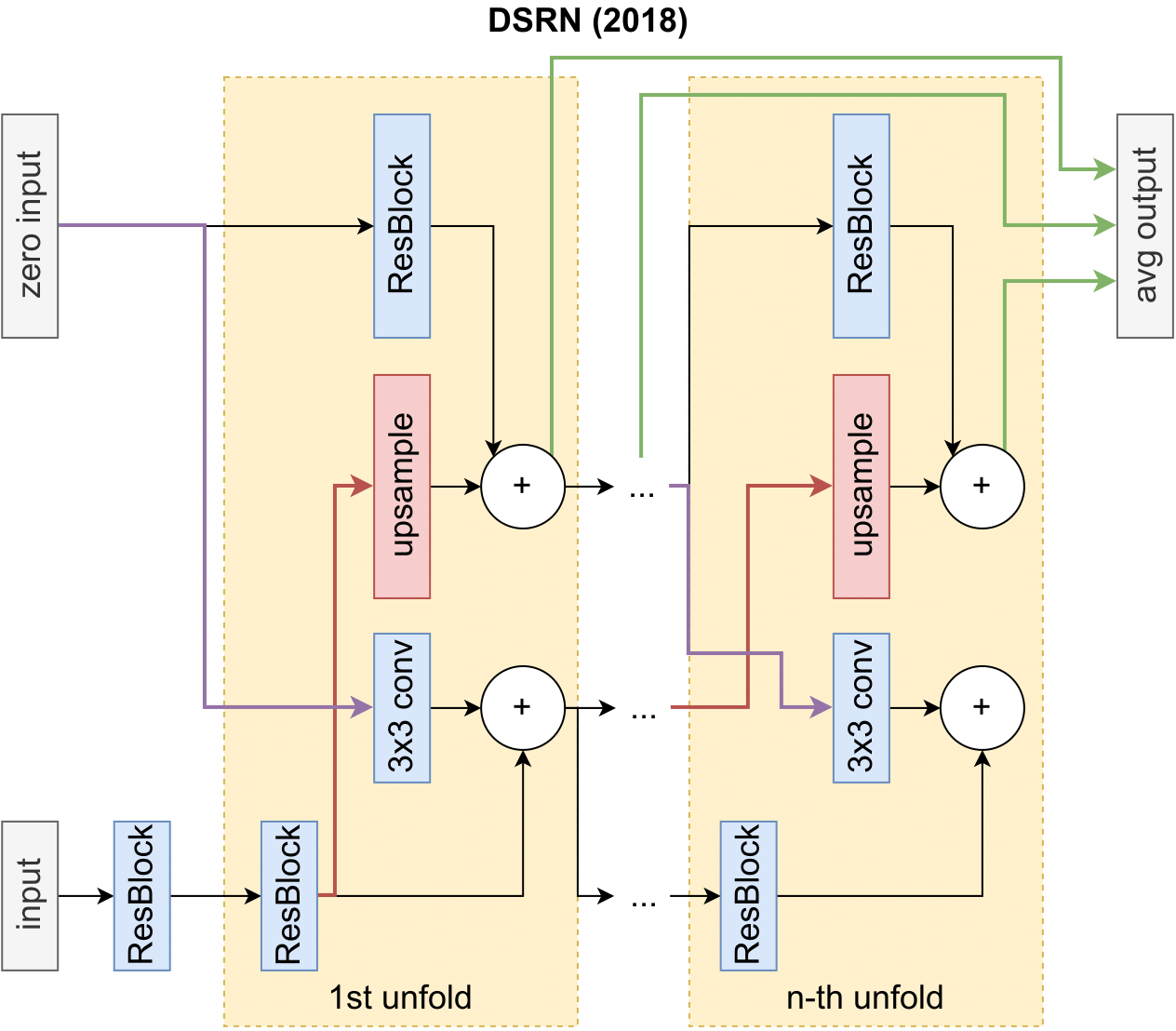}
        \caption{\label{fig:DSRN}
        Architecture design of DSRN \cite{han2018image}. It uses a dual-state design, where the state-to-state transitions (LR to LR and HR to HR) are realized with residual unit like paths. To interchange information between the to spaces, it uses a convolution layer to downsample information (HR to LR) and a transposed convolution to upsample information (LR to HR).
        }
    \end{center}
\end{figure}

This section contains additional visualizations of recurrent CNN networks presented in the main text, namely SRFBN \cite{li2019feedback} (see \autoref{fig:SRFBN}) and DSRN \cite{han2018image} (see \autoref{fig:DSRN}).

\newpage

\section*{Notes about Lightweight Models}

\begin{figure}[!h]
    \begin{center}
        \includegraphics[width=.49\textwidth]{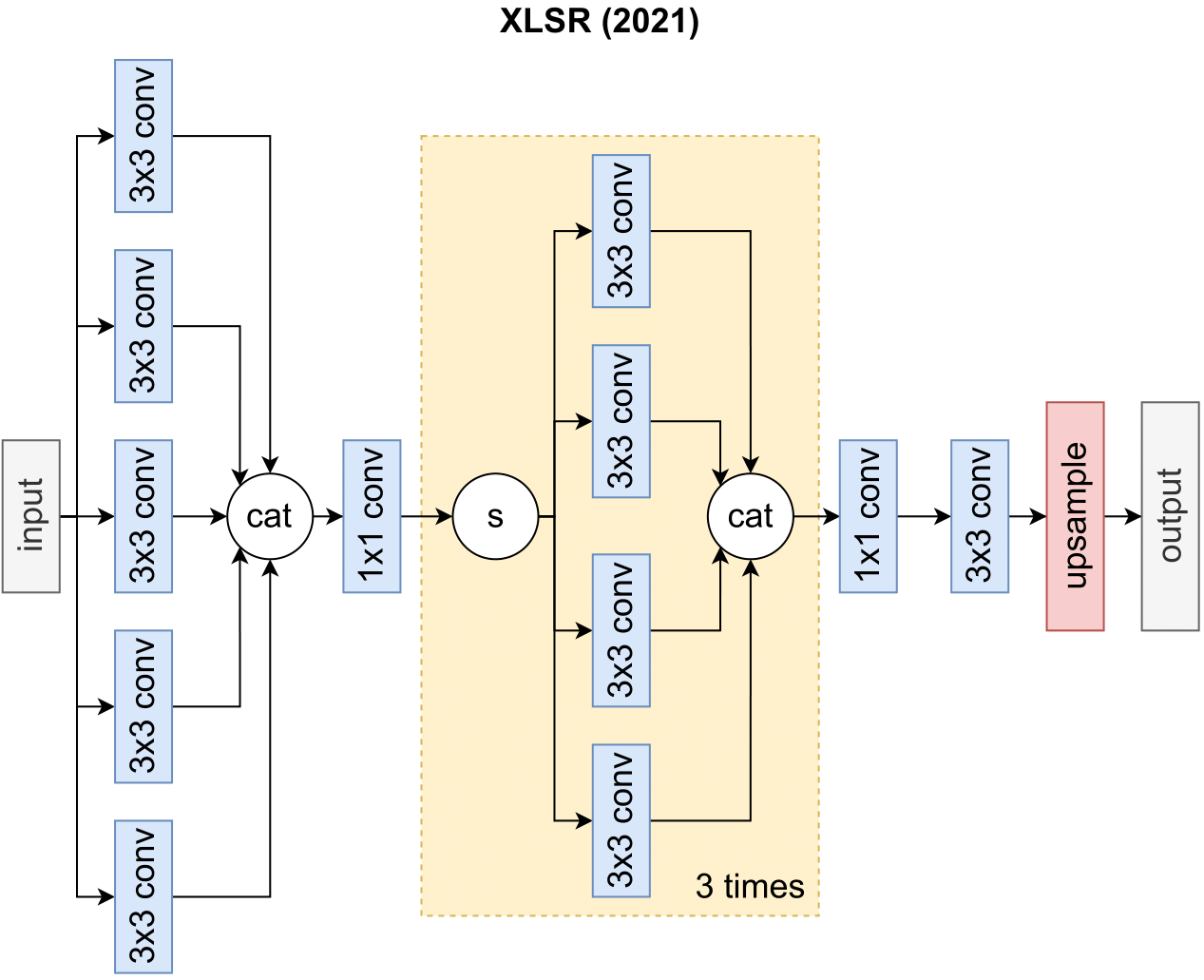}
        \caption{\label{fig:XLSR}
        Architecture design of XLSR \cite{ayazoglu2021extremely}. It applies multi-paths derive different parts of the feature map and 1x1 convolution to combine the features pixel-wise again. The operator ``s'' denotes a split operation. 
        }
    \end{center}
\end{figure}

This section contains a visualization of XLSR \cite{ayazoglu2021extremely} (see \autoref{fig:XLSR}) as well as \autoref{tab:exps_lightweightParams} that compares different lightweight models in terms of parameters and multi-adds.

\section*{Additional Figures: NAS}
\begin{figure}[!h]
    \begin{center}
        \includegraphics[width=.49\textwidth]{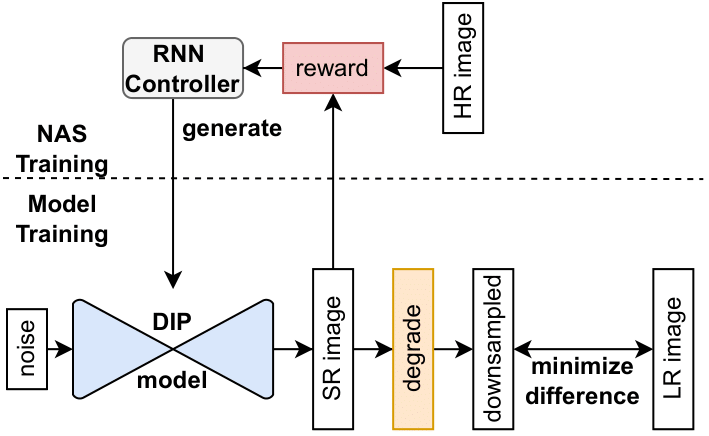}
        \caption{\label{fig:NASDIP}
        Visualization of NAS-DIP \cite{chen2020dip}. A controller derives an architecture. The architecture is then trained in a fashion described by DIP \cite{ulyanov2018deep}. It optimizes the loss function obtained by downsampling the SR output and comparing it with the corresponding LR output. The SR output is then used with the HR output to determine the reward.
        }
    \end{center}
\end{figure}

This section presents \autoref{fig:NASDIP}, a visualization of the NAS approach NAS-DIP \cite{chen2020dip}.

\newpage

\section*{Additional Figures: Unsupervised SR}
\begin{figure}[!h]
    \begin{center}
        \includegraphics[width=.49\textwidth]{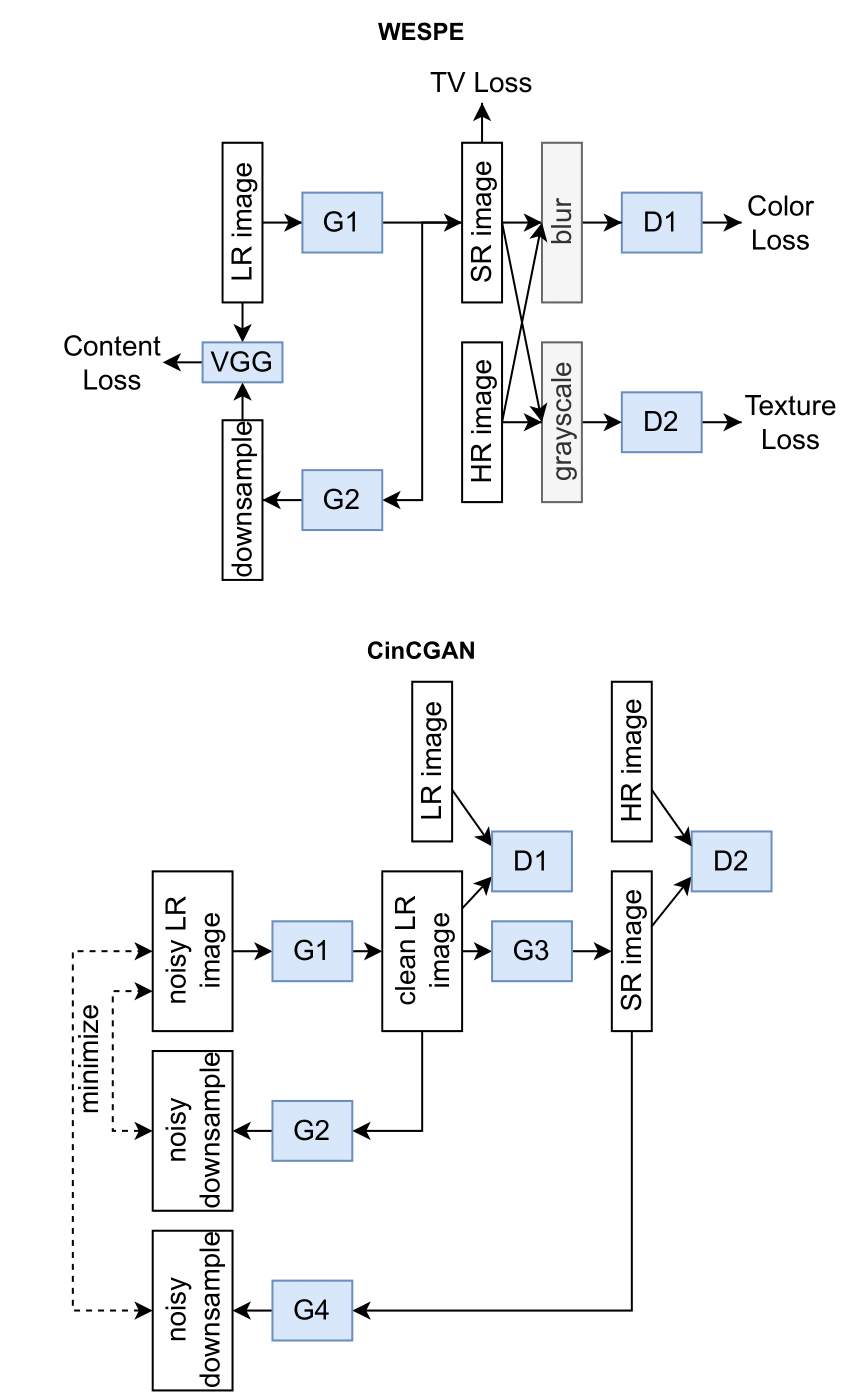}
        \caption{\label{fig:weaklySR}
        Visualization of GAN approaches WESPE \cite{ignatov2018wespe} and CinCGAN \cite{yuan2018unsupervised}. WESPE learns LR to HR relationship with generator G1, D1 and D2. It also learns the SR to LR relationship with G2 and content loss. CinCGAN works similar but uses an cycle-in-cycle adaption of CycleGAN \cite{zhu2017unpaired}. It learns the noisy LR to clean LR relationship and the inverse direction. Upon this, it learns the LR to HR and the HR to noisy LR relationship.
        }
    \end{center}
\end{figure}

This section presents \autoref{fig:weaklySR}, a visualization of the weakly unsupervised SR methods WESPE \cite{ignatov2018wespe} and CinCGAN \cite{yuan2018unsupervised}.

\newpage

\section*{Visual SR Results}
In addition to a benchmark table and various explanations, we want to give a visual impression of the methods presented in this work.
Therefore, we tested a selection of trained SR models on five images from the Set14 dataset \cite{zeyde2010single} with an upscaling factor of four and provided a visualization in the following. 
Moreover, we include the ground-truth image pairs (LR-HR) and a simple bicubic interpolated LR image for comparison. Thus, each visualization contains three images plus nine reconstructions of SR models.

The selection of trained SR models is: SRCNN \cite{dong2015image}, FSRCNN \cite{dong2016accelerating}, ESRGAN \cite{wang2018esrgan}, CARN-M \cite{ahn2018image}, CARN \cite{ahn2018image}, IDN \cite{hui2018fast}, IMDN \cite{hui2019lightweight}, SwinIR \cite{liu2021swin}, and CAR (with EDSR) \cite{sun2020learned}.
The selection shows not only the state-of-the-art methods but also gives a glimpse into the history of SR models based on deep learning as well as different categories of models (simple, lightweight, residual, GAN-based, and Transformer-based networks).
Each of the following visualizations incorporates two "zoomed-in" areas to highlight the reconstruction details provided by the models.

\begin{figure*}
    \begin{center}
        \includegraphics{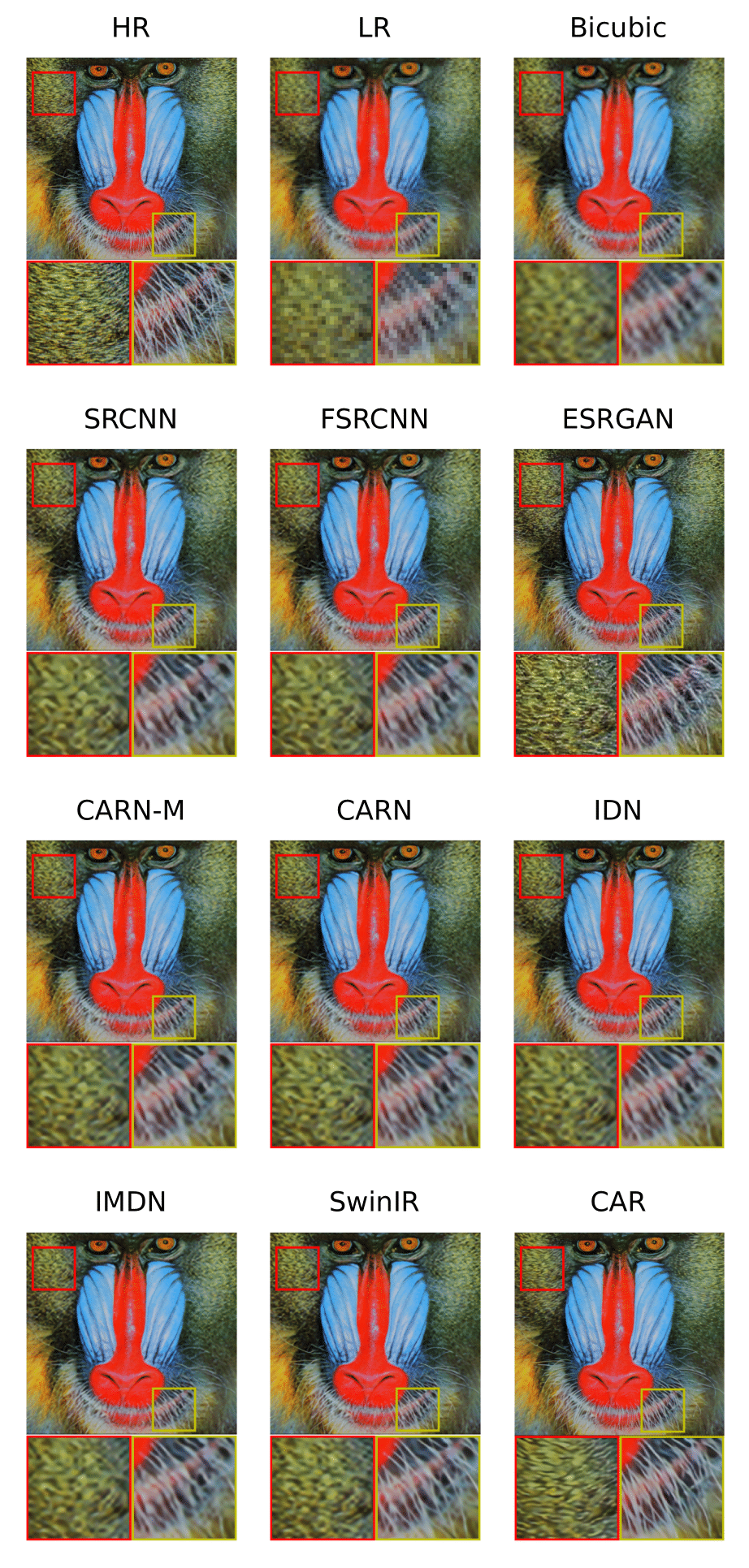}
        \caption{\label{fig:visual-0}
        Comparison of SR models for ''baboon'' from Set14 dataset (4x upscaling) \cite{zeyde2010single}. Two detail levels are shown in addition to the overall image. Shown SR models are SRCNN \cite{dong2015image}, FSRCNN \cite{dong2016accelerating}, ESRGAN \cite{wang2018esrgan}, CARN-M \cite{ahn2018image}, CARN \cite{ahn2018image}, IDN \cite{hui2018fast}, IMDN \cite{hui2019lightweight}, SwinIR \cite{liu2021swin}, and CAR \cite{sun2020learned}.}
        \label{fig:transposed}
    \end{center}
\end{figure*}

\begin{figure*}
    \begin{center}
        \includegraphics{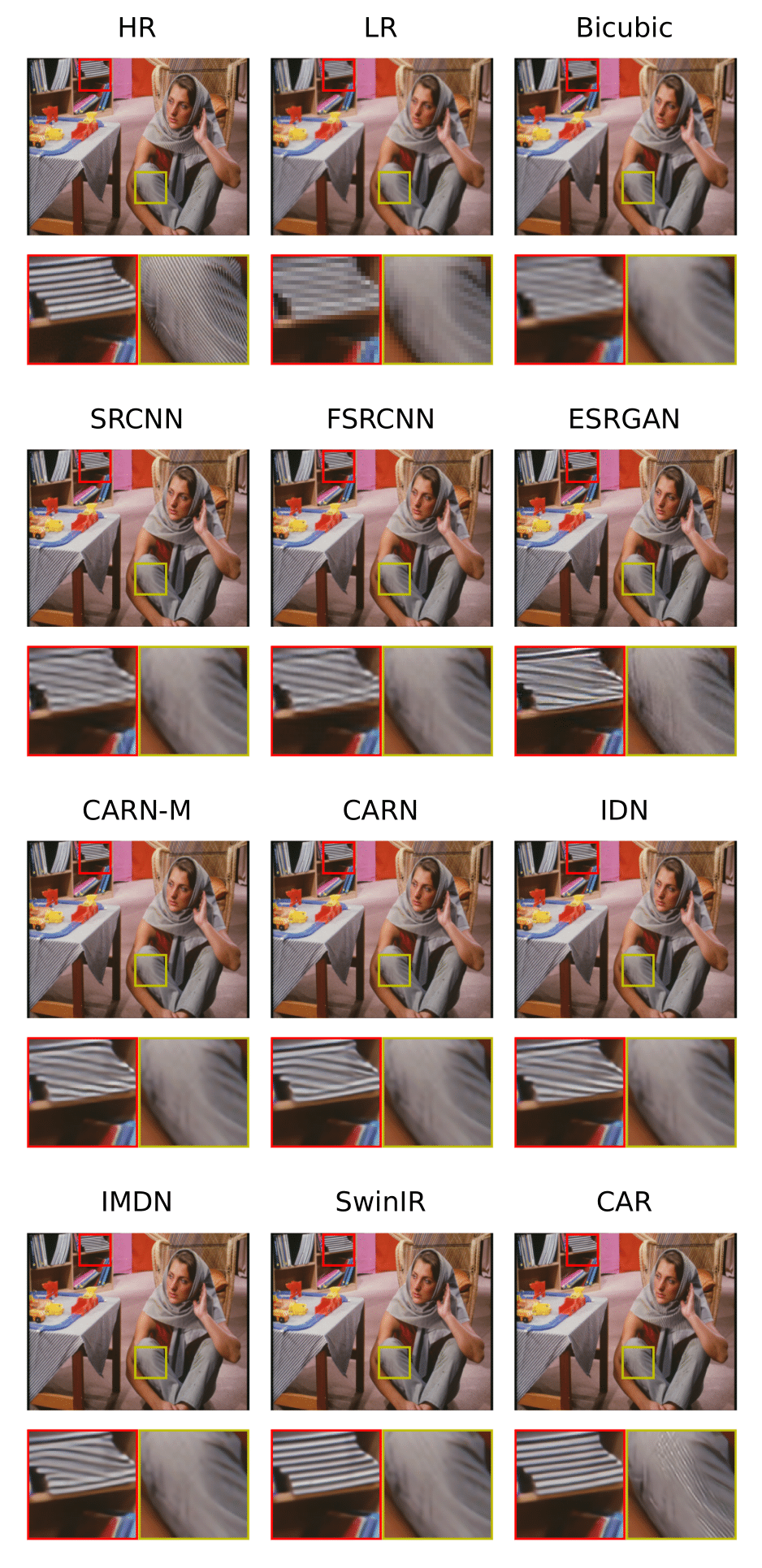}
        \caption{\label{fig:visual-1}
        Comparison of SR models for ''barbara'' from Set14 dataset (4x upscaling) \cite{zeyde2010single}. Two detail levels are shown in addition to the overall image. Shown SR models are SRCNN \cite{dong2015image}, FSRCNN \cite{dong2016accelerating}, ESRGAN \cite{wang2018esrgan}, CARN-M \cite{ahn2018image}, CARN \cite{ahn2018image}, IDN \cite{hui2018fast}, IMDN \cite{hui2019lightweight}, SwinIR \cite{liu2021swin}, and CAR \cite{sun2020learned}.}
        \label{fig:transposed}
    \end{center}
\end{figure*}

\begin{figure*}
    \begin{center}
        \includegraphics{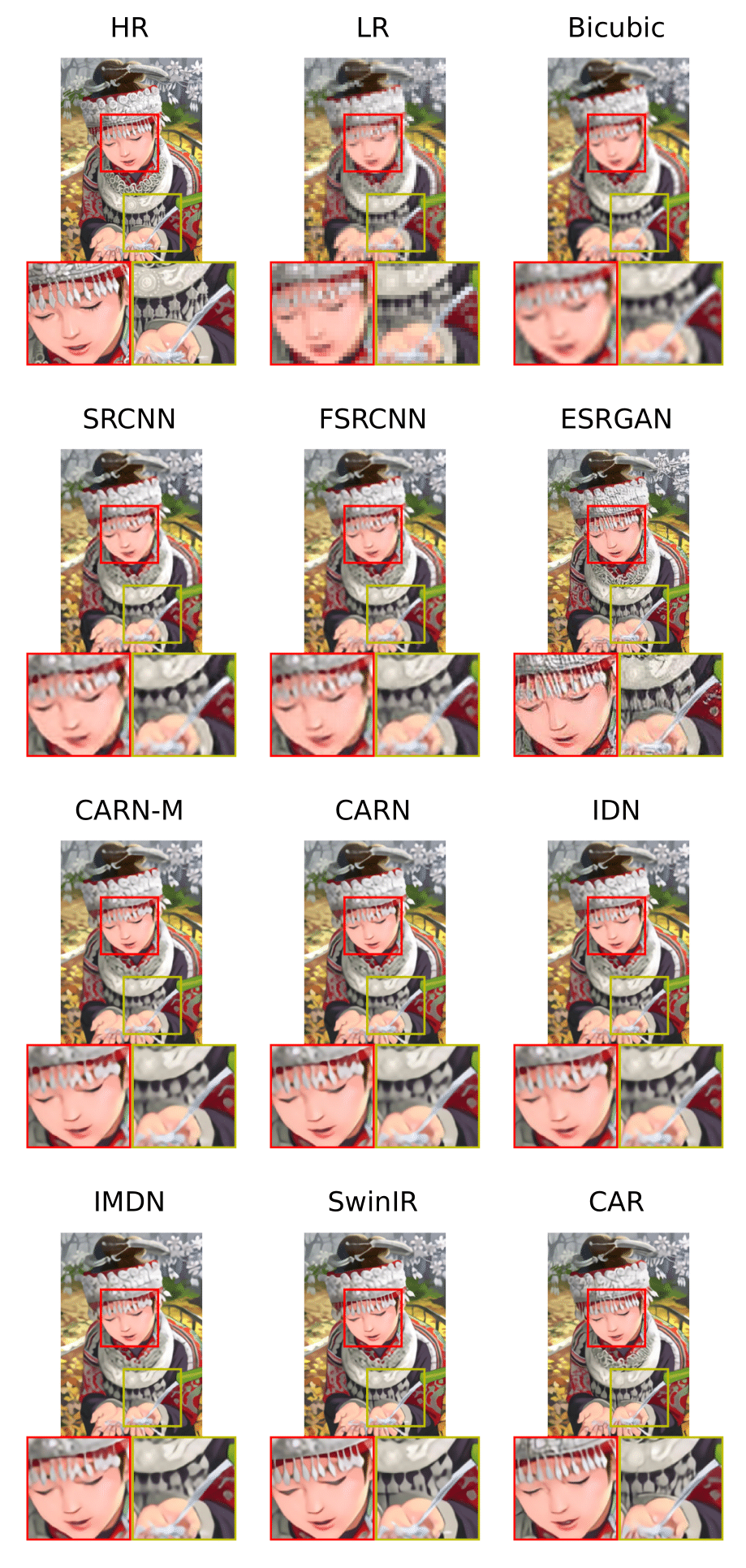}
        \caption{\label{fig:visual-2}
        Comparison of SR models for ''comic'' from Set14 dataset (4x upscaling) \cite{zeyde2010single}. Two detail levels are shown in addition to the overall image. Shown SR models are SRCNN \cite{dong2015image}, FSRCNN \cite{dong2016accelerating}, ESRGAN \cite{wang2018esrgan}, CARN-M \cite{ahn2018image}, CARN \cite{ahn2018image}, IDN \cite{hui2018fast}, IMDN \cite{hui2019lightweight}, SwinIR \cite{liu2021swin}, and CAR \cite{sun2020learned}.}
        \label{fig:transposed}
    \end{center}
\end{figure*}

\begin{figure*}
    \begin{center}
        \includegraphics{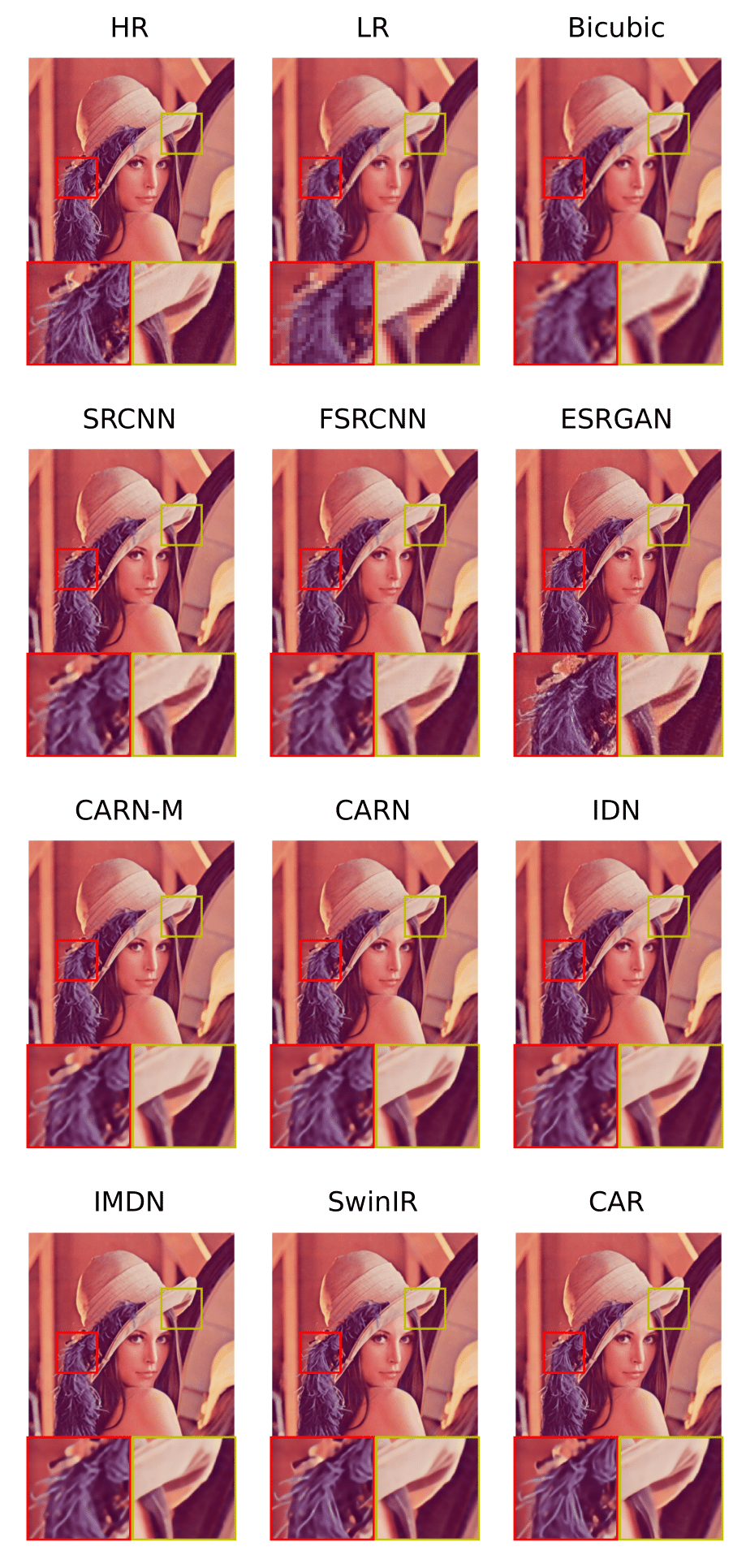}
        \caption{\label{fig:visual-3}
        Comparison of SR models for ''lenna'' from Set14 dataset (4x upscaling) \cite{zeyde2010single}. Two detail levels are shown in addition to the overall image. Shown SR models are SRCNN \cite{dong2015image}, FSRCNN \cite{dong2016accelerating}, ESRGAN \cite{wang2018esrgan}, CARN-M \cite{ahn2018image}, CARN \cite{ahn2018image}, IDN \cite{hui2018fast}, IMDN \cite{hui2019lightweight}, SwinIR \cite{liu2021swin}, and CAR \cite{sun2020learned}.}
        \label{fig:transposed}
    \end{center}
\end{figure*}

\begin{figure*}
    \begin{center}
        \includegraphics{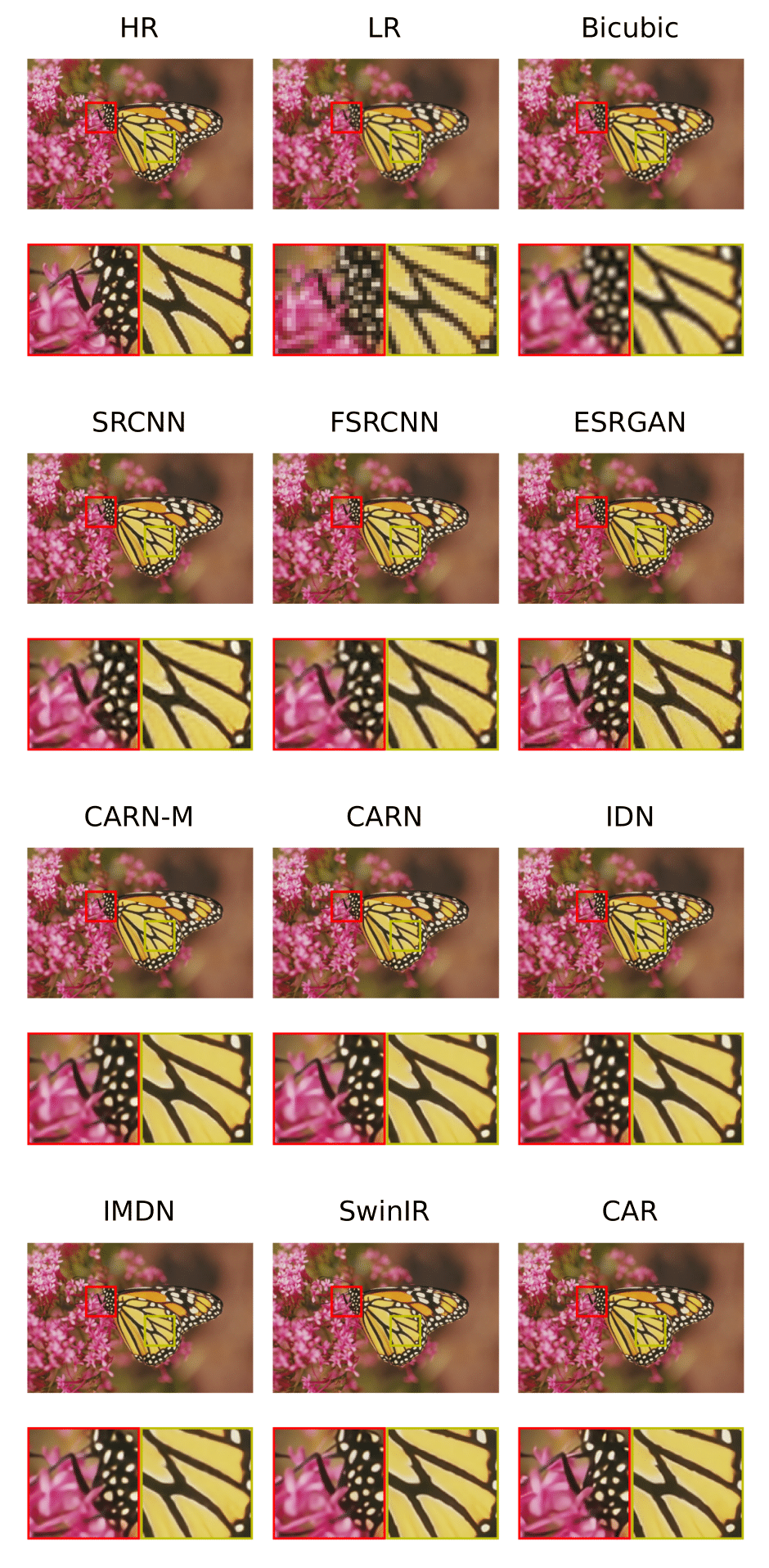}
        \caption{\label{fig:visual-4}
        Comparison of SR models for ''monarch'' from Set14 dataset (4x upscaling) \cite{zeyde2010single}. Two detail levels are shown in addition to the overall image. Shown SR models are SRCNN \cite{dong2015image}, FSRCNN \cite{dong2016accelerating}, ESRGAN \cite{wang2018esrgan}, CARN-M \cite{ahn2018image}, CARN \cite{ahn2018image}, IDN \cite{hui2018fast}, IMDN \cite{hui2019lightweight}, SwinIR \cite{liu2021swin}, and CAR \cite{sun2020learned}.}
        \label{fig:transposed}
    \end{center}
\end{figure*}

\end{document}